\documentclass[acmlarge]{acmart}
\usepackage{color,array}

\usepackage{graphicx}

\usepackage{mathrsfs}
\usepackage{bookmark}
\usepackage{amsmath,bm}

\usepackage{tabularx}
\usepackage{lineno}
\usepackage{epsfig}
\usepackage{chngpage}
\usepackage{float}
\usepackage{graphicx}
\usepackage{array}
\usepackage{diagbox}
\usepackage{epstopdf}
\usepackage{algorithm}
\usepackage{algorithmicx}
\usepackage{algpseudocode}
\usepackage{threeparttable}
\usepackage{dsfont}
\usepackage{subfigure}
\usepackage{multirow}
\usepackage{longtable}
\usepackage{booktabs}
\usepackage{multirow}
\usepackage{amssymb}
\usepackage{booktabs}
\usepackage{amsthm}
\usepackage{url}
\usepackage{footnote}
\makesavenoteenv{table}
\AtBeginDocument{%
	\providecommand\BibTeX{{%
			\normalfont B\kern-0.5em{\scshape i\kern-0.25em b}\kern-0.8em\TeX}}}

\setcopyright{acmcopyright}
\copyrightyear{2022}
\acmYear{2022}
\acmDOI{XXXXXXX.XXXXXXX}

\acmJournal{TIST}
\acmVolume{0}
\acmNumber{0}
\acmArticle{0}
\acmMonth{4}

\begin{document}
	
	\title{Graph Learning for Anomaly Analytics: Algorithms, Applications, and Challenges}
	
	\author{Jing Ren}
	\affiliation{%
		\department{Institute of Innovation, Science and Sustainability}
		\institution{Federation University Australia}
		\city{Ballarat}
		\state{VIC}
		\postcode{3353}
		\country{Australia}
	}
	\email{ch.yum@outlook.com}

\author{Feng Xia}
\authornote{Corresponding author}
%\orcid{1234-5678-9012-3456}
\affiliation{%
	\department{Institute of Innovation, Science and Sustainability}
	\institution{Federation University Australia}
	\city{Ballarat}
	\state{VIC}
	\postcode{3353}
	\country{Australia}
}
\email{f.xia@ieee.org}

	\author{Ivan Lee}
	\affiliation{%
		\department{STEM}
		\institution{University of South Australia}
		\city{Adelaide}
		\state{SA}
		\country{Australia}
		\postcode{5001}}
	\email{ivan.lee@unisa.edu.au}
	
	\author{Azadeh Noori Hoshyar}
\affiliation{%
	\department{Institute of Innovation, Science and Sustainability}
	\institution{Federation University Australia}
	\city{Brisbane}
	\state{QLD}
	\postcode{4000}
	\country{Australia}
}
	\email{a.noorihoshyar@federation.edu.au}
	
	\author{Charu C. Aggarwal}
	\affiliation{%
		\institution{IBM T. J. Watson Research Center}
		\city{New York}
		\country{USA}
		\postcode{10598}}
	\email{charu@us.ibm.com}
	
	\renewcommand{\shortauthors}{Ren, et al.}
	
	\begin{abstract}
		Anomaly analytics is a popular and vital task in various research contexts, which has been studied for several decades. At the same time, deep learning has shown its capacity in solving many graph-based tasks like, node classification, link prediction, and graph classification. Recently, many studies are extending graph learning models for solving anomaly analytics problems, resulting in beneficial advances in graph-based anomaly analytics techniques. In this survey, we provide a comprehensive overview of graph learning methods for anomaly analytics tasks. We classify them into four categories based on their model architectures, namely graph convolutional network (GCN), graph attention network (GAT), graph autoencoder (GAE), and other graph learning models. The differences between these methods are also compared in a systematic manner. Furthermore, we outline several graph-based anomaly analytics applications across various domains in the real world. Finally, we discuss five potential future research directions in this rapidly growing field.
	\end{abstract}
	
	


	\maketitle
	
\section{Introduction}
Anomalies, which are also known as outliers, commonly exist in various real-world networks~\cite{chandola2009anomaly}, such as fake reviews in opinion networks~\cite{zafarani2019fake}, fake news in social networks~\cite{yu2022graph}, outlier members in collaboration networks~\cite{yu2020optimize,wan2019your}, flash crowds in traffic networks~\cite{kong2018lotad}, socially selfish nodes in mobile networks~\cite{xia2016signaling}, and network intrusions in computer networks~\cite{ding2012intrusion}. The exploration of anomaly detection research is dating back to 1960s and it has been a popular research field for several decades~\cite{grubbs1969procedures}. With the increasing demand and broad applications in different domains, anomaly analytic plays an increasingly important role in various communities such as data mining and machine learning.
	
With the advancement of deep learning, graph learning is proposed subsequently, which is coined for deep learning-based models that are applied into graph-structured data~\cite{zhang2020deep,xia2021graph}. Due to its convincing performance and explainability, recent years have witnessed, in varied disciplines, an increasing number of studies focusing on anomaly detection and prediction tasks by utilizing deep graph models~\cite{wang2019convolutional,zhou2020graph}, which is not limited to shallow network embedding such as random walks~\cite{hou2020network,xia2019random}. As a unique non-Euclidean data structure, graphs are able to represent entities and their relationships in different kinds of scenarios. However, this research direction faces several inevitable problem complexities to all detection methods when applying deep learning and artificial intelligence in real-world networks~\cite{liu2018artificial,wang2021decoupling}.
	
\begin{itemize}
		\item \textbf{Irregularity of graph structure.} Unlike other regular structured data, such as text, sequences, and images, nodes in a graph are unordered and can have distinct neighborhoods, which makes the structure of graphs irregular. Therefore, some traditional deep learning architectures cannot be directly applied, such as convolution and pooling operation in convolutional neural networks (CNNs)~\cite{niepert2016learning}.
		\item \textbf{Heterogeneous anomaly classes.} The types of nodes and links are generally not unitary in a graph, which leads to the emergence of heterogeneous information networks (HINs). HINs usually incorporate more complex information among entities and relationships, especially those containing different modalities~\cite{shi2016survey}, which are very important in identifying different types of anomalies in a specific graph.
		\item \textbf{Scalability to real-world networks.} Nowadays, real-world networks such as social networks are composed of millions or even billions of nodes, edges, and attribute information~\cite{xia2017big}. This kind of large-scale network definitely increases computational complexity. Therefore, it is imperative to devise scalable models having a linear time complexity with respect to the graph size.
		\item \textbf{Label scarcity.} Compared with manually generated graph data, there are mainly two reasons for the sparsity of real-world networks. The first one is the scale-free network structure nature that the degree of nodes in most real-world networks follows long-tailed distribution~\cite{zeng2017science}. The other one is limited by the collection technology and privacy protection in the process of crawling data. Moreover, due to the lack of labeled datasets, devising unsupervised anomaly detection models is becoming  important.
		\item \textbf{Diverse types of anomalies.} Several types of anomalies have been explored such as node, edge, subgraph, and path (shown in Fig~\ref{anomalytype}). Node anomalies are  entities that show anomalous behaviours in the whole graph compared with other nodes, e.g., users who spread fake news in social networks. Other types of anomalies have similar concepts and their own real-world applications. Here, subgraph anomaly is difficult to detect because the individual nodes could show normal behaviours when extracted from an anomalous subgraph.
		
\end{itemize}
	
There have been a line of deep anomaly detection research demonstrating significantly better performance than conventional models on solving the above-mentioned challenges. Despite the fact that the adopted technologies vary from Graph Convolution Networks (GCNs) to Graph AutoEncoder (GAEs), most methods focus on detecting or predicting an anomaly in a specific situation due to the complexity of existing anomalies. To the best of our knowledge, little attention has been devoted to summarizing these methods in a comprehensive way and clearly analyzing how they are applied to solve real-world application scenarios.
\begin{figure*}[htb]
		\centering
		\includegraphics[width=0.9\textwidth]{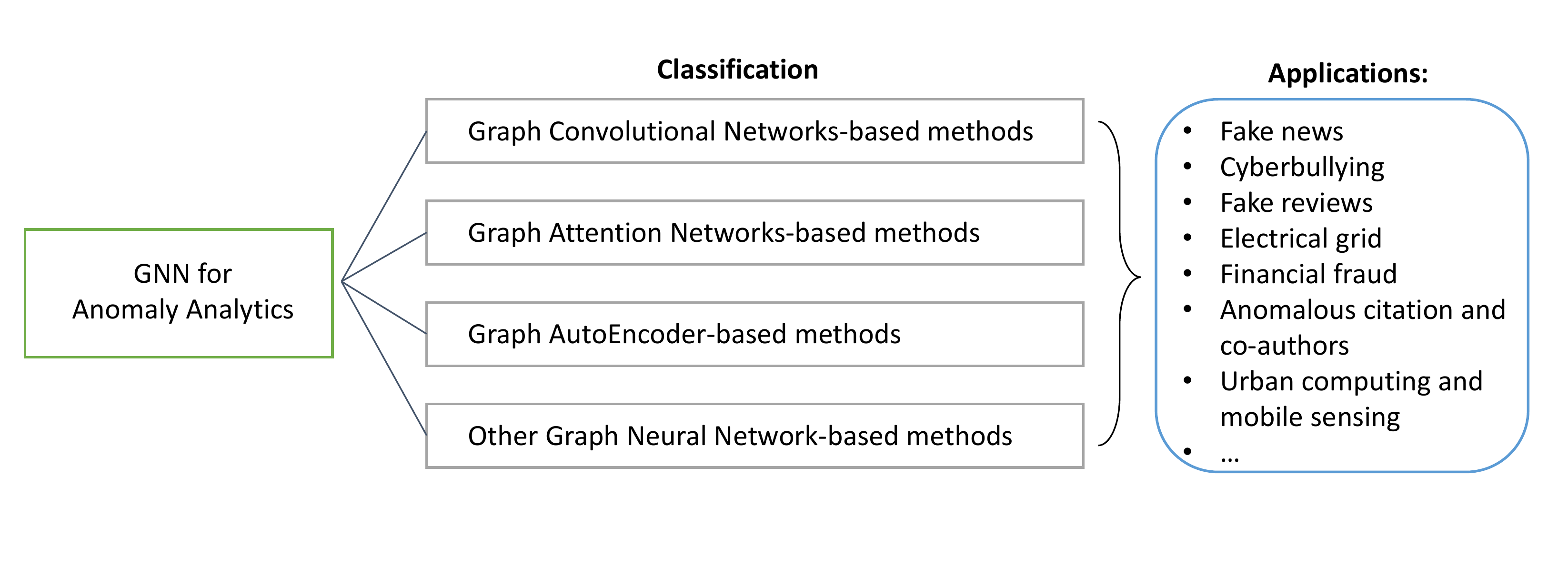}\\
		\caption{Classification of deep graph methods in solving anomaly analytics tasks and real-world applications.}\label{fm}
\end{figure*}
\subsection{Related surveys and novelty}
There are several surveys related to our work. Zamini et al. \cite{zamini2019comprehensive} summarized the anomaly detection techniques in four real-world application scenarios, namely banking, wireless sensor networks, social networks, and healthcare. Akoglu et al.~\cite{akoglu2015graph} reviewed the anomaly detection methods using graph metric-based techniques, Ranshous et al.~\cite{ranshous2015anomaly} only focused on anomaly detection methods in dynamic networks, while Bilgin et al.~\cite{bilgin2006dynamic} briefly reviewed some non-deep learning methods of detecting anomalies in dynamic networks. Both Chalapathy et al.~\cite{chalapathy2019deep} and Pang et al.~\cite{pang2021deep} concentrated on deep learning enabled anomaly detection in different kinds of data, which is not limited to graph data. \cite{ma2021comprehensive} reviewed the contemporary deep learning methods for graph anomaly detection and categorized existing work according to the anomalous graph objects. Actually, there are also some surveys focusing on introducing the main concepts and frameworks of Graph Neural Networks (GNNs) \cite{zhou2020graph, wu2020comprehensive}, and divided the corresponding methods according to the type of GNN models. Inspired by this classification strategy, we also divided the graph learning models in terms of the model type when introducing the specific anomaly detection tasks. However, the main focus of our survey and GNN survey are totally different in spite of the similar classification strategy.
	
This work is different from previous studies in that we aim to summarize the graph learning methods systematically and comprehensively for detecting anomalies in various graphs, ranging from homogeneous to heterogeneous, non-attributed to attributed, undirected to directed, rather than focusing on only one specific kind of graph. To fill this gap, we divide the existing methods into four categories based on their model architectures and training strategies: Graph Convolutional Networks (GCNs), Graph Attention Networks (GATs), Graph AutoEncoders (GAEs), and other GNN-based methods (shown in Figure \ref{fm}). The main characteristics of these methods are compared and summarized in Table~\ref{sum}. The characteristics of these basic models are briefly introduced in Section~\ref{sec2}. In summary, the contributions of this work are outlined as follows:
\begin{itemize}
		\item A systematic summarization and comparison of graph learning methods for anomaly analytics is presented. Specifically, we delineate their capabilities in addressing the existing problem complexities among all categories of the methods.
		\item An overview of major anomaly analysis tasks in various application domains is given.
		\item Insights into future research directions in this field are provided.
\end{itemize}
	
\subsection{Organization}
The rest of this survey is structured as follows. Section~\ref{sec2} presents the notations and preliminaries of graph learning models, which will be used in the subsequent sections. The anomaly analytics methods are reviewed in Sections~\ref{sec3} to \ref{sec6}. In Section~\ref{sec7}, we outline several real-world applications of anomaly analytics that can be solved with deep graph models, and discuss some future research directions and challenges in Section~\ref{sec8}. Finally, we briefly conclude this survey in Section~\ref{sec9}.
	
\begin{figure*}[htb]
		\centering
		\includegraphics[width=0.9\textwidth]{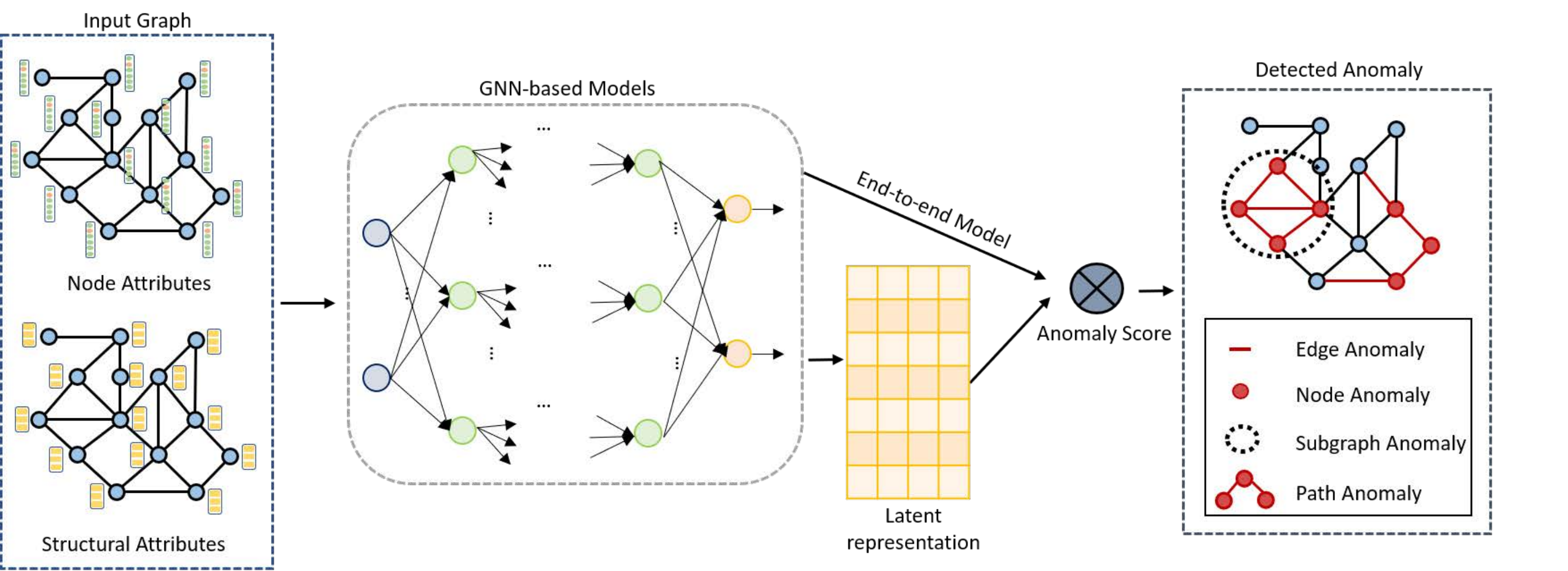}\\
		\caption{Illustration of the whole process of detecting anomalies in graph data with deep graph models. The models are mainly divided into two parts according to whether anomaly score is calculated by latent representation or directly generated by end-to-end models. There are mainly four types of graph anomalies, namely node, edge, (sub)graph, and path anomaly.}\label{anomalytype}
\end{figure*}

\begin{table*}
	\caption{\label{sum}Summary of graph learning models in detecting and predicting anomalies}
	\begin{tabular}{|p{1cm}<{\centering}|p{2.3cm}<{\centering}|p{3cm}<{\centering}|p{1.5cm}<{\centering}|p{4cm}<{\centering}|p{3.5cm}<{\centering}|}
		\hline
		\multirow{2}{*}{\textbf{Type}} &\multirow{2}{*}{ \textbf{Method}}&\multirow{2}{*}{\textbf{Graph Type}}&\textbf{Anomaly Type}&\multirow{2}{*}{\textbf{Dataset}}&\multirow{2}{*}{\textbf{Application}}\\
		\hline
		\multirow{20}{*}{GCN}	
		&GEM~\cite{liu2018heterogeneous}	&HIN &	Node&	-	&Malicious account\\ \cline{2-6}
		&\multirow{2}{*}{GCNSI~\cite{dong2019multiple}}& \multirow{2}{*}{Undirected network}&	\multirow{2}{*}{Node}&	Karate, Dolphin, Jazz, US Power grid, Ego-Facebook\footnotemark[1]&		\multirow{2}{*}{Rumor source detection}\\  \cline{2-6}
		&	GAS~\cite{li2019spam}&	HIN&	Node\&Edge&	- &	Spam review detection\\ \cline{2-6}
		& 	SpecAE~\cite{li2019specae}&	Attributed networks&	Node&	Cora, Pubmed, PolBlog~\cite{perozzi2014focused}	&-\\ \cline{2-6}
		&	\multirow{2}{*}{DOMINANT~\cite{ding2019deep}}&	\multirow{2}{*}{Attributed networks}&	\multirow{2}{*}{Node}&	BlogCatalog \cite{wang2010discovering},Flickr \cite{tang2009relational}, ACM \cite{tang2008arnetminer}&\multirow{2}{*}{-}\\
		\cline{2-6}
		
		&GCNwithMRF \cite{wu2020graph}	&Directed graph&Node& TwitterSH~\cite{lee2011seven}, 1KS-10KN~\cite{yang2012analyzing}&Social spammer detection\\	\cline{2-6}
		
		&Bi-GCN~\cite{bian2020rumor}&	-&	Edge&	Weibo~\cite{ma2016detecting}, Twitter~\cite{ma2017detect} &Rumor detection\\	\cline{2-6}
		
		&GCAN~\cite{lu2020gcan}&Weighted graph&	Node&	Twitter &	Fake news detection\\	\cline{2-6}
		& 	TPC-GCN~\cite{zhong2020integrating}&HIN	&	Node&	Weibo~\cite{ma2016detecting}, Reddit~\cite{hessel2019something} &Controversy detection\\	 \cline{2-6}
		& 	AANE~\cite{duan2020aane}&-	&Edge&	Disney\footnotemark[2], Enron\footnotemark[3] & -\\  \cline{2-6}
		&	HMGNN~\cite{zhu2020heterogeneous}&Vanilla graph	&	Node&	- &Fraud invitation\\	\cline{2-6}
		&		GraphRfi~\cite{zhang2020gcn}	&Bipartite Graph&	Node&	Yelp~\cite{rayana2015collective}, Amazon~\cite{mcauley2013amateurs}&-\\ 	\cline{2-6}
		%		&		\multirow{3}{*}{Zhong et al.~\cite{zhong2019graph}}	&\multirow{3}{*}{Feature Similarity graph}&	\multirow{3}{*}{-}&	UCF-Crime~\cite{sultani2018real}, ShanghaiTech~\cite{luo2017revisit}, UCSD-Peds~\cite{li2013anomaly}&\multirow{3}{*}{Fraudster detection}\\ 	\cline{2-6}
		&		AddGraph~\cite{zheng2019addgraph}	&Dynamic graph&	Edge&UCI~\cite{opsahl2009clustering}, Digg~\cite{de2009social}	&-\\ \cline{2-6}
		&	ST-GCAE~\cite{markovitz2020graph}	&ST Graph&	Graph&ShanghaiTech~\cite{luo2017revisit}	&Anomalous action\\
		\cline{2-6}
		&StrGNN~\cite{cai2021structural}	&Temporal Graph& Node	& UCI, Digg~\cite{de2009social}&-\\	 \cline{2-6}
		&ANEMONE~\cite{jin2021anemone}	&- & Node	& Cora, Citeseer, Pubmed&-\\	 \cline{2-6}
		&CoLA~\cite{liu2021anomaly}	&Attributed network& Node	&  \cite{wang2010discovering}, \cite{tang2009relational}, \cite{tang2008arnetminer}&-\\ \cline{2-6}
		&GCCAD~\cite{chen2022gccad}	&Attributed network& Node	& Aminer, MAS, Alpha, Yelp&-\\
		\hline
		
		\multirow{9}{*}{GAT}	&
		HAGNE~\cite{wang2019heterogeneousgraph}&HINs&Graph&	-	&Unknown malware \\ \cline{2-6}	
		&HACUD \cite{hu2019cash}	&Attributed HINs&	Node& -&Cash-out user detection \\ \cline{2-6}	
		& SemiGNN~\cite{wang2019semi}&Multiview graph	&Node&- &Financial fraud\\	\cline{2-6}
		&AA-HGNN~\cite{ren2020adversarial}&HINs	&	Node&	BuzzFeed\footnotemark[4] &Fake news detection\\		\cline{2-6}
		& 	mHGNN~\cite{fan2020metagraph}&Attributed HINs	&	Subgraph&	-&Illicit traded product \\	\cline{2-6}
		%			&	GraphConsis~\cite{liu2020alleviating}&	HINs	&	Node& Yelp, Pubmed~\cite{sen2008collective}, NEL~\cite{carlson2010toward} &Fraud detection\\		\cline{2-6}
		&	GDN~\cite{deng2021graph}	&Directed graph&	Node& SWaT~\cite{mathur2016swat}, WADI~\cite{ahmed2017wadi}& Anomalous sensors \\ 	\cline{2-6}
		& TGBULLY~\cite{ge2021improving}	&	Temporal graph&	Subgraph&	Instagram~\cite{hosseinmardi2015analyzing}, Vine~\cite{rafiq2015careful}&Cyberbullying detection\\ \cline{2-6}
		& TADDY~\cite{liu2021anomalyTKDE} & Dynamic graph&Node& Email~\footnotemark[5], AS-Topology~\footnotemark[6]&-\\
		\hline
		
		\multirow{8}{*}{GAE}	&		AEHE~\cite{fan2018abnormal}	&HINs&	Path&	ACM~\cite{tang2008arnetminer}	&	Co-authored event \\ \cline{2-6}	
		&AEGIS~\cite{ding2020inductive}&Attributed networks&	Node&BlogCatalog,Flickr, ACM	 &-\\	\cline{2-6}	
		&DONE~\cite{bandyopadhyay2020outlier}&Attributed network&	Node& Cora, Citeseer, Pubmed~\footnotemark[7]&-\\ 	\cline{2-6}	
		&\multirow{2}{*}{DeepSphere~\cite{teng2018deep}}&\multirow{2}{*}{Dynamic networks}&	\multirow{2}{*}{Node}&NYC taxi trip~\footnotemark[8], HERMEVENT~\cite{di2017hermevent}&  	\multirow{2}{*}{-}\\\cline{2-6}	
		&UCD~\cite{cheng2020unsupervised}&Attributed networks&	-&	Instagram~\cite{hosseinmardi2015analyzing}, Vine~\cite{rafiq2015careful}  &Cyberbullying detection\\\cline{2-6}
		&SL-GAD~\cite{cai2021structural}&Attributed network&	Node& Cora, Citeseer, Pubmed&-\\ 		\hline
		
		\multirow{5}{*}{Other}	&		\multirow{2}{*}{GAL~\cite{zhao2020error}}&	\multirow{2}{*}{Bipartite graph}&	\multirow{2}{*}{Node}&	Bitcoin-Alpha~\cite{kumar2016edge}, Tencent-Weibo~\cite{jiang2016catching}&\multirow{2}{*}{-}\\ \cline{2-6}
		&CARE~\cite{dou2020enhancing}&Multi-relation graph&	Node&	Yelp, Amazon &Camouflaged fraudsters\\	\cline{2-6}
		& Meta-GDN~\cite{ding2021few}&	Cross-network&	Node&	 PubMed~\cite{sen2008collective}, Reddit~\cite{hamilton2017inductive}&-\\	\cline{2-6}
		%		& OCGNN~\cite{wang2020one}&	Attributed network&	Node&Cora, Citeseer and Pubmed~\cite{sen2008collective}&	-\\\cline{2-6}
		& GAAN~\cite{chen2020generative}&Attributed network&	Node&BlogCatalog, Flickr, ACM &-\\\cline{2-6}
		& MAHINDER~\cite{zhong2020financial}&	Multi-view HINs&	Node&-&Financial defaulter \\
		\hline
		
	\end{tabular}
\end{table*}		

\footnotetext[1]{\url{http://www-personal.umich.edu/~mejn/netdata/ \& http://snap.stanford.edu/}}
\footnotetext[2]{\url{https://www.ipd.kit.edu/˜muellere/consub/}}
\footnotetext[3]{\url{https://www.cs.cmu.edu/˜./enron/}}
\footnotetext[4]{\url{https://github.com/KaiDMML/FakeNewsNet/tree/old-version}}
\footnotetext[5]{\url{http://networkrepository.com/email-dnc}}
\footnotetext[6]{\url{http://networkrepository.com/tech-as-topology}}
\footnotetext[7]{\url{https://linqs.org/datasets/}}
\footnotetext[8]{\url{https://portal.311.nyc.gov/}}

\section{Notations and Preliminaries}\label{sec2}
\subsection{Notations}
A graph\footnote{Graph and network are used interchangeably in this paper.} is represented as $G = (V,E)$, where $V = \{v_1, ..., v_n\}$ is a set of $n$ nodes and $E \subseteq V \times V$ is a set of $m$ edges between nodes. A graph may have different types, such as weighted or unweighted, directed or undirected. Here, if a graph is directed, then $e_{ij} = (v_i,v_j) \in E$ denotes an edge pointing from $v_i$ to $v_j$. The neighborhood set of a node $v_i$ is defined as $N(v_i) = \{v_j \in V|(v_i,v_j) \in E\}$. The adjacency matrix of a graph is a $n \times n$ matrix, which is denoted as $\mathbf{A}$. We use $\mathbf{A}(i,:), \mathbf{A}(:,j), \mathbf{A}(i,j)$ to denote the $i^{th}$ row, $j^{th}$ column, and an element of $\mathbf{A}$, respectively. For an unweighted graph, the element of its adjacency matrix is defined as:
\begin{equation}
		\textbf{A}(i,j)=\left\{
		\begin{array}{rl}
			1 & \mbox{if $~e_{ij} \in E$} \\
			0 & \mbox{otherwise.}
		\end{array} \right.
\end{equation}
For a weighted graph, $\mathbf{A}(i,j)$ is defined as the weight of edge $e_{ij}$. A graph may have node attributes $\mathbf{X^V}$ and edge attributes $\mathbf{X^E}$, where $\mathbf{X^V}$ is the node attributes matrix and $\mathbf{X^E}$ is the edge attributes matrix, respectively. If the feature matrix is used as $\mathbf{X}$ for convenience, the default setting $\mathbf{X}$ refers to node attributes matrix. Functions are marked with curlicues, e.g., $\mathcal{F}(\cdot)$.
	
Throughout this paper, we use bold uppercase characters denoting matrices and bold lowercase characters representing vectors, like a matrix $\mathbf{A}$ and a vector $\mathbf{a}$. Unless particularly specified, the notations used in this paper are illustrated in Table~\ref{notations}. Then, we provide a formal definition and brief introduction of some predefined matrices to better understand the concepts described in this paper.
\subsection{Preliminaries}
Given an undirected graph, the Laplacian matrix is defined as $\mathbf{L} = \mathbf{D} - \mathbf{A}$, where $\mathbf{D} \in \mathbb{R}^{n \times n} $ is a diagonal degree matrix with $\mathbf{D}_{ii} = \sum_{j=1}^{n}\mathbf{A}_{ij}$.	$\mathbf{L} = \mathbf{Q\Lambda Q^T}$ denotes eigendecomposition, where $\mathbf{\Lambda} \in \mathbb{R}^{n \times n}$ is a diagonal matrix of eigenvalues in ascending order and $\mathbf{Q} \in \mathbb{R}^{n \times n}$  is composed of corresponding eigenvectors. The element $\mathbf{P}(i,j)$ of transition matrix $\mathbf{P} = \mathbf{D}^{-1}\mathbf{A}$ represents the probability of a random walk from node $v_i$ to node $v_j$.
	
As mentioned previously, this survey aims to introduce existing research on graph anomaly detection and prediction. A graph is an abstract data type consisting of a set of nodes (a.k.a. vertices) representing entities, with edges between nodes representing relations or connections. Anomalies in graph data fall into four main categories: node anomaly, edge anomaly, path anomaly, and (sub)graph anomaly. The whole process of detecting anomalies with deep graph models is briefly illustrated in Figure~\ref{anomalytype}.
	
When learning a deep model on graphs for anomaly analytics tasks, we divid the models into four categories based on their model architectures and training strategies. Here, we briefly introduce the process and potential mechanism of these graph neural network models.
\begin{itemize}
		\item \textbf{Graph Convolutional Networks (GCNs)}
		Considering that graphs lack a grid structure like image and text, it is impractical to directly apply standard convolution operation on graphs. Graph convolution is generally divided into two categories, \textit{spectral convolution}, which performs Fourier transform on graph signals, and \textit{spatial convolution}, which learns structural information by aggregating node neighbors~\cite{welling2017semi}. The graph signal $\mathbf{X}$ in spectral methods is filtered by:
\begin{equation}
	\mathbf{Z}=f(\mathbf{X},\mathbf{A})=\widetilde{\mathbf{D}}^{-\frac{1}{2}}\widetilde{\mathbf{A}}\widetilde{\mathbf{D}}^{-\frac{1}{2}}\mathbf{X}\mathbf{\Theta}
\end{equation}
where $\mathbf{\Theta}$ is a matrix of learnable parameters and $\mathbf{Z}$ is the convolved signal matrix. 
		
In addition, the equation of learning node representation of node $i$ in a Graph Convolution Network (GCN) can be written as:
\begin{equation}
	\mathbf{h_i}=\sigma \left( \sum_{j\in N(v_i)}\alpha_{ij}\mathbf{W} \mathbf{h_j} \right).
\end{equation}
where $\mathbf{W}$ is weight matrix to be learned, and $\alpha_{ij}$ is set as 1 in GCN. The calculation formula of $\alpha_{ij}$ will be introduced in GAT.
		
\item \textbf{Graph Attention Networks (GATs)}
	It is acknowledged that spatial convolution is to aggregate features from node neighbors to update the hidden state of this node in the next layer. The aggregation operation could be (weighted) summarization, averaging, and maximization. Graph Attention Network (GAT) is a special type of spatial convolution methods. Although some spatial methods also consider the node importance and allocate a predefined weight for every neighbor, a Graph Attention Network is proposed so that the weight of nodes can be learned automatically by applying the attention mechanism to neighbors in the model~\cite{velivckovic2017graph}. The influence $\alpha_{ij}$ of node $v_j$ on node $v_i$ in GAT is calculated as:
\begin{equation}
			\alpha_{ij}=\frac{exp(\sigma(\mathbf{a}^T[\mathbf{W}\mathbf{h}_i\parallel\mathbf{W}\mathbf{h}_j]))}{\sum_{k\in{N(i)}} exp(\sigma(\mathbf{a}^T[\mathbf{W}\mathbf{h}_i\parallel\mathbf{W}\mathbf{h}_k]))}.
\end{equation}
Here, $\mathbf{a}$ denotes a weight vector and the symbol $\parallel$ is the concatenation operation on two vectors.
\item \textbf{Graph AutoEncoder (GAE)} Graph autoencoder is a popular model used in unsupervised learning tasks~\cite{an2015variational,el2021weighted}. Similar to the general autoencoder model, GAE is composed of an \textit{encoder} compressing the sparse node vector (input) into a low-dimensional representation through learning node structural features, and a \textit{decoder} reconstructing the dense vector into a high-dimensional vector similar to the input as much as possible. Based on this mechanism, an essential part of loss function in GAE models is to minimize the difference between the input and output vectors:
\begin{equation}
	\mathop{min}\limits_{\Theta} \mathcal{L}_2= \parallel \mathbf{A}-\mathbf{\hat{A}} \parallel_2 + \parallel \mathbf{X}-\mathbf{\hat{X}} \parallel_2,
\end{equation}
where $\mathbf{A}$ and $\mathbf{X}$ are the input node adjacency and attribute matrix, and $\mathbf{\hat{A}}$ and $\mathbf{\hat{X}}$ are the reconstructed node structure and attribute matrix.
It should be noted that the encoder could be any kind of neural network like MLP, Recurent Neural Network (RNN), and GCN. Therefore, there are a line of anomaly detection methods combining GAE and GCN. We refer readers to~\cite{zhang2020deep} for more details about deep graph models and their applications.
		
\end{itemize}

\begin{table}[htbp]
		\caption{\label{notations}Commonly used notations}
	\begin{tabular}{ll}
			\toprule
			\textbf{Notations} & \textbf{Descriptions}  \\
			\midrule
			$G = (V, E)$ & A graph.  \\
			$N(v)$ & The neighbors of a node $v$.\\
			$\mathbf{A}$ & The graph adjacency matrix.\\
			$\mathbf{D}$ & The diagonal degree matrix. \\
			$\mathbf{X}$ & The graph feature matrix.\\
			$\mathbf{D}_{ii} = \sum_{j=1}^{n}\mathbf{A}_{ij}$& The degree of node $i$. \\
			$\mathbf{L} = \mathbf{D}-\mathbf{A}$ & The Laplacian matrix.\\
			$\mathbf{U}$ & The eigenvector matrix of Laplacian matrix.\\
			$\mathbf{\Lambda}$ & The eigenvalue matrix of Laplacian matrix.\\
			$\mathbf{A}^T$ & The transpose of the matrix $\mathbf{A}$.\\
			$\mathbf{A}^n$ & The $n^{th}$ power of $\mathbf{A}$.\\
			$\mathbf{H}^{(l)}$ & The hidden representation in the $l^{th}$ layer.\\
			$\mathbf{W}$ & The weight parameter matrix.\\
			$\mathbf{b}$ & The bias parameter vector.\\
			$\mathbf{Z}$ & The generated node embedding matrix.\\
			$\Theta$ & Learnable parameters. \\
			$\parallel$ & The concatenation of two vectors. \\
			$\sigma(\cdot)$ & The sigmoid activation function. \\
			$|\cdot|$ & The length of a set.\\
			\bottomrule
	\end{tabular}
\end{table}
\section{GCN-based Methods}\label{sec3}
\begin{figure*}[htb]
		\centering
		\includegraphics[width=0.9\textwidth]{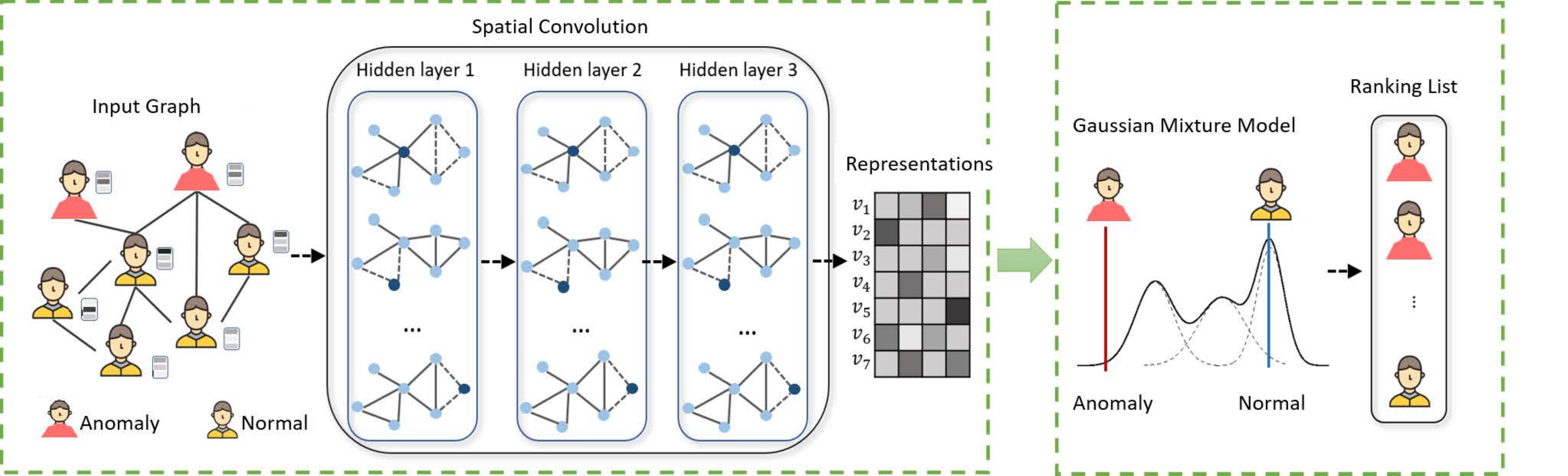}\\
		\caption{An illustration of applying spatial convolution operation in anomalous user detection in social networks, where nodes are affected only by their immediate neighbors. Both attribute feature and structure feature could be learned with a GCN model to get the final anomaly ranking list.}\label{gcn}
\end{figure*}

As the most popular structure among the deep graph models, Graph Convolutional Networks (GCNs) can learn and generate node embeddings through the operation of convolution, which refers to the process of aggregating information from the nodes' local neighborhoods. In this section, we introduce the GCN-based anomaly detection and prediction methods, which is also the most popular model type among all anomaly analytics models. The methods are divided into two classifications according to whether the methods are devised for specific anomaly detection tasks or not, namely, general models and task-driven models. A toy model of how anomalous users are detected in social networks with spatial convolution operation is shown in Figure \ref{gcn}. The main characteristics of these methods are summarized in Table \ref{gcntab}.
\subsection{General models}
%	
%Considering that graph is temporarily incomplete, Zhao et al.~\cite{zhao2020early} proposed ELAND to detect anomalous users at an early stage. It is composed of two major components, namely, a graph anomaly detection module and a behavior forecasting module. In the process of node embedding, the graph convolution operation of each GCN layer is defined as:
%\begin{equation}\label{GCN}
%		\mathbf{H}^{l+1} \leftarrow \sigma(\widetilde{\mathbf{D}}^{-\frac{1}{2}}\widetilde{\mathbf{A}}\widetilde{\mathbf{D}}^{-\frac{1}{2}}\mathbf{H}^{(l)}\mathbf{W}^{(l)}),
%\end{equation}
%where $\widetilde{\mathbf{A}}$=$\mathbf{A}$+$\mathbf{I}$, i.e., adjacency matrix with added self connection, and $\widetilde{\mathbf{D}}$ is the diagonal degree matrix with $\widetilde{D}_{ii}=\sum_{j}\widetilde{A}_{ij}$. Then, the graph anomaly detection model can be written as:
%\begin{equation}
%		\hat{y},\mathbf{Z}=g_{fd-gnn}(\mathbf{A},\mathbf{X};\Theta),
%\end{equation}
%where $\hat{y}$ is the predicted label of the node, and $\mathbf{Z}$ is the node embedding matrix generated by the last layer. Here, binary cross entropy is used as the loss function when training the model.
	
In~\cite{li2019specae}, the authors defined two types of node anomaly according to its global location and topological network structure, named global anomaly and community anomaly, respectively. When learning global anomaly node embeddings, an autoencoder is applied to extract node attributes $X$. As for community anomaly representation, the authors designed a convolutional encoder and deconvolutional decoder networks based on their neighborhoods. Then, the anomalous node could be detected by measuring the embedding's energy in the Gaussian Mixture Model. Similarly, Ding et al.~\cite{ding2019deep} and Zhu et al. \cite{zhu2020deepad} proposed to learn node embeddings by combining AutoEncoder with Graph Convolutional Networks in attributed networks. Specifically, the encoder module extends the operation of convolution in the spectral domain and learns a layer-wise new latent representation. Then, the structure reconstruction decoder $\mathbf{A}-\hat{\mathbf{A}}$ and attribute reconstruction decoder $\mathbf{X}-\hat{\mathbf{X}} $ are jointly learned to compute the anomaly score, which can be formulated as:
\begin{equation}
		score(v_i)=(1-\alpha)\lVert \mathbf{a_i}-\hat{\mathbf{a_i}} \rVert_2 + \alpha \lVert \mathbf{x_i}-\hat{\mathbf{x_i}} \rVert_2,
\end{equation}
where $\alpha$ is a hyper-parameter balancing the importance of reconstructed structure and attribute information.

Instead of detecting anomalous nodes, Duan et al.~\cite{duan2020aane} generated node embeddings by combining GAE and GCN to detect anomalous links. They assumed that a link with a lower value of predicted presence probability is regarded as anomalous, which is calculated as:
\begin{equation}
		\mathbf{P}_{u,v} < \underset{v'\in N_u}{\rm MEAN}\mathbf{P}_{u,v'}-\mu \cdot \underset{v'\in N_u}{\rm STD}\mathbf{P}_{u,v'},
\end{equation}
where MEAN and STD represent the mean and standard operations respectively, and $\mu$ is a parameter. $\mathbf{P}$ is the predicted presence probability matrix.

In \cite{zheng2019addgraph}, the authors aimed to incorporate all possible features in the proposed framework AddGraph, including structural, content, and temporal features. In AddGraph, they used a GCN incorporating content and structural features, with an attention-based GRU (Gated Recurrent Unit), which can combine short-term and long-term states. After obtaining the hidden states of nodes at timestamp $t$, the anomalous score of an edge is computed as:
\begin{equation}
		\mathcal{F}(i,j,w)=w \cdot \sigma(\beta \cdot(\parallel \mathbf{a}\odot \mathbf{h}_i + \mathbf{b}\odot \mathbf{h}_j\parallel_2^2-\mu)),
\end{equation}
where $\mathbf{h}_i$ and $\mathbf{h}_j$ are the hidden states of the $i$-th node and $j$-th node, respectively. Other characteristics are parameters to be adjusted. $\mathbf{a}$ and $\mathbf{b}$ are parameters to optimize the output layer, and $\beta$ and $\mu$ are hyper-parameters.
	
Jin et al. \cite{jin2021anemone} leveraged a multi-scale contrastive learning technique to capture node anomalies in multiple scales. ANEMONE simultaneously performs patch- and context-level contrastive learning via two GCN models. Anomaly is identified by leveraging the statistics of multi-round contrastive scores.
	
Similarly, Liu et al. \cite{liu2021anomaly} also proposed to detect node anomalies in attributed network in a contrastive learning way. The objective of their model CoLA is to discriminate the agreement between the elements within the selected instance pairs, which is finally used to calculate the anomalous scores of nodes. The difference between CoLA and ANEMONE is the process of sampling. CoLA selected the local subgraph including the target node as positive sample, while local graph without target node is negative sample.

Differing from existing graph contrastive learning frameworks for GNN pre-training, Chen et al. \cite{chen2022gccad} performed contrastive learning in a supervised learning way. In other words, the negative samples are constrained to anomalous nodes instead of being constructed according to some rules. Considering that the bottleneck of anomaly detection tasks is the lack of sufficient anomaly labels, they proposed to construct pseudo anomalies via corrupting the original graph.

\subsection{Task-driven models}
	
To detect malicious account at a mobile cashless payment platform, Liu et al.~\cite{liu2018heterogeneous} jointly learned the topology of a heterogeneous graph and the features of local structures of the nodes. Specifically, they constructed "homogeneous connected subgraph" based on an assumption that an edge $(i, i')$ is added if both account $i$ and $i'$ login to the same device in the original heterogeneous graph. This subgraph is composed of only accounts as nodes. The function to learn the embeddings of nodes is defined as:
\begin{equation}
		\mathbf{H}^{(l+1)}\leftarrow \sigma(\mathbf{X} \cdot \mathbf{W} + \sum_{d=1}^{|D|} softmax(\alpha_{d})\cdot \mathbf{A}^{(d)} \cdot \mathbf{H}^{(l)} \cdot \mathbf{V}_d),
\end{equation}
where $|D|$ is the number of subgraphs extracted from the original graph, and $\mathbf{V}_{d}$ is a parameter controlling the shape of the function. Moreover, an attention mechanism is also utilized in the learning process of different types of subgraphs, i.e., $softmax(\alpha_d)=\frac{exp~\alpha_{d}}{\sum_{i}exp~\alpha_{i}}$, and $\alpha=[\alpha_1, ...,\alpha_{|D|}]^T$ is a free parameter to be learned.
	
Based on an assumption that representing nodes with the information of its neighbors will effectively improve the performance of the source node detection task, Dong et al.~\cite{dong2019multiple} designed a model GCNSI to locate multiple rumor sources without prior knowledge of the underlying propagation model. This model learns node embeddings by adopting convolution in the spectral domain, which considers its multi-hop neighbors’ information. The propagation strategy of GCNSI is modified based on LPSI~\cite{wang2017multiple}.
	
Concerning the task of spam review detection, Li et al.~\cite{li2019spam} aimed to capture the local context and global context of a comment. The proposed model GAS simultaneously integrates a heterogeneous bipartite graph and a homogeneous comment graph. The comment edge embedding in bipartite graph is to aggregate the hidden states of three variables in the previous layer, i.e., the edge itself and its linked two nodes:
\begin{equation}
		\mathbf{H}_e^{l}=\sigma(\mathbf{W}_E^l \cdot AGG_E^l(\mathbf{H}_e^{l-1}, \mathbf{H}_{U(e)}^{l-1}, \mathbf{H}_{I(e)}^{l-1})),
\end{equation}
where
\begin{equation}
		AGG_E^l(\mathbf{H}_e^{l-1}, \mathbf{H}_{U(e)}^{l-1}, \mathbf{H}_{I(e)}^{l-1})=\mathbf{H}_e^{l-1} \parallel \mathbf{H}_{U(e)}^{l-1} \parallel \mathbf{H}_{I(e)}^{l-1}.
\end{equation}
Here, $U(e)$ and $I(e)$ are user node set and item node set linked by edge $e$, respectively. Similarly, the user and item embedding are calculated in the same way and the comment embedding in the comment graph can be obtained from a general GCN model. Finally, the classification result $y$ can be calculated according to:
\begin{equation}
		y=classifier(z_i\parallel z_u\parallel z_c \parallel p_c),
\end{equation}
where $z_i$, $z_u$, and $z_c$ are item, user, and comment embeddings obtained from bipartite graph, and $p_c$ is the comment embedding from comment graph, respectively.

With the aim of detecting a rumor on social media, Bian et al.~\cite{bian2020rumor} proposed a top-down GCN (TD-GCN) to model the rumor propagation features, and a bottom-up GCN (BU-GCN) to model the rumor dispersion features, respectively. The node representations are learned over a two-layer GCN:
\begin{equation}\label{TD1}
		\mathbf{H}_1^{TD} \leftarrow \sigma(\hat{\mathbf{A}}^{TD} \mathbf{X}\mathbf{W}_0^{TD}),
\end{equation}
\begin{equation}\label{TD2}
		\mathbf{H}_2^{TD} \leftarrow \sigma(\hat{\mathbf{A}}^{TD} \mathbf{H}_1^{TD} \mathbf{W}_1^{TD}),
\end{equation}
where $\mathbf{H}_1^{TD}$ and $\mathbf{H}_2^{TD}$ refer to the two-layer hidden features of the TD-GCN. The bottom-up features of BU-GCN are calculated in the same manner as Eq.~\ref{TD1} and \ref{TD2}, while the adjacency matrix should be transposed. 

Another similar topic in social media is fake news detection. Lu et al.~\cite{lu2020gcan} aimed to model the interactions among users by creating a propagation graph as a part of the proposed model. The propagation graph $G_i=(U_i,E_i)$ is constructed by the set of users $U_i$ who share or retweet the topic $s_i$, and the edge is weighted by the cosine similarity between the feature vectors of users. Then, the user embeddings will be learned by a GCN model based on this weighted propagation graph. Similarly, Zhong et al.~\cite{zhong2020integrating} created a Topic-Post-Comment graph for target posts in the task of controversy detection, where the nodes represent topic, post, or comment, and the edges refer to the corresponding interactions between two nodes. The node representations are obtained through a two-layer GCN, the same as Eq.~\ref{TD1} and Eq.~\ref{TD2}.
	
As the first application of deep graph model in the task of fraud invitation detection, Zhu et al.~\cite{zhu2020heterogeneous} proposed HMGNN model by dividing the whole network into $|D|$ mini-graphs, which were represented by hypernodes. The hypergraph is generated by adding edges between mini-graphs. Based on this graph, the convolution for hypernodes is defined as:
\begin{equation}
		\mathbf{H}^{l+1} \leftarrow \sigma(\mathbf{X}_h \mathbf{U}^l + \sum_{d\in D}ATT_d(\widetilde{\mathbf{A}}_h^d\mathbf{H}^l\mathbf{W}_d^l+\mathbf{b}_d^l)),
\end{equation}
where $ATT_d$ is the attention mechanism, and $\mathbf{U}^l$ are free parameters to be trained. Here, $\mathbf{H}^0=[\mathbf{X};\mathbf{X}_h^1; ...; \mathbf{X}_h^{|D|}]$ is the initial representation of the whole graph which concatenates the feature matrix of normal- and hyper- nodes.

To detect social spammers in a semi-supervised way, Wu et al.~\cite{wu2020graph} combined Graph Convolutional Networks (GCNs) and Markov Random Field (MRF) on directed social networks. The layer-wise propagation rule of GCN is defined as:
\begin{equation}
	\mathbf{H}^{(l+1)}= \sigma(\mathbf{D}_i^{-1}\mathbf{A}_i\mathbf{H}^{(l)}\mathbf{W}_i^{(l)}+\mathbf{D}_o^{-1}\mathbf{A}_o\mathbf{H}^{(l)}\mathbf{W}_o^{(l)})		+\mathbf{\tilde{D}}_b^{-\frac{1}{2}}\mathbf{\tilde{A}}_b\mathbf{\tilde{D}}_b^{-\frac{1}{2}}\mathbf{H}^{(l)}\mathbf{W}_b^{(l)},
\end{equation}
where $A_i$, $A_o$, and $A_b$ are three types of adjacency matrices constructed according to three different definitions of neighbors. Considering that different characteristics of pairwise nodes can have different influences on social networks, the authors proposed to use MRF modeling for the joint probability distribution of users' identities. The MRF is formulated as a RNN in this paper to perform multi-step inference when computing the posterior distribution.
	
Since deliberately inserting fake feedback will cause the recommender system bias to the malicious users' favor, Zhang et al.~\cite{zhang2020gcn} presented a GCN-based user representation learning framework to perform robust recommendation and fraudster detection in a unified way. Given a weighted bipartite rating graph $G=(U\cup V,E)$, GCN is adopted to capture topological neighborhood information and side information of nodes. The user and item embedding are calculated as:
	\begin{equation}
		\mathbf{z}_u=\sigma(\mathbf{W} \cdot AGG(\mathbf{h}_k,\forall k \in N(u)) + \mathbf{b}),
	\end{equation}
	\begin{equation}
		\mathbf{z}_v=\sigma(\mathbf{W} \cdot AGG(\mathbf{h}_q,\forall q \in N(v)) + \mathbf{b}),
	\end{equation}
where $\mathbf{h}_k$ and $\mathbf{h}_q$ are the neighbor information for each node. Here, the attention mechanism is incorporated into the aggregation function.
	
It is acknowledged that noisy labels will influence the results of anomaly detection algorithms in some degree. Then, instead of directly generating latent representations, Zhong et al.~\cite{zhong2019graph} designed a GCN-based model to correct noisy labels before detecting anomalous actions in videos. Here, the feature similarity graph is constructed with nodes denoting snippets and edges referring to the similarity between two snippets. Another temporal consistency graph module is directly built upon the temporal structure of a video.
	
Pose estimation is the first step of detecting anomalous actions in videos, and the extracted poses can be embedded by deep graph models. In \cite{markovitz2020graph}, the authors proposed spatio-temporal graph convolution block, which is composed of a spatial-attention graph convolution, a temporal convolution, and a batch normalization. The generated latent vector is fed into a cluster layer to obtain a normality score. Here, the normality score is calculated by a Dirichlet Process Mixture Model (DPMM) for evaluating the distribution of proportional data.
	\begin{table}
		\caption{\label{gcntab}A comparison of the GCN-based models}
		\begin{tabular}{|p{1cm}<{\centering}|p{3cm}<{\centering}|p{1.5cm}<{\centering}|p{2.5cm}<{\centering}|p{6cm}<{\centering}|}
				
				\hline
		\textbf{Sec.}&	\textbf{Method}&\textbf{Type}&\textbf{Convolution}&\textbf{Characteristic}\\
				\hline
			\multirow{7}{*}{3.1}			&SpecAE	\cite{li2019specae}&Spectral &-	&AutoEncoder\\		\cline{2-5}
				& DOMINANT \cite{ding2019deep}&Spatial &First-order &	-\\		\cline{2-5}
			&	DeepAD \cite{zhu2020deepad}&Spatial &First-order &-	\\		\cline{2-5}
			&AANE	\cite{duan2020aane}&Spatial &First-order &-	\\	\cline{2-5}
				&AddGraph	\cite{zheng2019addgraph}&Spatial &First-order & GRU + Attention	\\	\cline{2-5}
				&ANEMONE	\cite{jin2021anemone}&Spatial &First-order & Contrastive learning	\\		\cline{2-5}
				&CoLA	\cite{liu2021anomaly}&Spatial &First-order & Contrastive learning	\\	\cline{2-5}
				&GCCAD	\cite{chen2022gccad}&Spatial &First-order & Contrastive learning	\\	
					\hline

			\multirow{11}{*}{3.2}	&	GEM \cite{liu2018heterogeneous}	& Spatial&First-order &Attention mechanism\\ \cline{2-5}
			&	GCNSI \cite{dong2019multiple}& Spectral&First-order &Semi-supervised learning\\	\cline{2-5}
			&GAS	\cite{li2019spam}&Spatial &First-order &	Attention mechanism\\		\cline{2-5}
			&Bi-GCN	\cite{bian2020rumor}&Spectral &Polynomial &	DropEdge\\		\cline{2-5}
			&GCAN	\cite{lu2020gcan}&Spatial &First-order &-	\\		\cline{2-5}
			&TPC-GCN	\cite{zhong2020integrating}&Spatial &First-order &-	\\		\cline{2-5}
				
			&	HMGNN \cite{zhu2020heterogeneous}	&	Spatial&	First-order &	Adversarial learning +Attention\\		\cline{2-5}
				
			&GCNwithMRF	\cite{wu2020graph}&Spatial &First-order &-	\\		\cline{2-5}
			&GraphRfi	\cite{zhang2020gcn}&Spatial &First-order &	Attention mechanism\\		\cline{2-5}
			&TSN	\cite{zhong2019graph}&Spatial &First-order & -	\\		\cline{2-5}
				
			&ST-GCAE	\cite{markovitz2020graph}&Spatial &First-order &Attention mechanism	\\		\hline
			\end{tabular}
	\end{table}
\subsection{Discussion}
As it can be found from the GCN-based anomaly detection models we have discussed above, the modern GCN model could learn both local and global structure features of a graph with convolution and pooling operations. To improve the training efficiency when imposing GCN on large-scale graphs, neighborhood samplings and layer-wise samplings are two common strategies to deal with the phenomenon that some nodes have high degrees (too many neighbors). In addition to node and edge anomaly, a characteristic of GCN is that it is more suitable to detect (sub)graph or group anomaly when compared with other GNN models.
	
The aforementioned models mostly focus on learning node features and graph structures, ignoring another important element consisting of a graph, i.e., edge. In some real-world networks, edges generally contain abundant information, such as edge types and corresponding attributes, which could play a key role in anomaly detection tasks. Therefore, incorporating edge features into graph anomaly detection models could be considered as a future work~\cite{chen2018supervised}. Besides, although applying GCN to an inductive setting is verified~\cite{hamilton2017inductive}, conducting inductive GCN for graphs without explicit features remains an open problem.

\section{GAT-based Methods}\label{sec4}
\begin{figure*}[htb]
	\centering
	\includegraphics[width=0.9\textwidth]{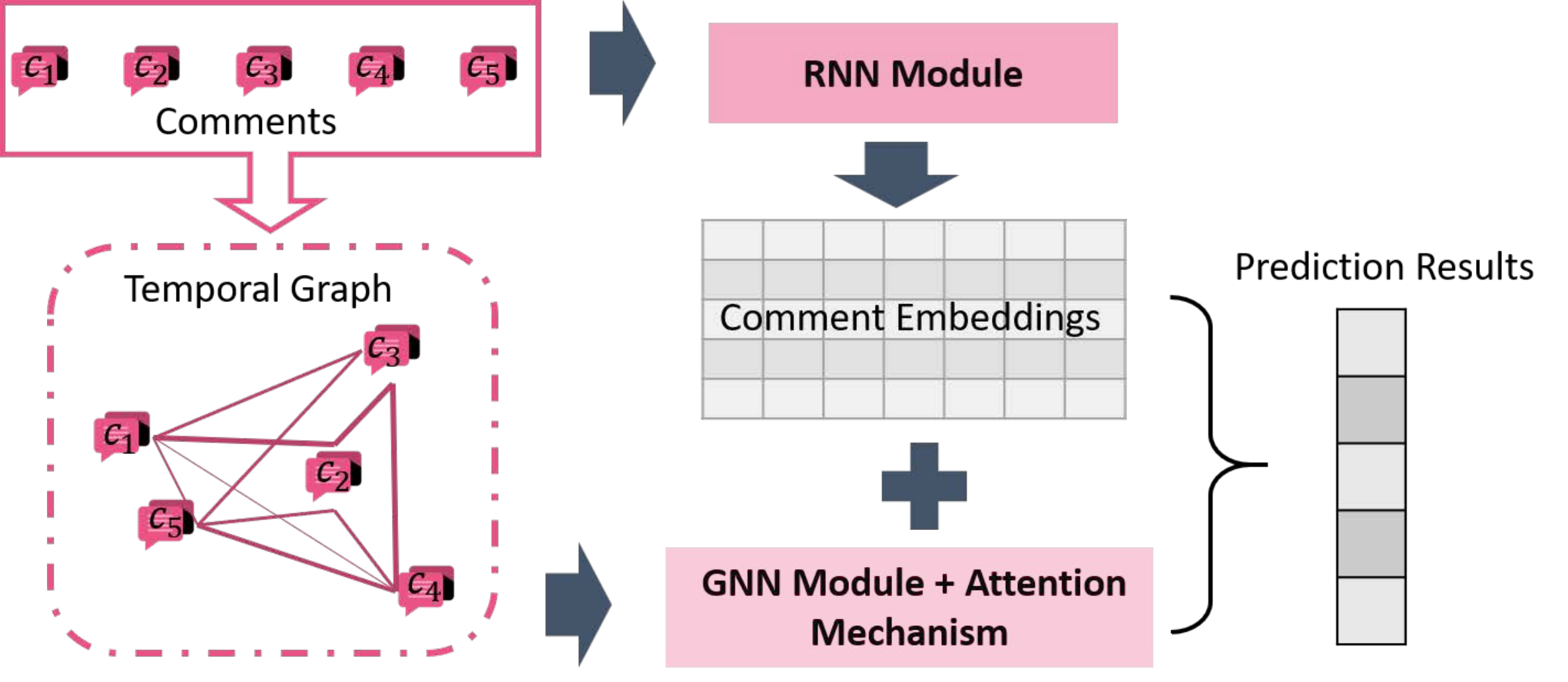}\\
	\caption{An illustration of how attention mechanism is applied into cyberbullying detection. Each comment is first encoded by a RNN framework as the initial vector, and the comments are constructed as a temporal graph where nodes represent user comments and edges represent time intervals between two comments. Then, the attention mechanism is applied to learn the temporal information among these comments for final anomaly detection. }\label{gat}
\end{figure*}

In deep graph models, the weights of node neighbors are defined as an equal or default setting. However, the importance of neighbors is mostly different in terms of their attribute and structural features. Motivated by the attention mechanism, Velivckovic et al.~\cite{velivckovic2017graph} proposed a graph attention network (GAT) by applying the attention mechanism to the spatial convolution operation of GCN. A toy example of how attention mechanism is applied into cyberbullying detection is shown in Figure~\ref{gat}. In this section, we summarize and introduce the anomaly analytics algorithms using graph attention networks. The methods are divided into 2 subsections in terms of the anomaly type, i.e., node anomaly detection and (sub)graph anomaly detection. The main characteristics of these methods are summarized in Table \ref{gattab}.

\subsection{Node anomaly detection}
To detect anomalous nodes in an Attributed Heterogeneous Information Network (AHIN), Hu et al.~\cite{hu2019cash}  applied feature and path attention mechanism to differentiate the importance of meta-paths as well as attribute information. As a basic analysis tool for heterogeneous graph, a meta-path captures the proximity among multiple nodes from a specific semantic perspective, which could be seen as a high-order structure. For example, the meta path ``Author-Paper-Author” (APA) describes that two authors collaborated with each other in a particular paper. The feature attention of neighbor node $i$ on node $u$ in path $\rho$ is calculated as:
\begin{equation}
	\hat{\alpha}_{u,i}^{\rho}=\frac{exp(\alpha_{u,i}^{\rho})}{\sum_{j=1}^{K}exp(\alpha_{u,j}^{\rho})}.
\end{equation}
The attention weight of path $\rho$ for node $u$ is defined as:
\begin{equation}
	\beta_{u,\rho}=\frac{exp(z^{\rho^{T}}\cdot \tilde{f}_u^C)}{\sum_{\rho' \in \mathbb{P}}exp(z^{\rho'^{T}}\cdot \tilde{f}_u^C)}.
\end{equation}
Here, $z^{\rho}$ is the attention vector of meta-path $\rho$, and $\tilde{f}_u^C$ is the collection of user representations w.r.t. all meta-paths. The cash-out probability (i.e., anomalous score) is calculated via a regression layer with a sigmoid unit.
\begin{table*}
	\caption{\label{gattab}A comparison among different GAT-based models}
	\begin{tabular}{|p{2.5cm}<{\centering}|p{2.5cm}<{\centering}|p{4cm}<{\centering}|p{5.5cm}<{\centering}|}
			
			\hline
			\textbf{Method}&\textbf{Attention Level}&\textbf{Objective Function}&\textbf{Other Characteristics}\\
			\hline
			HAGNE~\cite{wang2019heterogeneousgraph}&node \& path& Error sum of squares&Siamese network for graph matching\\ \hline	
			HACUD \cite{hu2019cash}	&feature \& path & Maximum likelihood estimation&Hierarchical attention mechanism\\ \hline
			mHGNN~\cite{fan2020metagraph}&node	& Cross-entropy&Metagraph-guided neighbor search\\	\hline	
			SemiGNN~\cite{wang2019semi}& node \& view		& Cross-entropy+Structure similarity&Semi-supervised learning\\	\hline
			AA-HGNN~\cite{ren2020adversarial}&node \& schema	 &Cross-entropy	&  Adversarial active learning\\		\hline
%			GraphConsis~\cite{liu2020alleviating}&		node	& Cross-entropy&-\\		\hline
			GDN~\cite{deng2021graph}	& node	& Mean squared error&Graph deviation scoring\\ 	\hline
			TGBULLY~\cite{ge2021improving}	&node		&- &Temporal graph interaction learning\\		\hline
			MAHINDER~\cite{zhong2020financial} & node \& path &Cross-entropy&-\\ \hline
			
	\end{tabular}
\end{table*}

To detect user fraud in financial networks, Wang et al.~\cite{wang2019semi} designed a hierarchical attention structure in graph neural network from node-level attention to view-level attention when generating graph embeddings.
View-attention mechanism is applied to fuse multiple views of data information into user embeddings. Finally, a softmax function is used on the representations of the embedding layer to get the classification result.

In~\cite{ren2020adversarial}, the authors constructed a news-oriented heterogeneous information network with nodes of creators, subjects, and articles, and two links of write and belong-to. Based on this network, they proposed AA-HGNN to solve the problem of fake news detection. From the perspective of node-level attention, the model first aggregates the importance of the same-type neighbors for each news node and generates an integrated embedding of schema node. By using a transformation matrix, the embeddings of the nodes can be mapped into the same dimension. The logistic regression layer works as the classification layer to generate the detection results.

%When using graph learning models to detect fraudsters in different kinds of networks, Liu et al.~\cite{liu2020alleviating} proposed three types of inconsistency according to the network structures and node attributes, namely, context, feature, and relation inconsistency. To deal with the relation inconsistency, the authors designed a self-attention mechanism to assign weights for the sampled neighbors. 
	
To give direct explanations like how anomalies deviate from normal behaviors,~\cite{deng2021graph} proposed to use graph attention mechanism to predict the future behavior of a node. The anomaly score of node $i$ is defined as the difference between the expected behavior and observed behavior at time $t$:
	\begin{equation}
		\mathbf{Err}_i(t)=|\mathbf{s}_i^{(t)}-\hat{\mathbf{s}}_i^{(t)}|.
	\end{equation}
For session-level cyberbullying detection, the final embedding is fed into a single-layer dense network and predict its label.

In financial default user detection over online credit payment service, Zhong et al.~\cite{zhong2020financial} devised a meta-path-based encoder to capture local structural feature of nodes and links. The path representation is defined as the concatenation of node and link embeddings. Moreover, attention mechanism is applied to capture different importance of nodes/links of a path. After modeling the node and link interactions above, the learned representation is fed into several fully connected neural networks and a regression layer with a sigmoid unit for anomaly classification.

\subsection{(Sub)graph anomaly detection}
Wang et al.~\cite{wang2019heterogeneousgraph} proposed HAGNE to detect unknown malicious programs in computer systems. Instead of setting one-hop nodes as neighbors, the authors construct a contextual neighborhood set by searching for meta-paths. Then, three kinds of aggregators are applied to generate graph embeddings based on  the generated meta-path set $\mathbb{M}=\{M_1,M_2,...,M_{|\mathbb{M}|}\}$, namely, node-wise attentional neural aggregator, which is defined as:
\begin{equation}
	\mathbf{h}_v^{(i)(k)}= AGG_{node}(\mathbf{h}_v^{(i)(k-1)},\{\mathbf{h}_u^{(i)(k-1)}\}_{u\in \mathcal{N}_v^{i}}),
\end{equation}
where $i\in \{1,2,...,|\mathbb{M}|\}$, $k\in\{1,2,...,K\}$ denotes the layer index, and $\mathbf{h}_v^{(i)(k)}$ is the feature vector of node $v$ at the $k$-th layer in meta-path $M_i$; layer-wise dense-connected neural aggregator, which is inspired by DENSENET~\cite{huang2017densely}:
\begin{equation}
	\mathbf{h}_v^{(i)(K+1)}=AGG_{layer}(\mathbf{h}_v^{(0)},\mathbf{h}_v^{(1)},...,\mathbf{h}_v^{(K)});
\end{equation}
and path-wise attentional neural aggregator, whose attentional weight is defined as:
\begin{equation}
	\alpha_{ij}=\frac{exp(\sigma(\mathbf{b}[\mathbf{W}_b \mathbf{h}_v^{(i)(K+1)} \parallel \mathbf{W}_b \mathbf{h}_v^{(j)(K+1)}]))}{\sum_{j'\in{|\mathbb{M}|}} exp(\sigma(\mathbf{b}[\mathbf{W}_b \mathbf{h}_v^{(i)(K+1)} \parallel \mathbf{W}_b \mathbf{h}_v^{(j')(K+1)}]))},
\end{equation}
Then, the graph embedding can be calculated from the joint representation of all meta-paths:
\begin{equation}
	\mathbf{h}_G=AGG_{path}=\sum_{i=1}^{|M|}ATT(\mathbf{h}_v^{(i)(K+1)})\mathbf{h}_v^{(i)(K+1)}.
\end{equation}
Graph matching is used to measure the anomalous level of a program~\cite{ren2021matching}. An alert will be triggered if the highest similarity score among all the existing programs is below the threshold. The similarity score is calculated as:
\begin{equation}
	Sim(G_{i(1)},G_{i(2)})=\frac{\mathbf{h}_{G_{i(1)}} \cdot \mathbf{h}_{G_{i(2)}}}{\parallel \mathbf{h}_{G_{i(1)}} \parallel \cdot \parallel \mathbf{h}_{G_{i(2)}} \parallel}.
\end{equation}

Subsequently, Fan et al.~\cite{fan2020metagraph} identified the illicit traded product in underground market with a similar process. After constructing the neighbors set by the meta path-based method, the authors applied an attention mechanism to learn product and buyer embeddings, respectively. Finally, their embeddings are generated by concatenating each embedding based on a specific metagraph.

Social media contains multi-modal information such as comment, user, time, and history. Ge et al.~\cite{ge2021improving} proposed to use temporal graph interaction learning module as a building block to detect cyberbullying in social networks. In this work, the authors incorporated GATs to automatically aggregate information from neighbor nodes to the central node in a temporal graph. Edge in the temporal graph denotes time dynamics, and the weight of the node pair $(i,j)$ is defined as:
\begin{equation}
	\alpha(\mathbf{z}_i,\mathbf{z}_j,t_{i},t_{j})=tanh((\mathbf{W}_o\mathbf{z}_i)^{T}\mathbf{z}_j+W_t(t_j-t_i)).
\end{equation}

\subsection{Discussion}
As we have explained, GAT is a branch of GCN. To improve GCN, GAT-based methods are separated as a unique section in which the importance of different neighbors on the central node is considered. The difference between these two sections is that the utilized traditional attention mechanism of Section~\ref{sec4} is only applied on nodes, while models in Section~\ref{sec3} are either using attention mechanism on other parts of the framework or using simple GCN function without any attention mechanism. 
	
\section{GAE-based Methods}\label{sec5}
	
\begin{figure}[htb]
	\centering
	\includegraphics[width=0.8\textwidth]{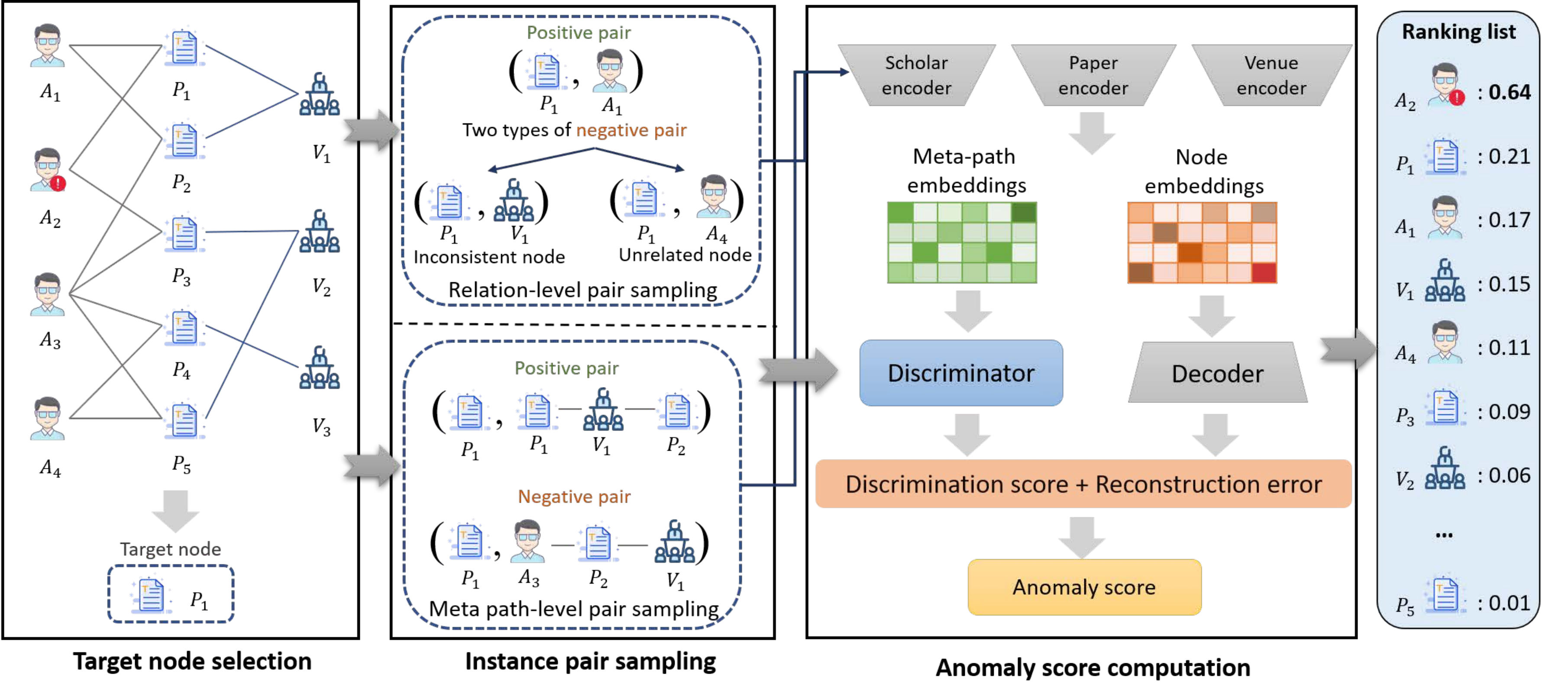}\\
	\caption{An illustration of combining graph autoencoder (GAE) with contrastive learning for anomalous academic’s detection in heterogeneous networks. After selecting a target node, the second step is to sample positive and negative instances from the network for contrastive learning. Then, different kinds of nodes are encoded with different encoders and a common decoder. The encoder aims to learn structure and attribute feature of nodes and generate low-dimensional vectors, and the decoder aims to reconstruct the input vector as similar as possible. The anomaly score is calculated by combining the discrimination score generated by the discriminator and the reconstruction loss generated by the AutoEncoder. }\label{gae}
\end{figure}

Graph AutoEncoder (GAE) is an unsupervised structure to generate low-dimensional representations, with the aim of minimizing the loss between the input of encoder and the output of decoder~\cite{tian2014learning}. In this section, we present the GAE-based algorithms that are applied to anomaly analytics. The methods are classified into three types according to the training and learning schema, namely, General AutoEncoder, Adversarial Training, and Hypersphere Learning. The main characteristics of these methods are summarized in Table \ref{gaetab}. In figure \ref{gae}, we present a GAE-based model for detecting anomalous citation behaviors in a heterogeneous network.
\subsection{General AutoEncoder}
Only considering the structure of a heterogeneous network is not sufficient for abnormal event detection due to the sparsity of a network. Fan et al.~\cite{fan2018abnormal} proposed AEHE to learn both attribute embedding and the second-order structure-preserving node embedding. The heterogeneous attribute embedding of a node is generated by a Multilayer Perceptron (MLP) component, which consists of two hidden layers with ReLU as the activation function. As for the second-order structure embedding, the authors constructed a homogeneous graph by extracting symmetry meta-paths. Autoencoder is used to model the neighborhood structures, which is composed by an encoder:
	\begin{equation}
		\mathbf{r}_i^t=\sigma(\mathbf{W}_1^t \cdot \mathbf{s}_i^t+\mathbf{b}_1^t),
	\end{equation}
and a decoder:
\begin{equation}
	\mathbf{\hat{s}}_i^t= \sigma(\mathbf{\hat{W}}_1^t \cdot \mathbf{r}_i^t+\mathbf{\hat{b}}_1^t),
\end{equation}
where $\mathbf{r}_i^t$ is the latent representation of entity $a_i^t$, and $\mathbf{\hat{s}}_i^t$ is the reconstructed representation of $\mathbf{s}_i^t$. It should be noted that $\mathbf{s}_i^t$ is the $i$th row of the adjacency matrix, not just a node feature vector.
	\begin{table*}
		\caption{\label{gaetab}A comparison among different GAE-based models}
		\begin{tabular}{|p{2.3cm}<{\centering}|p{3cm}<{\centering}|p{5cm}<{\centering}|p{3.5cm}<{\centering}|}
			
			\hline
			\textbf{Method}&\textbf{Type}&\textbf{Objective Function}&\textbf{Other Characteristics}\\
			\hline
			AEHE~\cite{fan2018abnormal}&GAE & Cross-entropy+L2-reconstruction& Negative sampling\\ \hline	
			AEGIS~\cite{ding2020inductive}	&GAE+GAN &Cross-entropy&GDN Encoder\\ \hline
			DONE~\cite{bandyopadhyay2020outlier}&GAE+Discriminator	& L2-reconstruction+Cross entropy&Adversarial learning\\	\hline	
			DeepSphere~\cite{teng2018deep}& GAE+LSTM		&L2-reconstruction &Hypersphere learning\\	\hline
			UCD~\cite{cheng2020unsupervised}& GAE+GCN	 &L2-reconstruction	&  Attention mechanism\\		\hline
			
		\end{tabular}
	\end{table*}
Research have shown that human behaviors reflect self-selection bias and peer influence in online social network, which is closely associated with cyberbullying behaviors~\cite{festl2013social}. In this regard, Cheng et al.~\cite{cheng2020unsupervised} used a GCN encoder and an inner product decoder to learn a latent matrix $\mathbf{Z}$ by minimizing the following reconstruction error:
\begin{equation}
	\mathcal{F}(v_i)=\parallel \mathbf{A}-\mathbf{\hat{A}}\parallel_2^2,
\end{equation}
where $\mathbf{\hat{A}}=\sigma(\mathbf{Z}\mathbf{Z}^T)$, and $\mathbf{Z}=GCN(\mathbf{X},\mathbf{A})$. Then, the anomalous session could be detected by measuring the embedding's energy in the Gaussian Mixture Model, which follows \cite{li2019specae}.
	
\subsection{Adversarial training}
Adversarial methods such as GAN and adversarial attacks are popular in the machine learning community in recent years. In \cite{pan2018adversarially}, the authors incorporated an adversarial training scheme into GAEs as an additional regularization term. Motivated by this work, Ding et al.~\cite{ding2020inductive} proposed AEGIS to learn anomaly-aware node representations through graph differentiation networks (GDNs) for inductive anomaly detection. AEGIS is composed of a GAE to learn node embeddings for training new networks, and a GAN to calculate the anomaly scores of nodes. The autoencoder network is built with the graph differentiative layers. Specifically, a GDN layer has a hierarchical attention structure from node level:
\begin{equation}
	\mathbf{h}_i^{(l)}=\sigma(\mathbf{W}_1\mathbf{h}_i^{(l-1)}+\sum_{j\in N_i}\alpha_{ij}\mathbf{W}_2\Delta_{i,j}^{(l-1)}),
\end{equation}
where $\Delta_{i,j}$ denotes the embedding difference between nodes $i$ and $j$; and neighbor level:
\begin{equation}
	\mathbf{h}_i^l=\sum_{k=1}^{K}\beta_{i}^k\mathbf{h}_i^{(l,k)}.
\end{equation}
Finally, the anomaly score of node $i$ is computed according to the output of a discriminator:
\begin{equation}
	score(\mathbf{z}_i)=1-D(\mathbf{z}_i^{'}).
\end{equation}

In real-world networks, community outliers deviate significantly from other nodes in the same community in terms of link structures and attributes. To alleviate the influence of these outliers and generate robust node embeddings, Bandyopadhyay et al.~\cite{bandyopadhyay2020outlier} mapped every vertex to a low-dimensional vector and detected outliers via a deep autoencoder-based architecture. Moreover, the authors introduced adversarial learning for outlier resistant network embedding. Here, a discriminator is combined with two parallel autoencoders to align the embeddings in terms of link structure and node attributes.
	%\begin{figure}[htb]
	%	\centering
	%	\includegraphics[width=0.45\textwidth]{gdl}\\
	%	\caption{An illustration of graph differentiative layer proposed in~\cite{ding2020inductive}, where $\alpha_{ij}$ is the attention coefficient between node $i$ and node $j$, and $\beta_{k}$ denotes the attention coefficient on $k$th-order representation }\label{gdl}
	%\end{figure}

\subsection{Hypersphere learning}
	
Inspired by hypersphere learning, Wang et al.~\cite{wang2021one} proposed One-Class Graph Neural Network (OCGNN) with the aim of minimizing the volume of a hypersphere that encloses normal nodes as much as possible. Then, the nodes out of the hypersphere are regarded as abnormal.
	
With the aim of identifying anomalous sample cases and nested anomalies within the anomalous tensors, Teng et al.~\cite{teng2018deep} proposed DeepSphere by incorporating hypersphere learning into a LSTM Autoencoder model in a mutual supportive manner. Here, attention mechanism is also applied to differentiate and aggregate different neighbors. The motivation of DeepSphere is that the learned representations at large distance from the outside of hypersphere are regarded as anomalous, while the ones with small distances from the inside of the hypersphere tend to be normal.

\subsection{Discussion}
GAE is the most popular model in tackling unsupervised graph learning tasks, which can only consider the structural patterns by using graph adjacency matrix. However, GCN-based models are semi-supervised and could capture both node attributes and graph structures. Despite the different architectures between GAE and GCN, existing research have shown that it is possible to combine them together in a unified framework~\cite{bojchevski2018deep}. When applying GCN as the encoder, GAE could be applied to the inductive learning settings where node attributes are incorporated. Considering that the aim of GAE is to reconstruct the input embedding as similar as possible, it should be cautious when selecting the appropriate similarity metrics which have significant influence on subsequent anomaly detection results.

\begin{figure*}[htb]
	\centering
	\includegraphics[width=0.9\textwidth]{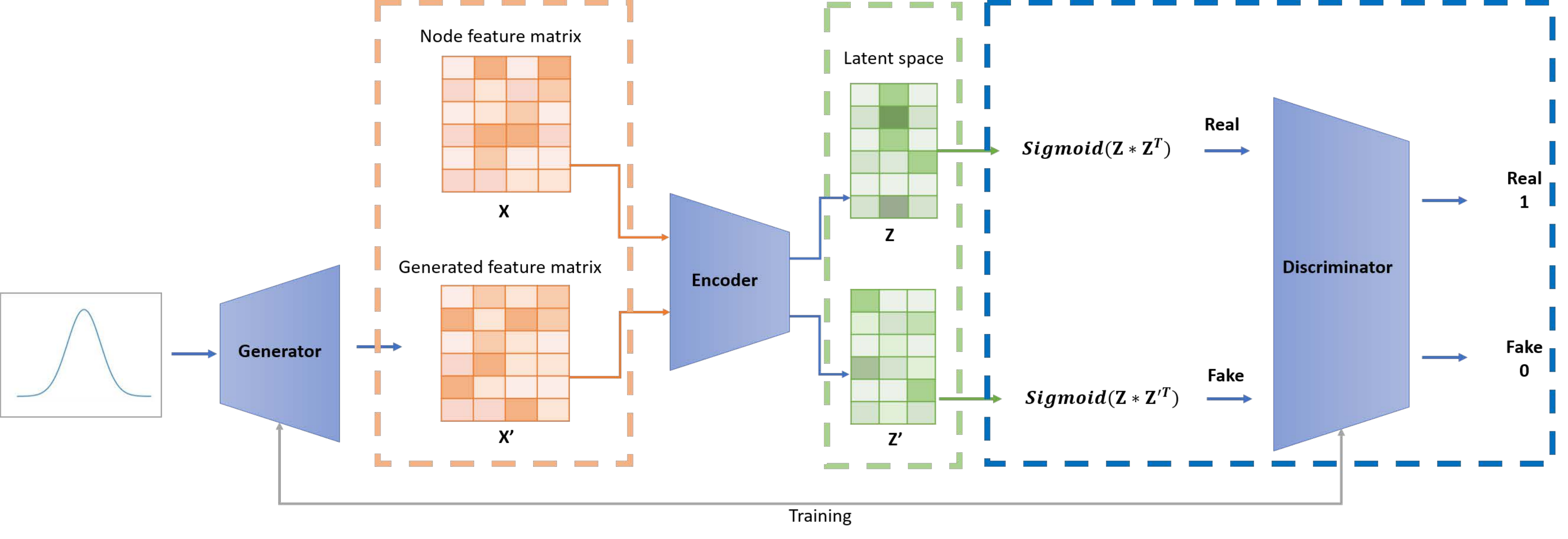}\\
	\caption{The Generative Adversarial Network (GAN)-based anomaly detection model is composed of three main parts:  a Generator sampling similar node attributes, an Encoder generating low-dimensional node representations, and a Discriminator differentiating real nodes embeddings from generated nodes embeddings.}\label{gan}
\end{figure*}
	
\section{Other Methods}\label{sec6}
	
Apart from the deep graph models mentioned above, there are many other popular deep learning models that can be used for anomaly analytics tasks, such as Generative Adversarial Methods~\cite{wang2019learning}, Meta-learning~\cite{zhou2019meta}, and Graph Reinforcement Learning~\cite{do2019graph}. In this section, we summarize these different deep graph models that are utilized to solve anomaly analytics tasks. The process of detecting anomalies with a Generative Adversarial Network (GAN) is shown in Figure~\ref{gan}. The main characteristics of these methods are summarized in Table \ref{gantab}.
	
\subsection{GAN-based methods}
With the rapid growth of research in Generative Adversarial Network (GAN) for high-dimensional data distribution approximation, Chen et al.~\cite{chen2020generative} proposed to detect anomalies with a Generative Adversarial Attributed Network (GAAN), which is composed of three parts: a Generator sampling similar node attributes, an Encoder generating low-dimensional node representations, and a Discriminator differentiating real nodes embeddings from generated nodes embeddings. The anomalous score is defined based on a context reconstruction loss $L_G$ and a structure discriminator loss $L_D$:
\begin{equation}
	\mathcal{F}(v_i)=\alpha \mathcal{L}_G(v_i) + (1-\alpha)\mathcal{L}_D(v_i),
\end{equation}
where $\mathcal{L}_G(v_i)=\parallel \mathbf{x}_i-\mathbf{x}_i' \parallel_2$, and $\mathcal{L}_D(v_i)$ is defined as:
\begin{equation}
	\mathcal{L}_D(v_i)=\sum_{j=1}^{n}\mathbf{A}_{ij}\cdot \sigma(\mathbf{\hat{A}}_{ij},1) / \sum_{j=1}^{n}\mathbf{A}_{ij}.
\end{equation}
Larger value of $\mathcal{F}(v_i)$ indicates the node $v_i$ is more likely to be anomalous.
\subsection{Reinforcement learning-based method}
In \cite{zhao2020error}, the authors divided the graph anomaly detection tasks into two classifications, i.e., outlier detection and unexpected dense block detection. When applying graph learning models to generate embeddings of nodes, a new loss function was designed as:
\begin{equation}
	\mathcal{L}(\mathbf{u})=\mathbb{E}_{u_+\sim U_{u_+},u_-\sim U_{u_-}}max\{0,g(u, u_-)-g(u,u_+)+\bigtriangleup_{y_u}\},
\end{equation}
where $\bigtriangleup_{y_u}=\frac{C}{n_{y_u}^{1/4}}$. $g()$ is a function to denote the similarity of the representations between any two user nodes. Here, $U_{u_+}$ denotes the set of user nodes that has the same label as $u$,  $U_{u_-}$ refers to $ U\backslash U_{u_+}$, and $n_{y_u}=|U_{u_+}|$. The construction of sets $U_{u_+}$ and $U_{u_-}$ have different strategies for corresponding tasks. 
	\begin{table}
		\caption{\label{gantab}A comparison among other deep graph models}
		\begin{tabular}{|p{2.5cm}<{\centering}|p{12cm}<{\centering}|}
			
			\hline
			\textbf{Method}&\textbf{Method/Problem Innovation}\\
			\hline
			GAAN~\cite{chen2020generative}& Generative adversarial network	\\		\hline
			GAL~\cite{zhao2020error}& New loss function\\ \hline	
			CARE-GNN~\cite{dou2020enhancing}	&Reinforcement learning\\ \hline
			Meta-GDN~\cite{ding2021few}&Cross-network meta-learning\\	\hline	
			OCGNN~\cite{wang2021one}& 	 One-Class graph neural network\\	\hline
			
			MAHINDER~\cite{zhong2020financial}& Financial default
			user detection over online credit payment service	\\		\hline
		\end{tabular}
	\end{table}
It is impractical to exactly detect camouflaged fraudsters with graph learning detectors. Dou et al.~\cite{dou2020enhancing} proposed three neural models to enhance the deep graph models against two kinds of camouflages, i.e. feature camouflage and relationship camouflage. Because camouflaged nodes should be filtered when selecting similarity-aware neighbors, a reinforcement learning process is formulated as a Bernoulli Multiarmed Bandit (BMAB) to find the optimal thresholds. The reward mechanism of epoch $e$ is defined as:
	\begin{equation}
		\mathcal{R}(p_r^{(l)},a_r^{(l)})^{(e)}=\left\{
		\begin{aligned}
			+1,& G(\mathcal{D}_r^{(l)})^{(e-1)}-G(\mathcal{D}_r^{(l)})^{(e)} \geq 0, \\
			-1,& G(\mathcal{D}_r^{(l)})^{(e-1)}-G(\mathcal{D}_r^{(l)})^{(e)} < 0.
		\end{aligned}
		\right.
	\end{equation}
Here, $G(\mathcal{D}_r^{(l)})^{(e)}$ refers to the average neighbor distances for relationship $r$ at the $l$-th layer for epoch $e$. If the average distance of newly selected neighbors at epoch $e$ is less than that of epoch $e-1$, then the reward is positive.

	\subsection{Few-shot learning-based method}
	To investigate the novel problem of few-shot network anomaly detection under the cross-network setting, Ding et al.~\cite{ding2021few} designed a new graph learning architecture, namely Graph Deviation Networks (GDNs). GDN in this paper is composed of three key building blocks: an encoder to generate node embeddings, an abnormality valuator to compute the anomaly score of nodes, and a deviation loss for optimization. Concretely, the GDN model can be formally represented as:
	\begin{equation}
		\mathcal{F}_{\theta}(\mathbf{A},\mathbf{X})=\mathcal{F}_{\theta_s}(\mathcal{F}_{\theta_e}(\mathbf{A},\mathbf{X})),
	\end{equation}
	which directly maps the input network to anomaly scores (scalar). After detecting anomalies in arbitrary networks, a meta-learner is learned to initialize GDN from multiple auxiliary networks, which possesses the ability to distill comprehensive knowledge of anomalies.

	\subsection{Discussion}
	There are also many other neural network structures and learning strategies being applied to detect graph anomalies, such as adversarial learning and reinforcement learning. Considering that the number of these methods is relatively less, we summarize all these methods into one section instead of different sections. Other aspects of common deep graph learning models include but not limited to graph reinforcement learning and graph adversarial learning. 
	
	It is well-known that the advantage of reinforcement learning is to actively learn from the feedback. In graph anomaly detection tasks, reinforcement learning could help in optimal selection of neighbors and aggregating them together for more informative node embeddings. Adversarial methods have shown its capacity in generating realistic entities, which improve the detection performance of anomalies that are hardly reconstructed from the latent space. However, this kind of anomaly detection methods faces multiple problems during the training process, such as failure to converge and mode collapse.

	\section{Applications}~\label{sec7}
	Thus far, we have reviewed different graph learning methods in anomaly analytics tasks. In this section, we briefly introduce their applications in different kinds of networks.
	\subsection{Fake news}
	With the rapid growth of the Internet, social media provides a platform for people to participate and discuss online news more conveniently, like communicating news without the physical distance barrier among individuals and acquiring news at an unprecedented rate. In general, fake news in social media is defined as the verifiably false information that is generated by malicious users or social bots intentionally, with the aim of misleading the public. There have been research showing that fake news spread more quickly and broadly than true news~\cite{vosoughi2018spread}.
	
	Detecting fake news, especially at an early stage, is complex and challenging due to the characteristics of fake news. Various types of information are integrated when designing detection strategies, including news-related and social-related features~\cite{zhou2019fake}. Among the whole process of detection algorithms, extracting information from network-based features is a procedure to improve the performance of detection results. In social media, users form different kinds of networks in terms of interests, topics, and relations. For example, \cite{nguyen2020fang} proposed a heterogeneous graph to incorporate all major social actors and their interactions into node representations, which is constructed by user, news, and sources. Other types of networks also exist, for instance, co-occurrence network indicating user engagements, friendship network showing the following relationships, and diffusion network tracking the source of fake news. We refer readers to~\cite{zhou2020survey, shu2017fake} for more information about the research on fake news.

	\subsection{Cyberbullying}
	Based on the definition of bullying, cyberbullying is, by extension, defined as an aggressive act intentionally carried out by a group or individual using an electronic device, against people who cannot easily defend themselves. Research have shown that cyberbullying is quite prevalent on social media with  54\% of young people reportedly cyberbullied on Facebook~\cite{salawu2017approaches}. Apart from the traditional research using merely content-based features, recent years have witnessed a proliferation of research focusing on incorporating network-based features (e.g., number of friends, uploads, likes and so on) in detection systems~\cite{squicciarini2015identification}. For example, Cheng et al.~\cite{cheng2019xbully} refined the cyberbullying detection problem within a multi-modal context. Then, the problem is a process of multiple modalities exploited in a collaborative fashion.
	
	\subsection{Fake reviews}
	Rating platforms require aggregating a large-scale collection of user reviews and ratings about items (e.g., products, movies, or other users), which play a central role in deciding what service to purchase, restaurant to patronize, and movie to watch, to name but a few. However, fraudulent users give fake ratings and even malicious reviews out of personal interest or prejudice. Therefore, it is necessary to detect such users and eliminate the influence of malicious competition among peers on rating systems.
	
	The algorithm of detecting fraudulent users is formulized as a process to calculate the trustworthiness of a user and its ratings. Networks used in this line of research are diverse, ranging from homogeneous to heterogeneous, such as user-product bipartite network with signed edges \cite{akoglu2013opinion}, homogeneous co-review graph with weighted or unweighted edges \cite{kaghazgaran2018combating}, and a bipartite rating graph with directed and weighted edges~\cite{kumar2018rev2}.
	
	\subsection{Electrical grid}
	A power grid, also known as an electrical grid, can be constructed as a graph, with nodes denoting generators and edges indicating power lines. Several research questions about anomaly detection or prediction need to be solved in an electrical system. For example, when an electrical component has failed or is going to fail, how could we detect or predict it accurately? Another more challenging problem is to determine the locations of a limited budget of sensors, then it is easier to detect and predict grid anomalies in advance.
	
	Existing anomaly detection algorithms mainly focus on graph theory-based measures instead of graph learning methods. For example, Hooi et al.~\cite{hooi2018gridwatch} detected sensor-level anomalies by designing detectors for three common types of anomalies, and constructed an optimization strategy for sensor placement, with the aim of maximizing the probability of detecting an anomaly. Li et al.~\cite{li2020dynamic} proposed an index to measure the distance between each past graph and the current graph, thereby generating anomaly scores of a graph in a specific timestamp.
	
\subsection{Financial defaulter}
Despite the huge benefits created by online financial services to the society, we have been witnessing a huge growth in financial frauds. The types of frauds in financial scenarios include cash-out behaviors~\cite{hu2019cash}, insurance fraud~\cite{wang2019semi}, and default users~\cite{zhong2020financial}. These frauds severely damage the security of users and service providers, which is a serious problem that needs to be solved.
	
In financial systems, users engage in interactions and have multiple sources of information. These data form a large multi-view network that conventional methods cannot fully exploit. By integrating the features of various kinds of objects and their interactions, \cite{hu2019cash} aims to identify whether a user is a cash-out user or not. Default user is defined as a user who is likely to fail to make required payments on time in the future~\cite{zhong2020financial}. Hence, these kinds of research questions are generally formulated as binary classification problems.
	
\subsection{Anomalous citations and co-authors}
In the context of big scholarly data, the concept of Academic Social Networks (ASNs) is created. ASNs are complex heterogeneous networks formed by academic entities and their relationships~\cite{kong2019academic}, such as co-authorship network, co-citation network, and co-word network. Among these complex relationships and interactions, abnormal academic behaviors (e.g., citations and collaborations) commonly and implicitly exist in different kinds of ASNs~\cite{bai2016identifying}.
	
In~\cite{jolly2020unsupervised}, the authors proposed five heuristic rules to define five types of anomalous citation from the perspective of journal-level citation count. Another kind of anomalous citation is defined in terms of citation context, which is identified by analyzing the context similarity between two publications. As for academic collaboration, Fan et al.~\cite{fan2018abnormal} analyzed the author-paper-author meta-path (co-authorship) to discover rare pattern events in a heterogeneous information network, where each event is denoted by a specific meta-path. To detect anomalous citations, Liu et al. \cite{liu2022deep} first applied transfer learning to automatically identify unmarked citation purposes and then, applied a deep graph learning framework for anomalous citation detection.
\subsection{Urban computing and mobile sensing}
In the process of constructing a smart city, urban anomalies like traffic congestion widely occur and sometimes it may bring serious environmental, economic, and social threats to the public~\cite{zhang2020urban}. To prevent tragedies, the use of smart devices and sensors to detect urban anomalies is of great value. Since the urban data are collected in real time through mobile devices or distributed sensors, they are generally modeled as spatial-temporal graphs that have timesteps and location tags.

The emergence of urban big data inspired many novel research on anomaly detection and prediction, such as air quality prediction~\cite{zheng2013u}, traffic speed prediction~\cite{xie2020deep}, and crime detection~\cite{wang2020deep}. By modeling the urban data as a global cross-region hypergraph, \cite{xia2021spatial} proposed to encode crime dependent representations and spatial temporal dynamics for crime prediction. As for intelligent transportation system, \cite{wang2022a2djp} proposed a model based on integration of a modified GCN and LSTM to predict anomaly distribution and duration.

\section{Future directions}~\label{sec8}
There are several ongoing or future research directions that are worthy of discussion. In this section, we summarize five potential research directions of anomaly analytics on graph data.
\subsection{Anomaly detection on graphs with complex types}
Most of the existing research focus on detecting anomalies on simple graphs, while real-world networks are more complicated and have different types, such as heterogeneous graph with multiple node types~\cite{zhang2019heterogeneous, zhao2021heterogeneous}, spatio-temporal graph evolving with time~\cite{huang2021temporal}, and hypergraph with relations not limited to pairwise relations~\cite{wang2021metro}. Detecting and predicting anomalies on these complex graphs involve technical challenges. For example, as nodes and links which are representing entities and relationships in real-world networks are constantly evolving over time, anomalous entities/relationships might sometimes present normal behaviors as other entities in static networks. This will decrease the accuracy of anomaly detection methods~\cite{bilgin2006dynamic}. So, how to model the temporal characteristic of dynamic networks and update real-time graph embeddings remain as important challenges. As for heterogeneous graph anomaly detection, how to incorporate both attribute and structure information of various types of nodes and edges into the graph learning model is a key problem~\cite{wang2019heterogeneous}. Therefore, anomaly detection and prediction on complex graphs is still a potential research direction need to be further explored. 
\subsection{Interpretable and robust anomaly detection algorithms}
Despite the fact that representation learning methods relieve much of the cost of handling features manually, a major limitation of current graph embedding approaches is the lack of interpretability. Unlike the general tasks, an interpretable model for anomaly analytics can help people to understand the results, thereby avoiding the potential model risks and human bias. Apart from result visualization and benchmark evaluation, efforts must be devoted to improving the interpretability of graph learning methods from the perspective of neural network structures. Interpretable models for anomaly analytics can be presented clearly and are likely to be accepted by the public. For example, \cite{nguyen2020fang} could identify which neighbor of an anomalous node influenced most by differentiating the edge weights generated by the attention mechanism. Moreover, it is acknowledged that adversarial attacks will influence the model's accuracy and performance. Therefore, how to enhance the robustness of a model is another challenge. Several studies regarding interpretability and robustness can be found in~\cite{ying2019gnnexplainer, zhu2019robust, du2019techniques}.
	
\subsection{Anomalous subgraph detection}
Recent years have substantially witnessed superior performance on detecting point anomalies, while users in real-world tend to carry out abnormal behaviors in groups, such as spreading rumors and telecommunications fraud. Apart from this, graph data have diverse structures and forms, while existing methods are not available for all situations. Methods regarding group or subgraph anomaly detection have been less explored~\cite{yu2020detecting}, especially for complex network structures like hypergraph and multi-modal graph.
\subsection{Novel applications of anomaly prediction}
	
While most of the works we reviewed aim to detect existing anomalies, there are still significant works to be done in predicting anomalies in advance. For example, predicting traffic jams ahead of time in transportation networks can help people map out another travel route and avoid congestion situations~\cite{kong2019short}. Therefore, developing representation learning frameworks that are truly appropriate to anomaly prediction settings in a timely manner is necessary to prevent accidents, huge financial loss, or even deaths.
	
As a special data structure, graphs are often employed as an auxiliary tool to combine with many research fields, such as biology, chemistry, and social science. Considering that anomalies are defined quite different in various scenarios, domain knowledge is thereby necessary when applying anomaly prediction models into novel applications.
\subsection{Fairness in anomaly analysis}
Recent years have witnessed a surge of attention in fair machine learning models~\cite{liu2018delayed}. Consequently, several fairness metrics have been proposed as the constraints of objective function in various machine learning models to guarantee the equality of the prediction results. As for anomaly detection tasks, whether users can trust the detection results of the models is still a significant problem~\cite{zhang2021towards}. It is due to the fact that incorrect anomaly detection results may sometimes lead to serious consequences, such as wrong  object detection when dealing with criminals and fraudsters. In  \cite{shekhar2021fairod}, the authors formally defined fairness-aware outlier detection problem and proposed a model to satisfy the fairness criteria. However, fairness on graph anomaly detection is still of concern and deserve further attention.

\section{Conclusion}~\label{sec9}
In this survey, we comprehensively reviewed anomaly analytics methods using graph learning models. The algorithms are divided into four classifications: graph convolutional network-based methods, graph attention network-based methods, graph autoencoder-based methods, and other graph learning models. A thorough comparison and summarization of these methods are provided in this paper. Then, we enumerated and briefly introduced several real-world applications of graph anomaly analytics. Finally, we discussed five future research directions when applying deep learning methods into graph anomaly analytics.
	
\bibliographystyle{ACM-Reference-Format}
\bibliography{bib}


\begin{thebibliography}{142}

%%% the Web address.

\ifx \showCODEN    \undefined \def \showCODEN     #1{\unskip}     \fi
\ifx \showDOI      \undefined \def \showDOI       #1{#1}\fi
\ifx \showISBNx    \undefined \def \showISBNx     #1{\unskip}     \fi
\ifx \showISBNxiii \undefined \def \showISBNxiii  #1{\unskip}     \fi
\ifx \showISSN     \undefined \def \showISSN      #1{\unskip}     \fi
\ifx \showLCCN     \undefined \def \showLCCN      #1{\unskip}     \fi
\ifx \shownote     \undefined \def \shownote      #1{#1}          \fi
\ifx \showarticletitle \undefined \def \showarticletitle #1{#1}   \fi
\ifx \showURL      \undefined \def \showURL       {\relax}        \fi
% invisible to TeX
\providecommand\bibfield[2]{#2}
\providecommand\bibinfo[2]{#2}
\providecommand\natexlab[1]{#1}
\providecommand\showeprint[2][]{arXiv:#2}

\bibitem[\protect\citeauthoryear{Ahmed, Palleti, and Mathur}{Ahmed
  et~al\mbox{.}}{2017}]%
        {ahmed2017wadi}
\bibfield{author}{\bibinfo{person}{Chuadhry~Mujeeb Ahmed},
  \bibinfo{person}{Venkata~Reddy Palleti}, {and} \bibinfo{person}{Aditya~P
  Mathur}.} \bibinfo{year}{2017}\natexlab{}.
\newblock \showarticletitle{{WADI}: a water distribution testbed for research
  in the design of secure cyber physical systems}. In
  \bibinfo{booktitle}{\emph{Proceedings of the 3rd International Workshop on
  Cyber-Physical Systems for Smart Water Networks}}. \bibinfo{pages}{25--28}.
\newblock


\bibitem[\protect\citeauthoryear{Akoglu, Chandy, and Faloutsos}{Akoglu
  et~al\mbox{.}}{2013}]%
        {akoglu2013opinion}
\bibfield{author}{\bibinfo{person}{Leman Akoglu}, \bibinfo{person}{Rishi
  Chandy}, {and} \bibinfo{person}{Christos Faloutsos}.}
  \bibinfo{year}{2013}\natexlab{}.
\newblock \showarticletitle{Opinion fraud detection in online reviews by
  network effects}. In \bibinfo{booktitle}{\emph{Proceedings of the
  International AAAI Conference on Web and Social Media}},
  Vol.~\bibinfo{volume}{7}. \bibinfo{pages}{2--11}.
\newblock


\bibitem[\protect\citeauthoryear{Akoglu, Tong, and Koutra}{Akoglu
  et~al\mbox{.}}{2015}]%
        {akoglu2015graph}
\bibfield{author}{\bibinfo{person}{Leman Akoglu}, \bibinfo{person}{Hanghang
  Tong}, {and} \bibinfo{person}{Danai Koutra}.}
  \bibinfo{year}{2015}\natexlab{}.
\newblock \showarticletitle{Graph based anomaly detection and description: {A}
  survey}.
\newblock \bibinfo{journal}{\emph{Data mining and knowledge discovery}}
  \bibinfo{volume}{29}, \bibinfo{number}{3} (\bibinfo{year}{2015}),
  \bibinfo{pages}{626--688}.
\newblock


\bibitem[\protect\citeauthoryear{An and Cho}{An and Cho}{2015}]%
        {an2015variational}
\bibfield{author}{\bibinfo{person}{Jinwon An} {and} \bibinfo{person}{Sungzoon
  Cho}.} \bibinfo{year}{2015}\natexlab{}.
\newblock \showarticletitle{Variational autoencoder based anomaly detection
  using reconstruction probability}.
\newblock \bibinfo{journal}{\emph{Special Lecture on IE}} \bibinfo{volume}{2},
  \bibinfo{number}{1} (\bibinfo{year}{2015}), \bibinfo{pages}{1--18}.
\newblock


\bibitem[\protect\citeauthoryear{Bai, Xia, Lee, Zhang, and Ning}{Bai
  et~al\mbox{.}}{2016}]%
        {bai2016identifying}
\bibfield{author}{\bibinfo{person}{Xiaomei Bai}, \bibinfo{person}{Feng Xia},
  \bibinfo{person}{Ivan Lee}, \bibinfo{person}{Jun Zhang}, {and}
  \bibinfo{person}{Zhaolong Ning}.} \bibinfo{year}{2016}\natexlab{}.
\newblock \showarticletitle{Identifying anomalous citations for objective
  evaluation of scholarly article impact}.
\newblock \bibinfo{journal}{\emph{PloS one}} \bibinfo{volume}{11},
  \bibinfo{number}{9} (\bibinfo{year}{2016}), \bibinfo{pages}{e0162364}.
\newblock


\bibitem[\protect\citeauthoryear{Bandyopadhyay, Vivek, and Murty}{Bandyopadhyay
  et~al\mbox{.}}{2020}]%
        {bandyopadhyay2020outlier}
\bibfield{author}{\bibinfo{person}{Sambaran Bandyopadhyay},
  \bibinfo{person}{Saley~Vishal Vivek}, {and} \bibinfo{person}{MN Murty}.}
  \bibinfo{year}{2020}\natexlab{}.
\newblock \showarticletitle{Outlier resistant unsupervised deep architectures
  for attributed network embedding}. In \bibinfo{booktitle}{\emph{Proceedings
  of the 13th International Conference on Web Search and Data Mining}}.
  \bibinfo{pages}{25--33}.
\newblock


\bibitem[\protect\citeauthoryear{Bian, Xiao, Xu, Zhao, Huang, Rong, and
  Huang}{Bian et~al\mbox{.}}{2020}]%
        {bian2020rumor}
\bibfield{author}{\bibinfo{person}{Tian Bian}, \bibinfo{person}{Xi Xiao},
  \bibinfo{person}{Tingyang Xu}, \bibinfo{person}{Peilin Zhao},
  \bibinfo{person}{Wenbing Huang}, \bibinfo{person}{Yu Rong}, {and}
  \bibinfo{person}{Junzhou Huang}.} \bibinfo{year}{2020}\natexlab{}.
\newblock \showarticletitle{Rumor detection on social media with bi-directional
  graph convolutional networks}. In \bibinfo{booktitle}{\emph{Proceedings of
  the AAAI Conference on Artificial Intelligence}}, Vol.~\bibinfo{volume}{34}.
  \bibinfo{pages}{549--556}.
\newblock


\bibitem[\protect\citeauthoryear{Bilgin and Yener}{Bilgin and Yener}{2006}]%
        {bilgin2006dynamic}
\bibfield{author}{\bibinfo{person}{Cemal~Cagatay Bilgin} {and}
  \bibinfo{person}{B{\"u}lent Yener}.} \bibinfo{year}{2006}\natexlab{}.
\newblock \showarticletitle{Dynamic network evolution: Models, clustering,
  anomaly detection}.
\newblock \bibinfo{journal}{\emph{IEEE Networks}} \bibinfo{number}{1}
  (\bibinfo{year}{2006}).
\newblock


\bibitem[\protect\citeauthoryear{Bojchevski and G{\"u}nnemann}{Bojchevski and
  G{\"u}nnemann}{2018}]%
        {bojchevski2018deep}
\bibfield{author}{\bibinfo{person}{Aleksandar Bojchevski} {and}
  \bibinfo{person}{Stephan G{\"u}nnemann}.} \bibinfo{year}{2018}\natexlab{}.
\newblock \showarticletitle{Deep Gaussian Embedding of Graphs: Unsupervised
  Inductive Learning via Ranking}. In \bibinfo{booktitle}{\emph{International
  Conference on Learning Representations}}. \bibinfo{pages}{1--13}.
\newblock


\bibitem[\protect\citeauthoryear{Cai, Chen, Luo, Gui, Ni, Li, and Chen}{Cai
  et~al\mbox{.}}{2021}]%
        {cai2021structural}
\bibfield{author}{\bibinfo{person}{Lei Cai}, \bibinfo{person}{Zhengzhang Chen},
  \bibinfo{person}{Chen Luo}, \bibinfo{person}{Jiaping Gui},
  \bibinfo{person}{Jingchao Ni}, \bibinfo{person}{Ding Li}, {and}
  \bibinfo{person}{Haifeng Chen}.} \bibinfo{year}{2021}\natexlab{}.
\newblock \showarticletitle{Structural temporal graph neural networks for
  anomaly detection in dynamic graphs}. In
  \bibinfo{booktitle}{\emph{Proceedings of the 30th ACM International
  Conference on Information \& Knowledge Management}}.
  \bibinfo{pages}{3747--3756}.
\newblock


\bibitem[\protect\citeauthoryear{Chalapathy and Chawla}{Chalapathy and
  Chawla}{2019}]%
        {chalapathy2019deep}
\bibfield{author}{\bibinfo{person}{Raghavendra Chalapathy} {and}
  \bibinfo{person}{Sanjay Chawla}.} \bibinfo{year}{2019}\natexlab{}.
\newblock \showarticletitle{Deep learning for anomaly detection: A survey}.
\newblock \bibinfo{journal}{\emph{arXiv preprint arXiv:1901.03407}}
  (\bibinfo{year}{2019}).
\newblock


\bibitem[\protect\citeauthoryear{Chandola, Banerjee, and Kumar}{Chandola
  et~al\mbox{.}}{2009}]%
        {chandola2009anomaly}
\bibfield{author}{\bibinfo{person}{Varun Chandola}, \bibinfo{person}{Arindam
  Banerjee}, {and} \bibinfo{person}{Vipin Kumar}.}
  \bibinfo{year}{2009}\natexlab{}.
\newblock \showarticletitle{Anomaly detection: A survey}.
\newblock \bibinfo{journal}{\emph{ACM computing surveys (CSUR)}}
  \bibinfo{volume}{41}, \bibinfo{number}{3} (\bibinfo{year}{2009}),
  \bibinfo{pages}{1--58}.
\newblock


\bibitem[\protect\citeauthoryear{Chen, Zhang, Zhang, Dong, Song, Zhang, Xu,
  Kharlamov, and Tang}{Chen et~al\mbox{.}}{2022}]%
        {chen2022gccad}
\bibfield{author}{\bibinfo{person}{Bo Chen}, \bibinfo{person}{Jing Zhang},
  \bibinfo{person}{Xiaokang Zhang}, \bibinfo{person}{Yuxiao Dong},
  \bibinfo{person}{Jian Song}, \bibinfo{person}{Peng Zhang},
  \bibinfo{person}{Kaibo Xu}, \bibinfo{person}{Evgeny Kharlamov}, {and}
  \bibinfo{person}{Jie Tang}.} \bibinfo{year}{2022}\natexlab{}.
\newblock \showarticletitle{GCCAD: Graph Contrastive Learning for Anomaly
  Detection}.
\newblock \bibinfo{journal}{\emph{IEEE Transactions on Knowledge and Data
  Engineering}} (\bibinfo{year}{2022}).
\newblock


\bibitem[\protect\citeauthoryear{Chen, Li, and Bruna}{Chen
  et~al\mbox{.}}{2019}]%
        {chen2018supervised}
\bibfield{author}{\bibinfo{person}{Zhengdao Chen}, \bibinfo{person}{Lisha Li},
  {and} \bibinfo{person}{Joan Bruna}.} \bibinfo{year}{2019}\natexlab{}.
\newblock \showarticletitle{Supervised Community Detection with Line Graph
  Neural Networks}. In \bibinfo{booktitle}{\emph{International Conference on
  Learning Representations}}.
\newblock


\bibitem[\protect\citeauthoryear{Chen, Liu, Wang, Dai, Lv, and Bo}{Chen
  et~al\mbox{.}}{2020}]%
        {chen2020generative}
\bibfield{author}{\bibinfo{person}{Zhenxing Chen}, \bibinfo{person}{Bo Liu},
  \bibinfo{person}{Meiqing Wang}, \bibinfo{person}{Peng Dai},
  \bibinfo{person}{Jun Lv}, {and} \bibinfo{person}{Liefeng Bo}.}
  \bibinfo{year}{2020}\natexlab{}.
\newblock \showarticletitle{Generative Adversarial Attributed Network Anomaly
  Detection}. In \bibinfo{booktitle}{\emph{Proceedings of the 29th ACM
  International Conference on Information \& Knowledge Management}}.
  \bibinfo{pages}{1989--1992}.
\newblock


\bibitem[\protect\citeauthoryear{Cheng, Li, Silva, Hall, and Liu}{Cheng
  et~al\mbox{.}}{2019}]%
        {cheng2019xbully}
\bibfield{author}{\bibinfo{person}{Lu Cheng}, \bibinfo{person}{Jundong Li},
  \bibinfo{person}{Yasin~N Silva}, \bibinfo{person}{Deborah~L Hall}, {and}
  \bibinfo{person}{Huan Liu}.} \bibinfo{year}{2019}\natexlab{}.
\newblock \showarticletitle{{XBully}: Cyberbullying detection within a
  multi-modal context}. In \bibinfo{booktitle}{\emph{Proceedings of the Twelfth
  ACM International Conference on Web Search and Data Mining}}.
  \bibinfo{pages}{339--347}.
\newblock


\bibitem[\protect\citeauthoryear{Cheng, Shu, Wu, Silva, Hall, and Liu}{Cheng
  et~al\mbox{.}}{2020}]%
        {cheng2020unsupervised}
\bibfield{author}{\bibinfo{person}{Lu Cheng}, \bibinfo{person}{Kai Shu},
  \bibinfo{person}{Siqi Wu}, \bibinfo{person}{Yasin~N Silva},
  \bibinfo{person}{Deborah~L Hall}, {and} \bibinfo{person}{Huan Liu}.}
  \bibinfo{year}{2020}\natexlab{}.
\newblock \showarticletitle{Unsupervised Cyberbullying Detection via
  Time-Informed Gaussian Mixture Model}. In
  \bibinfo{booktitle}{\emph{Proceedings of the 29th ACM International
  Conference on Information \& Knowledge Management}}.
  \bibinfo{pages}{185--194}.
\newblock


\bibitem[\protect\citeauthoryear{De~Choudhury, Sundaram, John, and
  Seligmann}{De~Choudhury et~al\mbox{.}}{2009}]%
        {de2009social}
\bibfield{author}{\bibinfo{person}{Munmun De~Choudhury}, \bibinfo{person}{Hari
  Sundaram}, \bibinfo{person}{Ajita John}, {and}
  \bibinfo{person}{Dor{\'e}e~Duncan Seligmann}.}
  \bibinfo{year}{2009}\natexlab{}.
\newblock \showarticletitle{Social synchrony: Predicting mimicry of user
  actions in online social media}. In \bibinfo{booktitle}{\emph{2009
  International conference on computational science and engineering}},
  Vol.~\bibinfo{volume}{4}. IEEE, \bibinfo{pages}{151--158}.
\newblock


\bibitem[\protect\citeauthoryear{Deng and Hooi}{Deng and Hooi}{2021}]%
        {deng2021graph}
\bibfield{author}{\bibinfo{person}{Ailin Deng} {and} \bibinfo{person}{Bryan
  Hooi}.} \bibinfo{year}{2021}\natexlab{}.
\newblock \showarticletitle{Graph neural network-based anomaly detection in
  multivariate time series}. In \bibinfo{booktitle}{\emph{Proceedings of the
  AAAI Conference on Artificial Intelligence}}, Vol.~\bibinfo{volume}{35}.
  \bibinfo{pages}{4027--4035}.
\newblock


\bibitem[\protect\citeauthoryear{Di~Crescenzo, Gavazzi, Legnaro, Troccoli,
  Bordino, and Gullo}{Di~Crescenzo et~al\mbox{.}}{2017}]%
        {di2017hermevent}
\bibfield{author}{\bibinfo{person}{Cristiano Di~Crescenzo},
  \bibinfo{person}{Giulia Gavazzi}, \bibinfo{person}{Giacomo Legnaro},
  \bibinfo{person}{Elena Troccoli}, \bibinfo{person}{Ilaria Bordino}, {and}
  \bibinfo{person}{Francesco Gullo}.} \bibinfo{year}{2017}\natexlab{}.
\newblock \showarticletitle{{HERMEVENT}: a news collection for emerging-event
  detection}. In \bibinfo{booktitle}{\emph{Proceedings of the 7th International
  Conference on Web Intelligence, Mining and Semantics}}.
  \bibinfo{pages}{1--10}.
\newblock


\bibitem[\protect\citeauthoryear{Ding, Li, Agarwal, and Liu}{Ding
  et~al\mbox{.}}{2020}]%
        {ding2020inductive}
\bibfield{author}{\bibinfo{person}{Kaize Ding}, \bibinfo{person}{Jundong Li},
  \bibinfo{person}{Nitin Agarwal}, {and} \bibinfo{person}{Huan Liu}.}
  \bibinfo{year}{2020}\natexlab{}.
\newblock \showarticletitle{Inductive anomaly detection on attributed
  networks}. In \bibinfo{booktitle}{\emph{29th International Joint Conference
  on Artificial Intelligence, IJCAI 2020}}. International Joint Conferences on
  Artificial Intelligence, \bibinfo{pages}{1288--1294}.
\newblock


\bibitem[\protect\citeauthoryear{Ding, Li, Bhanushali, and Liu}{Ding
  et~al\mbox{.}}{2019}]%
        {ding2019deep}
\bibfield{author}{\bibinfo{person}{Kaize Ding}, \bibinfo{person}{Jundong Li},
  \bibinfo{person}{Rohit Bhanushali}, {and} \bibinfo{person}{Huan Liu}.}
  \bibinfo{year}{2019}\natexlab{}.
\newblock \showarticletitle{Deep anomaly detection on attributed networks}. In
  \bibinfo{booktitle}{\emph{Proceedings of the 2019 SIAM International
  Conference on Data Mining}}. SIAM, \bibinfo{pages}{594--602}.
\newblock


\bibitem[\protect\citeauthoryear{Ding, Zhou, Tong, and Liu}{Ding
  et~al\mbox{.}}{2021}]%
        {ding2021few}
\bibfield{author}{\bibinfo{person}{Kaize Ding}, \bibinfo{person}{Qinghai Zhou},
  \bibinfo{person}{Hanghang Tong}, {and} \bibinfo{person}{Huan Liu}.}
  \bibinfo{year}{2021}\natexlab{}.
\newblock \showarticletitle{Few-shot network anomaly detection via
  cross-network meta-learning}. In \bibinfo{booktitle}{\emph{Proceedings of the
  Web Conference 2021}}. \bibinfo{pages}{2448--2456}.
\newblock


\bibitem[\protect\citeauthoryear{Ding, Katenka, Barford, Kolaczyk, and
  Crovella}{Ding et~al\mbox{.}}{2012}]%
        {ding2012intrusion}
\bibfield{author}{\bibinfo{person}{Qi Ding}, \bibinfo{person}{Natallia
  Katenka}, \bibinfo{person}{Paul Barford}, \bibinfo{person}{Eric Kolaczyk},
  {and} \bibinfo{person}{Mark Crovella}.} \bibinfo{year}{2012}\natexlab{}.
\newblock \showarticletitle{Intrusion as (anti) social communication:
  {Characterization} and detection}. In \bibinfo{booktitle}{\emph{Proceedings
  of the 18th ACM SIGKDD international conference on Knowledge discovery and
  data mining}}. \bibinfo{pages}{886--894}.
\newblock


\bibitem[\protect\citeauthoryear{Do, Tran, and Venkatesh}{Do
  et~al\mbox{.}}{2019}]%
        {do2019graph}
\bibfield{author}{\bibinfo{person}{Kien Do}, \bibinfo{person}{Truyen Tran},
  {and} \bibinfo{person}{Svetha Venkatesh}.} \bibinfo{year}{2019}\natexlab{}.
\newblock \showarticletitle{Graph transformation policy network for chemical
  reaction prediction}. In \bibinfo{booktitle}{\emph{Proceedings of the 25th
  ACM SIGKDD International Conference on Knowledge Discovery \& Data Mining}}.
  \bibinfo{pages}{750--760}.
\newblock


\bibitem[\protect\citeauthoryear{Dong, Zheng, Quoc Viet~Hung, Su, and Li}{Dong
  et~al\mbox{.}}{2019}]%
        {dong2019multiple}
\bibfield{author}{\bibinfo{person}{Ming Dong}, \bibinfo{person}{Bolong Zheng},
  \bibinfo{person}{Nguyen Quoc Viet~Hung}, \bibinfo{person}{Han Su}, {and}
  \bibinfo{person}{Guohui Li}.} \bibinfo{year}{2019}\natexlab{}.
\newblock \showarticletitle{Multiple rumor source detection with graph
  convolutional networks}. In \bibinfo{booktitle}{\emph{Proceedings of the 28th
  ACM International Conference on Information and Knowledge Management}}.
  \bibinfo{pages}{569--578}.
\newblock


\bibitem[\protect\citeauthoryear{Dou, Liu, Sun, Deng, Peng, and Yu}{Dou
  et~al\mbox{.}}{2020}]%
        {dou2020enhancing}
\bibfield{author}{\bibinfo{person}{Yingtong Dou}, \bibinfo{person}{Zhiwei Liu},
  \bibinfo{person}{Li Sun}, \bibinfo{person}{Yutong Deng}, \bibinfo{person}{Hao
  Peng}, {and} \bibinfo{person}{Philip~S Yu}.} \bibinfo{year}{2020}\natexlab{}.
\newblock \showarticletitle{Enhancing graph neural network-based fraud
  detectors against camouflaged fraudsters}. In
  \bibinfo{booktitle}{\emph{Proceedings of the 29th ACM International
  Conference on Information \& Knowledge Management}}.
  \bibinfo{pages}{315--324}.
\newblock


\bibitem[\protect\citeauthoryear{Du, Liu, and Hu}{Du et~al\mbox{.}}{2019}]%
        {du2019techniques}
\bibfield{author}{\bibinfo{person}{Mengnan Du}, \bibinfo{person}{Ninghao Liu},
  {and} \bibinfo{person}{Xia Hu}.} \bibinfo{year}{2019}\natexlab{}.
\newblock \showarticletitle{Techniques for interpretable machine learning}.
\newblock \bibinfo{journal}{\emph{Commun. ACM}} \bibinfo{volume}{63},
  \bibinfo{number}{1} (\bibinfo{year}{2019}), \bibinfo{pages}{68--77}.
\newblock


\bibitem[\protect\citeauthoryear{Duan, Tong, Li, Lu, Shi, and Zhang}{Duan
  et~al\mbox{.}}{2020}]%
        {duan2020aane}
\bibfield{author}{\bibinfo{person}{Dongsheng Duan}, \bibinfo{person}{Lingling
  Tong}, \bibinfo{person}{Yangxi Li}, \bibinfo{person}{Jie Lu},
  \bibinfo{person}{Lei Shi}, {and} \bibinfo{person}{Cheng Zhang}.}
  \bibinfo{year}{2020}\natexlab{}.
\newblock \showarticletitle{{AANE}: Anomaly Aware Network Embedding For
  Anomalous Link Detection}. In \bibinfo{booktitle}{\emph{2020 IEEE
  International Conference on Data Mining (ICDM)}}. IEEE,
  \bibinfo{pages}{1002--1007}.
\newblock


\bibitem[\protect\citeauthoryear{El-Fiqi, Wang, Kasmarik, Bezerianos, Tan, and
  Abbass}{El-Fiqi et~al\mbox{.}}{2021}]%
        {el2021weighted}
\bibfield{author}{\bibinfo{person}{Heba El-Fiqi}, \bibinfo{person}{Min Wang},
  \bibinfo{person}{Kathryn Kasmarik}, \bibinfo{person}{Anastasios Bezerianos},
  \bibinfo{person}{Kay~Chen Tan}, {and} \bibinfo{person}{Hussein~A Abbass}.}
  \bibinfo{year}{2021}\natexlab{}.
\newblock \showarticletitle{Weighted Gate Layer Autoencoders}.
\newblock \bibinfo{journal}{\emph{IEEE Transactions on Cybernetics}}
  \bibinfo{volume}{52}, \bibinfo{number}{8} (\bibinfo{year}{2021}),
  \bibinfo{pages}{7242--7253}.
\newblock


\bibitem[\protect\citeauthoryear{Fan, Shi, and Wang}{Fan et~al\mbox{.}}{2018}]%
        {fan2018abnormal}
\bibfield{author}{\bibinfo{person}{Shaohua Fan}, \bibinfo{person}{Chuan Shi},
  {and} \bibinfo{person}{Xiao Wang}.} \bibinfo{year}{2018}\natexlab{}.
\newblock \showarticletitle{Abnormal event detection via heterogeneous
  information network embedding}. In \bibinfo{booktitle}{\emph{Proceedings of
  the 27th ACM International Conference on Information and Knowledge
  Management}}. \bibinfo{pages}{1483--1486}.
\newblock


\bibitem[\protect\citeauthoryear{Fan, Ye, Peng, Zhang, Zhang, Xiao, Shi, Xiong,
  Shao, and Zhao}{Fan et~al\mbox{.}}{2020}]%
        {fan2020metagraph}
\bibfield{author}{\bibinfo{person}{Yujie Fan}, \bibinfo{person}{Yanfang Ye},
  \bibinfo{person}{Qian Peng}, \bibinfo{person}{Jianfei Zhang},
  \bibinfo{person}{Yiming Zhang}, \bibinfo{person}{Xusheng Xiao},
  \bibinfo{person}{Chuan Shi}, \bibinfo{person}{Qi Xiong},
  \bibinfo{person}{Fudong Shao}, {and} \bibinfo{person}{Liang Zhao}.}
  \bibinfo{year}{2020}\natexlab{}.
\newblock \showarticletitle{Metagraph Aggregated Heterogeneous Graph Neural
  Network for Illicit Traded Product Identification in Underground Market}. In
  \bibinfo{booktitle}{\emph{2020 IEEE International Conference on Data Mining
  (ICDM)}}. IEEE, \bibinfo{pages}{132--141}.
\newblock


\bibitem[\protect\citeauthoryear{Festl and Quandt}{Festl and Quandt}{2013}]%
        {festl2013social}
\bibfield{author}{\bibinfo{person}{Ruth Festl} {and} \bibinfo{person}{Thorsten
  Quandt}.} \bibinfo{year}{2013}\natexlab{}.
\newblock \showarticletitle{Social relations and cyberbullying: The influence
  of individual and structural attributes on victimization and perpetration via
  the internet}.
\newblock \bibinfo{journal}{\emph{Human communication research}}
  \bibinfo{volume}{39}, \bibinfo{number}{1} (\bibinfo{year}{2013}),
  \bibinfo{pages}{101--126}.
\newblock


\bibitem[\protect\citeauthoryear{Ge, Cheng, and Liu}{Ge et~al\mbox{.}}{2021}]%
        {ge2021improving}
\bibfield{author}{\bibinfo{person}{Suyu Ge}, \bibinfo{person}{Lu Cheng}, {and}
  \bibinfo{person}{Huan Liu}.} \bibinfo{year}{2021}\natexlab{}.
\newblock \showarticletitle{Improving cyberbullying detection with user
  interaction}. In \bibinfo{booktitle}{\emph{Proceedings of the Web Conference
  2021}}. \bibinfo{pages}{496--506}.
\newblock


\bibitem[\protect\citeauthoryear{Grubbs}{Grubbs}{1969}]%
        {grubbs1969procedures}
\bibfield{author}{\bibinfo{person}{Frank~E Grubbs}.}
  \bibinfo{year}{1969}\natexlab{}.
\newblock \showarticletitle{Procedures for detecting outlying observations in
  samples}.
\newblock \bibinfo{journal}{\emph{Technometrics}} \bibinfo{volume}{11},
  \bibinfo{number}{1} (\bibinfo{year}{1969}), \bibinfo{pages}{1--21}.
\newblock


\bibitem[\protect\citeauthoryear{Hamilton, Ying, and Leskovec}{Hamilton
  et~al\mbox{.}}{2017}]%
        {hamilton2017inductive}
\bibfield{author}{\bibinfo{person}{William~L Hamilton}, \bibinfo{person}{Rex
  Ying}, {and} \bibinfo{person}{Jure Leskovec}.}
  \bibinfo{year}{2017}\natexlab{}.
\newblock \showarticletitle{Inductive representation learning on large graphs}.
  In \bibinfo{booktitle}{\emph{Proceedings of the 31st International Conference
  on Neural Information Processing Systems}}. \bibinfo{pages}{1025--1035}.
\newblock


\bibitem[\protect\citeauthoryear{Hessel and Lee}{Hessel and Lee}{2019}]%
        {hessel2019something}
\bibfield{author}{\bibinfo{person}{Jack Hessel} {and} \bibinfo{person}{Lillian
  Lee}.} \bibinfo{year}{2019}\natexlab{}.
\newblock \showarticletitle{Something¡¯s Brewing! {Early} Prediction of
  Controversy-causing Posts from Discussion Features}. In
  \bibinfo{booktitle}{\emph{Proceedings of the 2019 Conference of the North
  American Chapter of the Association for Computational Linguistics: Human
  Language Technologies}}. \bibinfo{pages}{1648--1659}.
\newblock


\bibitem[\protect\citeauthoryear{Hooi, Eswaran, Song, Pandey, Jereminov,
  Pileggi, and Faloutsos}{Hooi et~al\mbox{.}}{2018}]%
        {hooi2018gridwatch}
\bibfield{author}{\bibinfo{person}{Bryan Hooi}, \bibinfo{person}{Dhivya
  Eswaran}, \bibinfo{person}{Hyun~Ah Song}, \bibinfo{person}{Amritanshu
  Pandey}, \bibinfo{person}{Marko Jereminov}, \bibinfo{person}{Larry Pileggi},
  {and} \bibinfo{person}{Christos Faloutsos}.} \bibinfo{year}{2018}\natexlab{}.
\newblock \showarticletitle{{GRIDWATCH}: Sensor placement and anomaly detection
  in the electrical grid}. In \bibinfo{booktitle}{\emph{Joint European
  Conference on Machine Learning and Knowledge Discovery in Databases}}.
  Springer, \bibinfo{pages}{71--86}.
\newblock


\bibitem[\protect\citeauthoryear{Hosseinmardi, Mattson, Rafiq, Han, Lv, and
  Mishra}{Hosseinmardi et~al\mbox{.}}{2015}]%
        {hosseinmardi2015analyzing}
\bibfield{author}{\bibinfo{person}{Homa Hosseinmardi},
  \bibinfo{person}{Sabrina~Arredondo Mattson}, \bibinfo{person}{Rahat~Ibn
  Rafiq}, \bibinfo{person}{Richard Han}, \bibinfo{person}{Qin Lv}, {and}
  \bibinfo{person}{Shivakant Mishra}.} \bibinfo{year}{2015}\natexlab{}.
\newblock \showarticletitle{Analyzing labeled cyberbullying incidents on the
  instagram social network}. In \bibinfo{booktitle}{\emph{International
  conference on social informatics}}. Springer, \bibinfo{pages}{49--66}.
\newblock


\bibitem[\protect\citeauthoryear{Hou, Ren, Zhang, Kong, Zhang, and Xia}{Hou
  et~al\mbox{.}}{2020}]%
        {hou2020network}
\bibfield{author}{\bibinfo{person}{Mingliang Hou}, \bibinfo{person}{Jing Ren},
  \bibinfo{person}{Da Zhang}, \bibinfo{person}{Xiangjie Kong},
  \bibinfo{person}{Dongyu Zhang}, {and} \bibinfo{person}{Feng Xia}.}
  \bibinfo{year}{2020}\natexlab{}.
\newblock \showarticletitle{Network embedding: Taxonomies, frameworks and
  applications}.
\newblock \bibinfo{journal}{\emph{Computer Science Review}}
  \bibinfo{volume}{38} (\bibinfo{year}{2020}), \bibinfo{pages}{100296}.
\newblock


\bibitem[\protect\citeauthoryear{Hu, Zhang, Shi, Zhou, Li, and Qi}{Hu
  et~al\mbox{.}}{2019}]%
        {hu2019cash}
\bibfield{author}{\bibinfo{person}{Binbin Hu}, \bibinfo{person}{Zhiqiang
  Zhang}, \bibinfo{person}{Chuan Shi}, \bibinfo{person}{Jun Zhou},
  \bibinfo{person}{Xiaolong Li}, {and} \bibinfo{person}{Yuan Qi}.}
  \bibinfo{year}{2019}\natexlab{}.
\newblock \showarticletitle{Cash-out user detection based on attributed
  heterogeneous information network with a hierarchical attention mechanism}.
  In \bibinfo{booktitle}{\emph{Proceedings of the AAAI Conference on Artificial
  Intelligence}}, Vol.~\bibinfo{volume}{33}. \bibinfo{pages}{946--953}.
\newblock


\bibitem[\protect\citeauthoryear{Huang, Liu, Van Der~Maaten, and
  Weinberger}{Huang et~al\mbox{.}}{2017}]%
        {huang2017densely}
\bibfield{author}{\bibinfo{person}{Gao Huang}, \bibinfo{person}{Zhuang Liu},
  \bibinfo{person}{Laurens Van Der~Maaten}, {and} \bibinfo{person}{Kilian~Q
  Weinberger}.} \bibinfo{year}{2017}\natexlab{}.
\newblock \showarticletitle{Densely connected convolutional networks}. In
  \bibinfo{booktitle}{\emph{Proceedings of the IEEE conference on computer
  vision and pattern recognition}}. \bibinfo{pages}{4700--4708}.
\newblock


\bibitem[\protect\citeauthoryear{Huang, Shi, Zhou, Wang, Jin, and Fu}{Huang
  et~al\mbox{.}}{2021}]%
        {huang2021temporal}
\bibfield{author}{\bibinfo{person}{Hong Huang}, \bibinfo{person}{Ruize Shi},
  \bibinfo{person}{Wei Zhou}, \bibinfo{person}{Xiao Wang}, \bibinfo{person}{Hai
  Jin}, {and} \bibinfo{person}{Xiaoming Fu}.} \bibinfo{year}{2021}\natexlab{}.
\newblock \showarticletitle{Temporal Heterogeneous Information Network
  Embedding.}. In \bibinfo{booktitle}{\emph{IJCAI}}.
  \bibinfo{pages}{1470--1476}.
\newblock


\bibitem[\protect\citeauthoryear{Jiang}{Jiang}{2016}]%
        {jiang2016catching}
\bibfield{author}{\bibinfo{person}{Meng Jiang}.}
  \bibinfo{year}{2016}\natexlab{}.
\newblock \showarticletitle{Catching social media advertisers with strategy
  analysis}. In \bibinfo{booktitle}{\emph{Proceedings of the First
  International Workshop on Computational Methods for CyberSafety}}.
  \bibinfo{pages}{5--10}.
\newblock


\bibitem[\protect\citeauthoryear{Jin, Liu, Zheng, Chi, Li, and Pan}{Jin
  et~al\mbox{.}}{2021}]%
        {jin2021anemone}
\bibfield{author}{\bibinfo{person}{Ming Jin}, \bibinfo{person}{Yixin Liu},
  \bibinfo{person}{Yu Zheng}, \bibinfo{person}{Lianhua Chi},
  \bibinfo{person}{Yuan-Fang Li}, {and} \bibinfo{person}{Shirui Pan}.}
  \bibinfo{year}{2021}\natexlab{}.
\newblock \showarticletitle{ANEMONE: Graph Anomaly Detection with Multi-Scale
  Contrastive Learning}. In \bibinfo{booktitle}{\emph{Proceedings of the 30th
  ACM International Conference on Information \& Knowledge Management}}.
  \bibinfo{pages}{3122--3126}.
\newblock


\bibitem[\protect\citeauthoryear{Jolly, Jain, Bera, and Chakraborty}{Jolly
  et~al\mbox{.}}{2020}]%
        {jolly2020unsupervised}
\bibfield{author}{\bibinfo{person}{Baani Leen~Kaur Jolly},
  \bibinfo{person}{Lavina Jain}, \bibinfo{person}{Debajyoti Bera}, {and}
  \bibinfo{person}{Tanmoy Chakraborty}.} \bibinfo{year}{2020}\natexlab{}.
\newblock \showarticletitle{Unsupervised anomaly detection in journal-level
  citation networks}. In \bibinfo{booktitle}{\emph{Proceedings of the ACM/IEEE
  Joint Conference on Digital Libraries in 2020}}. \bibinfo{pages}{27--36}.
\newblock


\bibitem[\protect\citeauthoryear{Kaghazgaran, Caverlee, and
  Squicciarini}{Kaghazgaran et~al\mbox{.}}{2018}]%
        {kaghazgaran2018combating}
\bibfield{author}{\bibinfo{person}{Parisa Kaghazgaran}, \bibinfo{person}{James
  Caverlee}, {and} \bibinfo{person}{Anna Squicciarini}.}
  \bibinfo{year}{2018}\natexlab{}.
\newblock \showarticletitle{Combating crowdsourced review manipulators: A
  neighborhood-based approach}. In \bibinfo{booktitle}{\emph{Proceedings of the
  Eleventh ACM International Conference on Web Search and Data Mining}}.
  \bibinfo{pages}{306--314}.
\newblock


\bibitem[\protect\citeauthoryear{Kong, Li, Jiang, and Song}{Kong
  et~al\mbox{.}}{2019a}]%
        {kong2019short}
\bibfield{author}{\bibinfo{person}{Fanhui Kong}, \bibinfo{person}{Jian Li},
  \bibinfo{person}{Bin Jiang}, {and} \bibinfo{person}{Houbing Song}.}
  \bibinfo{year}{2019}\natexlab{a}.
\newblock \showarticletitle{Short-term traffic flow prediction in smart
  multimedia system for {Internet of Vehicles} based on deep belief network}.
\newblock \bibinfo{journal}{\emph{Future Generation Computer Systems}}
  \bibinfo{volume}{93} (\bibinfo{year}{2019}), \bibinfo{pages}{460--472}.
\newblock


\bibitem[\protect\citeauthoryear{Kong, Shi, Yu, Liu, and Xia}{Kong
  et~al\mbox{.}}{2019b}]%
        {kong2019academic}
\bibfield{author}{\bibinfo{person}{Xiangjie Kong}, \bibinfo{person}{Yajie Shi},
  \bibinfo{person}{Shuo Yu}, \bibinfo{person}{Jiaying Liu}, {and}
  \bibinfo{person}{Feng Xia}.} \bibinfo{year}{2019}\natexlab{b}.
\newblock \showarticletitle{Academic social networks: Modeling, analysis,
  mining and applications}.
\newblock \bibinfo{journal}{\emph{Journal of Network and Computer
  Applications}}  \bibinfo{volume}{132} (\bibinfo{year}{2019}),
  \bibinfo{pages}{86--103}.
\newblock


\bibitem[\protect\citeauthoryear{Kong, Song, Xia, Guo, Wang, and Tolba}{Kong
  et~al\mbox{.}}{2018}]%
        {kong2018lotad}
\bibfield{author}{\bibinfo{person}{Xiangjie Kong}, \bibinfo{person}{Ximeng
  Song}, \bibinfo{person}{Feng Xia}, \bibinfo{person}{Haochen Guo},
  \bibinfo{person}{Jinzhong Wang}, {and} \bibinfo{person}{Amr Tolba}.}
  \bibinfo{year}{2018}\natexlab{}.
\newblock \showarticletitle{{LoTAD}: Long-term traffic anomaly detection based
  on crowdsourced bus trajectory data}.
\newblock \bibinfo{journal}{\emph{World Wide Web}} \bibinfo{volume}{21},
  \bibinfo{number}{3} (\bibinfo{year}{2018}), \bibinfo{pages}{825--847}.
\newblock


\bibitem[\protect\citeauthoryear{Kumar, Hooi, Makhija, Kumar, Faloutsos, and
  Subrahmanian}{Kumar et~al\mbox{.}}{2018}]%
        {kumar2018rev2}
\bibfield{author}{\bibinfo{person}{Srijan Kumar}, \bibinfo{person}{Bryan Hooi},
  \bibinfo{person}{Disha Makhija}, \bibinfo{person}{Mohit Kumar},
  \bibinfo{person}{Christos Faloutsos}, {and} \bibinfo{person}{VS
  Subrahmanian}.} \bibinfo{year}{2018}\natexlab{}.
\newblock \showarticletitle{Rev2: Fraudulent user prediction in rating
  platforms}. In \bibinfo{booktitle}{\emph{Proceedings of the Eleventh ACM
  International Conference on Web Search and Data Mining}}.
  \bibinfo{pages}{333--341}.
\newblock


\bibitem[\protect\citeauthoryear{Kumar, Spezzano, Subrahmanian, and
  Faloutsos}{Kumar et~al\mbox{.}}{2016}]%
        {kumar2016edge}
\bibfield{author}{\bibinfo{person}{Srijan Kumar}, \bibinfo{person}{Francesca
  Spezzano}, \bibinfo{person}{VS Subrahmanian}, {and} \bibinfo{person}{Christos
  Faloutsos}.} \bibinfo{year}{2016}\natexlab{}.
\newblock \showarticletitle{Edge weight prediction in weighted signed
  networks}. In \bibinfo{booktitle}{\emph{2016 IEEE 16th International
  Conference on Data Mining (ICDM)}}. IEEE, \bibinfo{pages}{221--230}.
\newblock


\bibitem[\protect\citeauthoryear{Lee, Eoff, and Caverlee}{Lee
  et~al\mbox{.}}{2011}]%
        {lee2011seven}
\bibfield{author}{\bibinfo{person}{Kyumin Lee}, \bibinfo{person}{Brian Eoff},
  {and} \bibinfo{person}{James Caverlee}.} \bibinfo{year}{2011}\natexlab{}.
\newblock \showarticletitle{Seven months with the devils: A long-term study of
  content polluters on twitter}. In \bibinfo{booktitle}{\emph{Proceedings of
  the international AAAI conference on web and social media}},
  Vol.~\bibinfo{volume}{5}. \bibinfo{pages}{185--192}.
\newblock


\bibitem[\protect\citeauthoryear{Li, Qin, Liu, Yang, and Li}{Li
  et~al\mbox{.}}{2019b}]%
        {li2019spam}
\bibfield{author}{\bibinfo{person}{Ao Li}, \bibinfo{person}{Zhou Qin},
  \bibinfo{person}{Runshi Liu}, \bibinfo{person}{Yiqun Yang}, {and}
  \bibinfo{person}{Dong Li}.} \bibinfo{year}{2019}\natexlab{b}.
\newblock \showarticletitle{Spam review detection with graph convolutional
  networks}. In \bibinfo{booktitle}{\emph{Proceedings of the 28th ACM
  International Conference on Information and Knowledge Management}}.
  \bibinfo{pages}{2703--2711}.
\newblock


\bibitem[\protect\citeauthoryear{Li, Pandey, Hooi, Faloutsos, and Pileggi}{Li
  et~al\mbox{.}}{2022}]%
        {li2020dynamic}
\bibfield{author}{\bibinfo{person}{Shimiao Li}, \bibinfo{person}{Amritanshu
  Pandey}, \bibinfo{person}{Bryan Hooi}, \bibinfo{person}{Christos Faloutsos},
  {and} \bibinfo{person}{Larry Pileggi}.} \bibinfo{year}{2022}\natexlab{}.
\newblock \showarticletitle{Dynamic Graph-Based Anomaly Detection in the
  Electrical Grid}.
\newblock \bibinfo{journal}{\emph{IEEE Transactions on Power Systems}}
  \bibinfo{volume}{37}, \bibinfo{number}{5} (\bibinfo{year}{2022}),
  \bibinfo{pages}{3408--3422}.
\newblock


\bibitem[\protect\citeauthoryear{Li, Huang, Li, Du, and Zou}{Li
  et~al\mbox{.}}{2019a}]%
        {li2019specae}
\bibfield{author}{\bibinfo{person}{Yuening Li}, \bibinfo{person}{Xiao Huang},
  \bibinfo{person}{Jundong Li}, \bibinfo{person}{Mengnan Du}, {and}
  \bibinfo{person}{Na Zou}.} \bibinfo{year}{2019}\natexlab{a}.
\newblock \showarticletitle{{SpecAE: Spectral} autoencoder for anomaly
  detection in attributed networks}. In \bibinfo{booktitle}{\emph{Proceedings
  of the 28th ACM International Conference on Information and Knowledge
  Management}}. \bibinfo{pages}{2233--2236}.
\newblock


\bibitem[\protect\citeauthoryear{Liu, Kong, Xia, Bai, Wang, Qing, and Lee}{Liu
  et~al\mbox{.}}{2018c}]%
        {liu2018artificial}
\bibfield{author}{\bibinfo{person}{Jiaying Liu}, \bibinfo{person}{Xiangjie
  Kong}, \bibinfo{person}{Feng Xia}, \bibinfo{person}{Xiaomei Bai},
  \bibinfo{person}{Lei Wang}, \bibinfo{person}{Qing Qing}, {and}
  \bibinfo{person}{Ivan Lee}.} \bibinfo{year}{2018}\natexlab{c}.
\newblock \showarticletitle{Artificial intelligence in the 21st century}.
\newblock \bibinfo{journal}{\emph{IEEE Access}}  \bibinfo{volume}{6}
  (\bibinfo{year}{2018}), \bibinfo{pages}{34403--34421}.
\newblock


\bibitem[\protect\citeauthoryear{Liu, Xia, Feng, Ren, and Liu}{Liu
  et~al\mbox{.}}{2022}]%
        {liu2022deep}
\bibfield{author}{\bibinfo{person}{Jiaying Liu}, \bibinfo{person}{Feng Xia},
  \bibinfo{person}{Xu Feng}, \bibinfo{person}{Jing Ren}, {and}
  \bibinfo{person}{Huan Liu}.} \bibinfo{year}{2022}\natexlab{}.
\newblock \showarticletitle{Deep Graph Learning for Anomalous Citation
  Detection}.
\newblock \bibinfo{journal}{\emph{IEEE Transactions on Neural Networks and
  Learning Systems}} \bibinfo{volume}{33}, \bibinfo{number}{6}
  (\bibinfo{year}{2022}), \bibinfo{pages}{2543--2557}.
\newblock


\bibitem[\protect\citeauthoryear{Liu, Dean, Rolf, Simchowitz, and Hardt}{Liu
  et~al\mbox{.}}{2018b}]%
        {liu2018delayed}
\bibfield{author}{\bibinfo{person}{Lydia~T Liu}, \bibinfo{person}{Sarah Dean},
  \bibinfo{person}{Esther Rolf}, \bibinfo{person}{Max Simchowitz}, {and}
  \bibinfo{person}{Moritz Hardt}.} \bibinfo{year}{2018}\natexlab{b}.
\newblock \showarticletitle{Delayed impact of fair machine learning}. In
  \bibinfo{booktitle}{\emph{International Conference on Machine Learning}}.
  PMLR, \bibinfo{pages}{3150--3158}.
\newblock


\bibitem[\protect\citeauthoryear{Liu, Li, Pan, Gong, Zhou, and Karypis}{Liu
  et~al\mbox{.}}{2021a}]%
        {liu2021anomaly}
\bibfield{author}{\bibinfo{person}{Yixin Liu}, \bibinfo{person}{Zhao Li},
  \bibinfo{person}{Shirui Pan}, \bibinfo{person}{Chen Gong},
  \bibinfo{person}{Chuan Zhou}, {and} \bibinfo{person}{George Karypis}.}
  \bibinfo{year}{2021}\natexlab{a}.
\newblock \showarticletitle{Anomaly detection on attributed networks via
  contrastive self-supervised learning}.
\newblock \bibinfo{journal}{\emph{IEEE transactions on neural networks and
  learning systems}} \bibinfo{volume}{33}, \bibinfo{number}{6}
  (\bibinfo{year}{2021}), \bibinfo{pages}{2378--2392}.
\newblock


\bibitem[\protect\citeauthoryear{Liu, Pan, Wang, Xiong, Wang, Chen, and
  Lee}{Liu et~al\mbox{.}}{2021b}]%
        {liu2021anomalyTKDE}
\bibfield{author}{\bibinfo{person}{Yixin Liu}, \bibinfo{person}{Shirui Pan},
  \bibinfo{person}{Yu~Guang Wang}, \bibinfo{person}{Fei Xiong},
  \bibinfo{person}{Liang Wang}, \bibinfo{person}{Qingfeng Chen}, {and}
  \bibinfo{person}{Vincent~CS Lee}.} \bibinfo{year}{2021}\natexlab{b}.
\newblock \showarticletitle{Anomaly detection in dynamic graphs via
  transformer}.
\newblock \bibinfo{journal}{\emph{IEEE Transactions on Knowledge and Data
  Engineering}} (\bibinfo{year}{2021}).
\newblock


\bibitem[\protect\citeauthoryear{Liu, Chen, Yang, Zhou, Li, and Song}{Liu
  et~al\mbox{.}}{2018a}]%
        {liu2018heterogeneous}
\bibfield{author}{\bibinfo{person}{Ziqi Liu}, \bibinfo{person}{Chaochao Chen},
  \bibinfo{person}{Xinxing Yang}, \bibinfo{person}{Jun Zhou},
  \bibinfo{person}{Xiaolong Li}, {and} \bibinfo{person}{Le Song}.}
  \bibinfo{year}{2018}\natexlab{a}.
\newblock \showarticletitle{Heterogeneous graph neural networks for malicious
  account detection}. In \bibinfo{booktitle}{\emph{Proceedings of the 27th ACM
  International Conference on Information and Knowledge Management}}.
  \bibinfo{pages}{2077--2085}.
\newblock


\bibitem[\protect\citeauthoryear{Lu and Li}{Lu and Li}{2020}]%
        {lu2020gcan}
\bibfield{author}{\bibinfo{person}{Yi-Ju Lu} {and} \bibinfo{person}{Cheng-Te
  Li}.} \bibinfo{year}{2020}\natexlab{}.
\newblock \showarticletitle{{GCAN}: Graph-aware Co-Attention Networks for
  Explainable Fake News Detection on Social Media}. In
  \bibinfo{booktitle}{\emph{Proceedings of the 58th Annual Meeting of the
  Association for Computational Linguistics}}. \bibinfo{pages}{505--514}.
\newblock


\bibitem[\protect\citeauthoryear{Luo, Liu, and Gao}{Luo et~al\mbox{.}}{2017}]%
        {luo2017revisit}
\bibfield{author}{\bibinfo{person}{Weixin Luo}, \bibinfo{person}{Wen Liu},
  {and} \bibinfo{person}{Shenghua Gao}.} \bibinfo{year}{2017}\natexlab{}.
\newblock \showarticletitle{A revisit of sparse coding based anomaly detection
  in stacked {RNN} framework}. In \bibinfo{booktitle}{\emph{Proceedings of the
  IEEE International Conference on Computer Vision}}.
  \bibinfo{pages}{341--349}.
\newblock


\bibitem[\protect\citeauthoryear{Ma, Gao, Mitra, Kwon, Jansen, Wong, and
  Cha}{Ma et~al\mbox{.}}{2016}]%
        {ma2016detecting}
\bibfield{author}{\bibinfo{person}{Jing Ma}, \bibinfo{person}{Wei Gao},
  \bibinfo{person}{Prasenjit Mitra}, \bibinfo{person}{Sejeong Kwon},
  \bibinfo{person}{Bernard~J Jansen}, \bibinfo{person}{Kam-Fai Wong}, {and}
  \bibinfo{person}{Meeyoung Cha}.} \bibinfo{year}{2016}\natexlab{}.
\newblock \showarticletitle{Detecting rumors from microblogs with recurrent
  neural networks}. In \bibinfo{booktitle}{\emph{25th International Joint
  Conference on Artificial Intelligence, IJCAI 2016}}. International Joint
  Conferences on Artificial Intelligence, \bibinfo{pages}{3818--3824}.
\newblock


\bibitem[\protect\citeauthoryear{Ma, Gao, and Wong}{Ma et~al\mbox{.}}{2017}]%
        {ma2017detect}
\bibfield{author}{\bibinfo{person}{Jing Ma}, \bibinfo{person}{Wei Gao}, {and}
  \bibinfo{person}{Kam-Fai Wong}.} \bibinfo{year}{2017}\natexlab{}.
\newblock \showarticletitle{Detect Rumors in Microblog Posts Using Propagation
  Structure via Kernel Learning}. In \bibinfo{booktitle}{\emph{Proceedings of
  the 55th Annual Meeting of the Association for Computational Linguistics
  (Volume 1: Long Papers)}}. \bibinfo{pages}{708--717}.
\newblock


\bibitem[\protect\citeauthoryear{Ma, Wu, Xue, Yang, Zhou, Sheng, Xiong, and
  Akoglu}{Ma et~al\mbox{.}}{2021}]%
        {ma2021comprehensive}
\bibfield{author}{\bibinfo{person}{Xiaoxiao Ma}, \bibinfo{person}{Jia Wu},
  \bibinfo{person}{Shan Xue}, \bibinfo{person}{Jian Yang},
  \bibinfo{person}{Chuan Zhou}, \bibinfo{person}{Quan~Z Sheng},
  \bibinfo{person}{Hui Xiong}, {and} \bibinfo{person}{Leman Akoglu}.}
  \bibinfo{year}{2021}\natexlab{}.
\newblock \showarticletitle{A comprehensive survey on graph anomaly detection
  with deep learning}.
\newblock \bibinfo{journal}{\emph{IEEE Transactions on Knowledge and Data
  Engineering}} (\bibinfo{year}{2021}).
\newblock
\urldef\tempurl%
\url{https://doi.org/10.1109/TKDE.2021.3118815}
\showDOI{\tempurl}


\bibitem[\protect\citeauthoryear{Markovitz, Sharir, Friedman, Zelnik-Manor, and
  Avidan}{Markovitz et~al\mbox{.}}{2020}]%
        {markovitz2020graph}
\bibfield{author}{\bibinfo{person}{Amir Markovitz}, \bibinfo{person}{Gilad
  Sharir}, \bibinfo{person}{Itamar Friedman}, \bibinfo{person}{Lihi
  Zelnik-Manor}, {and} \bibinfo{person}{Shai Avidan}.}
  \bibinfo{year}{2020}\natexlab{}.
\newblock \showarticletitle{Graph embedded pose clustering for anomaly
  detection}. In \bibinfo{booktitle}{\emph{Proceedings of the IEEE/CVF
  Conference on Computer Vision and Pattern Recognition}}.
  \bibinfo{pages}{10539--10547}.
\newblock


\bibitem[\protect\citeauthoryear{Mathur and Tippenhauer}{Mathur and
  Tippenhauer}{2016}]%
        {mathur2016swat}
\bibfield{author}{\bibinfo{person}{Aditya~P Mathur} {and}
  \bibinfo{person}{Nils~Ole Tippenhauer}.} \bibinfo{year}{2016}\natexlab{}.
\newblock \showarticletitle{{SWaT: A} water treatment testbed for research and
  training on ICS security}. In \bibinfo{booktitle}{\emph{2016 international
  workshop on cyber-physical systems for smart water networks (CySWater)}}.
  IEEE, \bibinfo{pages}{31--36}.
\newblock


\bibitem[\protect\citeauthoryear{McAuley and Leskovec}{McAuley and
  Leskovec}{2013}]%
        {mcauley2013amateurs}
\bibfield{author}{\bibinfo{person}{Julian~John McAuley} {and}
  \bibinfo{person}{Jure Leskovec}.} \bibinfo{year}{2013}\natexlab{}.
\newblock \showarticletitle{From amateurs to connoisseurs: {Modeling} the
  evolution of user expertise through online reviews}. In
  \bibinfo{booktitle}{\emph{Proceedings of the 22nd international conference on
  World Wide Web}}. \bibinfo{pages}{897--908}.
\newblock


\bibitem[\protect\citeauthoryear{Nguyen, Sugiyama, Nakov, and Kan}{Nguyen
  et~al\mbox{.}}{2020}]%
        {nguyen2020fang}
\bibfield{author}{\bibinfo{person}{Van-Hoang Nguyen}, \bibinfo{person}{Kazunari
  Sugiyama}, \bibinfo{person}{Preslav Nakov}, {and} \bibinfo{person}{Min-Yen
  Kan}.} \bibinfo{year}{2020}\natexlab{}.
\newblock \showarticletitle{Fang: Leveraging social context for fake news
  detection using graph representation}. In
  \bibinfo{booktitle}{\emph{Proceedings of the 29th ACM international
  conference on information \& knowledge management}}.
  \bibinfo{pages}{1165--1174}.
\newblock


\bibitem[\protect\citeauthoryear{Niepert, Ahmed, and Kutzkov}{Niepert
  et~al\mbox{.}}{2016}]%
        {niepert2016learning}
\bibfield{author}{\bibinfo{person}{Mathias Niepert}, \bibinfo{person}{Mohamed
  Ahmed}, {and} \bibinfo{person}{Konstantin Kutzkov}.}
  \bibinfo{year}{2016}\natexlab{}.
\newblock \showarticletitle{Learning convolutional neural networks for graphs}.
  In \bibinfo{booktitle}{\emph{International conference on machine learning}}.
  PMLR, \bibinfo{pages}{2014--2023}.
\newblock


\bibitem[\protect\citeauthoryear{Opsahl and Panzarasa}{Opsahl and
  Panzarasa}{2009}]%
        {opsahl2009clustering}
\bibfield{author}{\bibinfo{person}{Tore Opsahl} {and} \bibinfo{person}{Pietro
  Panzarasa}.} \bibinfo{year}{2009}\natexlab{}.
\newblock \showarticletitle{Clustering in weighted networks}.
\newblock \bibinfo{journal}{\emph{Social networks}} \bibinfo{volume}{31},
  \bibinfo{number}{2} (\bibinfo{year}{2009}), \bibinfo{pages}{155--163}.
\newblock


\bibitem[\protect\citeauthoryear{Pan, Hu, Long, Jiang, Yao, and Zhang}{Pan
  et~al\mbox{.}}{2018}]%
        {pan2018adversarially}
\bibfield{author}{\bibinfo{person}{Shirui Pan}, \bibinfo{person}{Ruiqi Hu},
  \bibinfo{person}{Guodong Long}, \bibinfo{person}{Jing Jiang},
  \bibinfo{person}{Lina Yao}, {and} \bibinfo{person}{Chengqi Zhang}.}
  \bibinfo{year}{2018}\natexlab{}.
\newblock \showarticletitle{Adversarially regularized graph autoencoder for
  graph embedding}. In \bibinfo{booktitle}{\emph{Proceedings of the 27th
  International Joint Conference on Artificial Intelligence}}.
  \bibinfo{pages}{2609--2615}.
\newblock


\bibitem[\protect\citeauthoryear{Pang, Shen, Cao, and Hengel}{Pang
  et~al\mbox{.}}{2021}]%
        {pang2021deep}
\bibfield{author}{\bibinfo{person}{Guansong Pang}, \bibinfo{person}{Chunhua
  Shen}, \bibinfo{person}{Longbing Cao}, {and} \bibinfo{person}{Anton Van~Den
  Hengel}.} \bibinfo{year}{2021}\natexlab{}.
\newblock \showarticletitle{Deep learning for anomaly detection: A review}.
\newblock \bibinfo{journal}{\emph{ACM Computing Surveys (CSUR)}}
  \bibinfo{volume}{54}, \bibinfo{number}{2} (\bibinfo{year}{2021}),
  \bibinfo{pages}{1--38}.
\newblock


\bibitem[\protect\citeauthoryear{Perozzi, Akoglu, Iglesias~S{\'a}nchez, and
  M{\"u}ller}{Perozzi et~al\mbox{.}}{2014}]%
        {perozzi2014focused}
\bibfield{author}{\bibinfo{person}{Bryan Perozzi}, \bibinfo{person}{Leman
  Akoglu}, \bibinfo{person}{Patricia Iglesias~S{\'a}nchez}, {and}
  \bibinfo{person}{Emmanuel M{\"u}ller}.} \bibinfo{year}{2014}\natexlab{}.
\newblock \showarticletitle{Focused clustering and outlier detection in large
  attributed graphs}. In \bibinfo{booktitle}{\emph{Proceedings of the 20th ACM
  SIGKDD international conference on Knowledge discovery and data mining}}.
  \bibinfo{pages}{1346--1355}.
\newblock


\bibitem[\protect\citeauthoryear{Rafiq, Hosseinmardi, Han, Lv, Mishra, and
  Mattson}{Rafiq et~al\mbox{.}}{2015}]%
        {rafiq2015careful}
\bibfield{author}{\bibinfo{person}{Rahat~Ibn Rafiq}, \bibinfo{person}{Homa
  Hosseinmardi}, \bibinfo{person}{Richard Han}, \bibinfo{person}{Qin Lv},
  \bibinfo{person}{Shivakant Mishra}, {and} \bibinfo{person}{Sabrina~Arredondo
  Mattson}.} \bibinfo{year}{2015}\natexlab{}.
\newblock \showarticletitle{Careful what you share in six seconds: Detecting
  cyberbullying instances in {Vine}}. In \bibinfo{booktitle}{\emph{2015
  IEEE/ACM International Conference on Advances in Social Networks Analysis and
  Mining (ASONAM)}}. IEEE, \bibinfo{pages}{617--622}.
\newblock


\bibitem[\protect\citeauthoryear{Ranshous, Shen, Koutra, Harenberg, Faloutsos,
  and Samatova}{Ranshous et~al\mbox{.}}{2015}]%
        {ranshous2015anomaly}
\bibfield{author}{\bibinfo{person}{Stephen Ranshous}, \bibinfo{person}{Shitian
  Shen}, \bibinfo{person}{Danai Koutra}, \bibinfo{person}{Steve Harenberg},
  \bibinfo{person}{Christos Faloutsos}, {and} \bibinfo{person}{Nagiza~F
  Samatova}.} \bibinfo{year}{2015}\natexlab{}.
\newblock \showarticletitle{Anomaly detection in dynamic networks: {A} survey}.
\newblock \bibinfo{journal}{\emph{Wiley Interdisciplinary Reviews:
  Computational Statistics}} \bibinfo{volume}{7}, \bibinfo{number}{3}
  (\bibinfo{year}{2015}), \bibinfo{pages}{223--247}.
\newblock


\bibitem[\protect\citeauthoryear{Rayana and Akoglu}{Rayana and Akoglu}{2015}]%
        {rayana2015collective}
\bibfield{author}{\bibinfo{person}{Shebuti Rayana} {and} \bibinfo{person}{Leman
  Akoglu}.} \bibinfo{year}{2015}\natexlab{}.
\newblock \showarticletitle{Collective opinion spam detection: Bridging review
  networks and metadata}. In \bibinfo{booktitle}{\emph{Proceedings of the 21th
  acm sigkdd international conference on knowledge discovery and data mining}}.
  \bibinfo{pages}{985--994}.
\newblock


\bibitem[\protect\citeauthoryear{Ren, Xia, Chen, Liu, Hou, Shehzad, Sultanova,
  and Kong}{Ren et~al\mbox{.}}{2021}]%
        {ren2021matching}
\bibfield{author}{\bibinfo{person}{Jing Ren}, \bibinfo{person}{Feng Xia},
  \bibinfo{person}{Xiangtai Chen}, \bibinfo{person}{Jiaying Liu},
  \bibinfo{person}{Mingliang Hou}, \bibinfo{person}{Ahsan Shehzad},
  \bibinfo{person}{Nargiz Sultanova}, {and} \bibinfo{person}{Xiangjie Kong}.}
  \bibinfo{year}{2021}\natexlab{}.
\newblock \showarticletitle{Matching Algorithms: Fundamentals, Applications and
  Challenges}.
\newblock \bibinfo{journal}{\emph{IEEE Transactions on Emerging Topics in
  Computational Intelligence}} \bibinfo{volume}{5}, \bibinfo{number}{3}
  (\bibinfo{year}{2021}), \bibinfo{pages}{332--350}.
\newblock


\bibitem[\protect\citeauthoryear{Ren, Wang, Zhang, and Chang}{Ren
  et~al\mbox{.}}{2020}]%
        {ren2020adversarial}
\bibfield{author}{\bibinfo{person}{Yuxiang Ren}, \bibinfo{person}{Bo Wang},
  \bibinfo{person}{Jiawei Zhang}, {and} \bibinfo{person}{Yi Chang}.}
  \bibinfo{year}{2020}\natexlab{}.
\newblock \showarticletitle{Adversarial Active Learning based Heterogeneous
  Graph Neural Network for Fake News Detection}. In
  \bibinfo{booktitle}{\emph{2020 IEEE International Conference on Data Mining
  (ICDM)}}. IEEE, \bibinfo{pages}{452--461}.
\newblock


\bibitem[\protect\citeauthoryear{Salawu, He, and Lumsden}{Salawu
  et~al\mbox{.}}{2017}]%
        {salawu2017approaches}
\bibfield{author}{\bibinfo{person}{Semiu Salawu}, \bibinfo{person}{Yulan He},
  {and} \bibinfo{person}{Joanna Lumsden}.} \bibinfo{year}{2017}\natexlab{}.
\newblock \showarticletitle{Approaches to automated detection of cyberbullying:
  A survey}.
\newblock \bibinfo{journal}{\emph{IEEE Transactions on Affective Computing}}
  \bibinfo{volume}{11}, \bibinfo{number}{1} (\bibinfo{year}{2017}),
  \bibinfo{pages}{3--24}.
\newblock


\bibitem[\protect\citeauthoryear{Sen, Namata, Bilgic, Getoor, Galligher, and
  Eliassi-Rad}{Sen et~al\mbox{.}}{2008}]%
        {sen2008collective}
\bibfield{author}{\bibinfo{person}{Prithviraj Sen}, \bibinfo{person}{Galileo
  Namata}, \bibinfo{person}{Mustafa Bilgic}, \bibinfo{person}{Lise Getoor},
  \bibinfo{person}{Brian Galligher}, {and} \bibinfo{person}{Tina Eliassi-Rad}.}
  \bibinfo{year}{2008}\natexlab{}.
\newblock \showarticletitle{Collective classification in network data}.
\newblock \bibinfo{journal}{\emph{AI magazine}} \bibinfo{volume}{29},
  \bibinfo{number}{3} (\bibinfo{year}{2008}), \bibinfo{pages}{93--93}.
\newblock


\bibitem[\protect\citeauthoryear{Shekhar, Shah, and Akoglu}{Shekhar
  et~al\mbox{.}}{2021}]%
        {shekhar2021fairod}
\bibfield{author}{\bibinfo{person}{Shubhranshu Shekhar}, \bibinfo{person}{Neil
  Shah}, {and} \bibinfo{person}{Leman Akoglu}.}
  \bibinfo{year}{2021}\natexlab{}.
\newblock \showarticletitle{Fairod: Fairness-aware outlier detection}. In
  \bibinfo{booktitle}{\emph{Proceedings of the 2021 AAAI/ACM Conference on AI,
  Ethics, and Society}}. \bibinfo{pages}{210--220}.
\newblock


\bibitem[\protect\citeauthoryear{Shi, Li, Zhang, Sun, and Philip}{Shi
  et~al\mbox{.}}{2016}]%
        {shi2016survey}
\bibfield{author}{\bibinfo{person}{Chuan Shi}, \bibinfo{person}{Yitong Li},
  \bibinfo{person}{Jiawei Zhang}, \bibinfo{person}{Yizhou Sun}, {and}
  \bibinfo{person}{S~Yu Philip}.} \bibinfo{year}{2016}\natexlab{}.
\newblock \showarticletitle{A survey of heterogeneous information network
  analysis}.
\newblock \bibinfo{journal}{\emph{IEEE Transactions on Knowledge and Data
  Engineering}} \bibinfo{volume}{29}, \bibinfo{number}{1}
  (\bibinfo{year}{2016}), \bibinfo{pages}{17--37}.
\newblock


\bibitem[\protect\citeauthoryear{Shu, Sliva, Wang, Tang, and Liu}{Shu
  et~al\mbox{.}}{2017}]%
        {shu2017fake}
\bibfield{author}{\bibinfo{person}{Kai Shu}, \bibinfo{person}{Amy Sliva},
  \bibinfo{person}{Suhang Wang}, \bibinfo{person}{Jiliang Tang}, {and}
  \bibinfo{person}{Huan Liu}.} \bibinfo{year}{2017}\natexlab{}.
\newblock \showarticletitle{Fake news detection on social media: A data mining
  perspective}.
\newblock \bibinfo{journal}{\emph{ACM SIGKDD explorations newsletter}}
  \bibinfo{volume}{19}, \bibinfo{number}{1} (\bibinfo{year}{2017}),
  \bibinfo{pages}{22--36}.
\newblock


\bibitem[\protect\citeauthoryear{Squicciarini, Rajtmajer, Liu, and
  Griffin}{Squicciarini et~al\mbox{.}}{2015}]%
        {squicciarini2015identification}
\bibfield{author}{\bibinfo{person}{Anna Squicciarini}, \bibinfo{person}{Sarah
  Rajtmajer}, \bibinfo{person}{Y Liu}, {and} \bibinfo{person}{Christopher
  Griffin}.} \bibinfo{year}{2015}\natexlab{}.
\newblock \showarticletitle{Identification and characterization of
  cyberbullying dynamics in an online social network}. In
  \bibinfo{booktitle}{\emph{Proceedings of the 2015 IEEE/ACM International
  Conference on Advances in Social Networks Analysis and Mining 2015}}.
  \bibinfo{pages}{280--285}.
\newblock


\bibitem[\protect\citeauthoryear{Tang, Zhang, Yao, Li, Zhang, and Su}{Tang
  et~al\mbox{.}}{2008}]%
        {tang2008arnetminer}
\bibfield{author}{\bibinfo{person}{Jie Tang}, \bibinfo{person}{Jing Zhang},
  \bibinfo{person}{Limin Yao}, \bibinfo{person}{Juanzi Li}, \bibinfo{person}{Li
  Zhang}, {and} \bibinfo{person}{Zhong Su}.} \bibinfo{year}{2008}\natexlab{}.
\newblock \showarticletitle{{ArnetMiner}: extraction and mining of academic
  social networks}. In \bibinfo{booktitle}{\emph{Proceedings of the 14th ACM
  SIGKDD international conference on Knowledge discovery and data mining}}.
  \bibinfo{pages}{990--998}.
\newblock


\bibitem[\protect\citeauthoryear{Tang and Liu}{Tang and Liu}{2009}]%
        {tang2009relational}
\bibfield{author}{\bibinfo{person}{Lei Tang} {and} \bibinfo{person}{Huan Liu}.}
  \bibinfo{year}{2009}\natexlab{}.
\newblock \showarticletitle{Relational learning via latent social dimensions}.
  In \bibinfo{booktitle}{\emph{Proceedings of the 15th ACM SIGKDD international
  conference on Knowledge discovery and data mining}}.
  \bibinfo{pages}{817--826}.
\newblock


\bibitem[\protect\citeauthoryear{Teng, Yan, Ertugrul, and Lin}{Teng
  et~al\mbox{.}}{2018}]%
        {teng2018deep}
\bibfield{author}{\bibinfo{person}{Xian Teng}, \bibinfo{person}{Muheng Yan},
  \bibinfo{person}{Ali~Mert Ertugrul}, {and} \bibinfo{person}{Yu-Ru Lin}.}
  \bibinfo{year}{2018}\natexlab{}.
\newblock \showarticletitle{Deep into hypersphere: robust and unsupervised
  anomaly discovery in dynamic networks}. In
  \bibinfo{booktitle}{\emph{Proceedings of the 27th International Joint
  Conference on Artificial Intelligence}}. \bibinfo{pages}{2724--2730}.
\newblock


\bibitem[\protect\citeauthoryear{Tian, Gao, Cui, Chen, and Liu}{Tian
  et~al\mbox{.}}{2014}]%
        {tian2014learning}
\bibfield{author}{\bibinfo{person}{Fei Tian}, \bibinfo{person}{Bin Gao},
  \bibinfo{person}{Qing Cui}, \bibinfo{person}{Enhong Chen}, {and}
  \bibinfo{person}{Tie-Yan Liu}.} \bibinfo{year}{2014}\natexlab{}.
\newblock \showarticletitle{Learning deep representations for graph
  clustering}. In \bibinfo{booktitle}{\emph{Proceedings of the Twenty-Eighth
  AAAI Conference on Artificial Intelligence}}, Vol.~\bibinfo{volume}{28}.
  \bibinfo{pages}{1293--1299}.
\newblock


\bibitem[\protect\citeauthoryear{Veli{\v{c}}kovi{\'c}, Cucurull, Casanova,
  Romero, Lio, and Bengio}{Veli{\v{c}}kovi{\'c} et~al\mbox{.}}{2018}]%
        {velivckovic2017graph}
\bibfield{author}{\bibinfo{person}{Petar Veli{\v{c}}kovi{\'c}},
  \bibinfo{person}{Guillem Cucurull}, \bibinfo{person}{Arantxa Casanova},
  \bibinfo{person}{Adriana Romero}, \bibinfo{person}{Pietro Lio}, {and}
  \bibinfo{person}{Yoshua Bengio}.} \bibinfo{year}{2018}\natexlab{}.
\newblock \showarticletitle{Graph attention networks}. In
  \bibinfo{booktitle}{\emph{Proceedings of the 7th International Conference on
  Learning Representations}}.
\newblock


\bibitem[\protect\citeauthoryear{Vosoughi, Roy, and Aral}{Vosoughi
  et~al\mbox{.}}{2018}]%
        {vosoughi2018spread}
\bibfield{author}{\bibinfo{person}{Soroush Vosoughi}, \bibinfo{person}{Deb
  Roy}, {and} \bibinfo{person}{Sinan Aral}.} \bibinfo{year}{2018}\natexlab{}.
\newblock \showarticletitle{The spread of true and false news online}.
\newblock \bibinfo{journal}{\emph{Science}} \bibinfo{volume}{359},
  \bibinfo{number}{6380} (\bibinfo{year}{2018}), \bibinfo{pages}{1146--1151}.
\newblock


\bibitem[\protect\citeauthoryear{Wan, Yuan, Xia, and Liu}{Wan
  et~al\mbox{.}}{2019}]%
        {wan2019your}
\bibfield{author}{\bibinfo{person}{Liangtian Wan}, \bibinfo{person}{Yuyuan
  Yuan}, \bibinfo{person}{Feng Xia}, {and} \bibinfo{person}{Huan Liu}.}
  \bibinfo{year}{2019}\natexlab{}.
\newblock \showarticletitle{To your surprise: Identifying serendipitous
  collaborators}.
\newblock \bibinfo{journal}{\emph{IEEE Transactions on Big Data}}
  \bibinfo{volume}{7}, \bibinfo{number}{3} (\bibinfo{year}{2019}),
  \bibinfo{pages}{574--589}.
\newblock


\bibitem[\protect\citeauthoryear{Wang, Lin, Cui, Jia, Wang, Fang, Yu, Zhou,
  Yang, and Qi}{Wang et~al\mbox{.}}{2019d}]%
        {wang2019semi}
\bibfield{author}{\bibinfo{person}{Daixin Wang}, \bibinfo{person}{Jianbin Lin},
  \bibinfo{person}{Peng Cui}, \bibinfo{person}{Quanhui Jia},
  \bibinfo{person}{Zhen Wang}, \bibinfo{person}{Yanming Fang},
  \bibinfo{person}{Quan Yu}, \bibinfo{person}{Jun Zhou},
  \bibinfo{person}{Shuang Yang}, {and} \bibinfo{person}{Yuan Qi}.}
  \bibinfo{year}{2019}\natexlab{d}.
\newblock \showarticletitle{A semi-supervised graph attentive network for
  financial fraud detection}. In \bibinfo{booktitle}{\emph{2019 IEEE
  International Conference on Data Mining (ICDM)}}. IEEE,
  \bibinfo{pages}{598--607}.
\newblock


\bibitem[\protect\citeauthoryear{Wang, Wang, Wang, Zhao, Zhang, Zhang, Li, Xie,
  and Guo}{Wang et~al\mbox{.}}{2019e}]%
        {wang2019learning}
\bibfield{author}{\bibinfo{person}{Hongwei Wang}, \bibinfo{person}{Jialin
  Wang}, \bibinfo{person}{Jia Wang}, \bibinfo{person}{Miao Zhao},
  \bibinfo{person}{Weinan Zhang}, \bibinfo{person}{Fuzheng Zhang},
  \bibinfo{person}{Wenjie Li}, \bibinfo{person}{Xing Xie}, {and}
  \bibinfo{person}{Minyi Guo}.} \bibinfo{year}{2019}\natexlab{e}.
\newblock \showarticletitle{Learning graph representation with generative
  adversarial nets}.
\newblock \bibinfo{journal}{\emph{IEEE Transactions on Knowledge and Data
  Engineering}} \bibinfo{volume}{33}, \bibinfo{number}{8}
  (\bibinfo{year}{2019}), \bibinfo{pages}{3090--3103}.
\newblock


\bibitem[\protect\citeauthoryear{Wang, Zhang, Wei, Hu, Piao, and Yin}{Wang
  et~al\mbox{.}}{2021c}]%
        {wang2021metro}
\bibfield{author}{\bibinfo{person}{Jingcheng Wang}, \bibinfo{person}{Yong
  Zhang}, \bibinfo{person}{Yun Wei}, \bibinfo{person}{Yongli Hu},
  \bibinfo{person}{Xinglin Piao}, {and} \bibinfo{person}{Baocai Yin}.}
  \bibinfo{year}{2021}\natexlab{c}.
\newblock \showarticletitle{Metro passenger flow prediction via dynamic
  hypergraph convolution networks}.
\newblock \bibinfo{journal}{\emph{IEEE Transactions on Intelligent
  Transportation Systems}} \bibinfo{volume}{22}, \bibinfo{number}{12}
  (\bibinfo{year}{2021}), \bibinfo{pages}{7891--7903}.
\newblock


\bibitem[\protect\citeauthoryear{Wang, Zhou, Wang, Wang, Fang, and Wang}{Wang
  et~al\mbox{.}}{2022}]%
        {wang2022a2djp}
\bibfield{author}{\bibinfo{person}{Kun Wang}, \bibinfo{person}{Zhengyang Zhou},
  \bibinfo{person}{Xu Wang}, \bibinfo{person}{Pengkun Wang},
  \bibinfo{person}{Qi Fang}, {and} \bibinfo{person}{Yang Wang}.}
  \bibinfo{year}{2022}\natexlab{}.
\newblock \showarticletitle{A2DJP: A Two Graph-based Component Fused Learning
  Framework for Urban Anomaly Distribution and Duration Joint-Prediction}.
\newblock \bibinfo{journal}{\emph{IEEE Transactions on Knowledge and Data
  Engineering}} (\bibinfo{year}{2022}).
\newblock
\urldef\tempurl%
\url{https://doi.org/10.1109/TKDE.2022.3176650}
\showDOI{\tempurl}


\bibitem[\protect\citeauthoryear{Wang, El-Fiqi, Hu, and Abbass}{Wang
  et~al\mbox{.}}{2019b}]%
        {wang2019convolutional}
\bibfield{author}{\bibinfo{person}{Min Wang}, \bibinfo{person}{Heba El-Fiqi},
  \bibinfo{person}{Jiankun Hu}, {and} \bibinfo{person}{Hussein~A Abbass}.}
  \bibinfo{year}{2019}\natexlab{b}.
\newblock \showarticletitle{Convolutional neural networks using dynamic
  functional connectivity for EEG-based person identification in diverse human
  states}.
\newblock \bibinfo{journal}{\emph{IEEE Transactions on Information Forensics
  and Security}} \bibinfo{volume}{14}, \bibinfo{number}{12}
  (\bibinfo{year}{2019}), \bibinfo{pages}{3259--3272}.
\newblock


\bibitem[\protect\citeauthoryear{Wang, Chen, Yu, Li, Ni, Tang, Gui, Li, Chen,
  and Yu}{Wang et~al\mbox{.}}{2019a}]%
        {wang2019heterogeneousgraph}
\bibfield{author}{\bibinfo{person}{Shen Wang}, \bibinfo{person}{Zhengzhang
  Chen}, \bibinfo{person}{Xiao Yu}, \bibinfo{person}{Ding Li},
  \bibinfo{person}{Jingchao Ni}, \bibinfo{person}{Lu-An Tang},
  \bibinfo{person}{Jiaping Gui}, \bibinfo{person}{Zhichun Li},
  \bibinfo{person}{Haifeng Chen}, {and} \bibinfo{person}{Philip~S Yu}.}
  \bibinfo{year}{2019}\natexlab{a}.
\newblock \showarticletitle{Heterogeneous Graph Matching Networks for Unknown
  Malware Detection}. In \bibinfo{booktitle}{\emph{28th International Joint
  Conference on Artificial Intelligence, IJCAI 2019}}. International Joint
  Conferences on Artificial Intelligence, \bibinfo{pages}{3762--3770}.
\newblock


\bibitem[\protect\citeauthoryear{Wang, Ji, Shi, Wang, Ye, Cui, and Yu}{Wang
  et~al\mbox{.}}{2019c}]%
        {wang2019heterogeneous}
\bibfield{author}{\bibinfo{person}{Xiao Wang}, \bibinfo{person}{Houye Ji},
  \bibinfo{person}{Chuan Shi}, \bibinfo{person}{Bai Wang},
  \bibinfo{person}{Yanfang Ye}, \bibinfo{person}{Peng Cui}, {and}
  \bibinfo{person}{Philip~S Yu}.} \bibinfo{year}{2019}\natexlab{c}.
\newblock \showarticletitle{Heterogeneous graph attention network}. In
  \bibinfo{booktitle}{\emph{The world wide web conference}}.
  \bibinfo{pages}{2022--2032}.
\newblock


\bibitem[\protect\citeauthoryear{Wang, Jin, Du, Cui, Tan, and Yang}{Wang
  et~al\mbox{.}}{2021a}]%
        {wang2021one}
\bibfield{author}{\bibinfo{person}{Xuhong Wang}, \bibinfo{person}{Baihong Jin},
  \bibinfo{person}{Ying Du}, \bibinfo{person}{Ping Cui},
  \bibinfo{person}{Yingshui Tan}, {and} \bibinfo{person}{Yupu Yang}.}
  \bibinfo{year}{2021}\natexlab{a}.
\newblock \showarticletitle{One-class graph neural networks for anomaly
  detection in attributed networks}.
\newblock \bibinfo{journal}{\emph{Neural computing and applications}}
  \bibinfo{volume}{33}, \bibinfo{number}{18} (\bibinfo{year}{2021}),
  \bibinfo{pages}{12073--12085}.
\newblock


\bibitem[\protect\citeauthoryear{Wang, Tang, Gao, and Liu}{Wang
  et~al\mbox{.}}{2010}]%
        {wang2010discovering}
\bibfield{author}{\bibinfo{person}{Xufei Wang}, \bibinfo{person}{Lei Tang},
  \bibinfo{person}{Huiji Gao}, {and} \bibinfo{person}{Huan Liu}.}
  \bibinfo{year}{2010}\natexlab{}.
\newblock \showarticletitle{Discovering overlapping groups in social media}. In
  \bibinfo{booktitle}{\emph{2010 IEEE international conference on data
  mining}}. IEEE, \bibinfo{pages}{569--578}.
\newblock


\bibitem[\protect\citeauthoryear{Wang, Ge, Li, and Chang}{Wang
  et~al\mbox{.}}{2020}]%
        {wang2020deep}
\bibfield{author}{\bibinfo{person}{Yaqian Wang}, \bibinfo{person}{Liang Ge},
  \bibinfo{person}{Siyu Li}, {and} \bibinfo{person}{Feng Chang}.}
  \bibinfo{year}{2020}\natexlab{}.
\newblock \showarticletitle{Deep Temporal Multi-Graph Convolutional Network for
  Crime Prediction}. In \bibinfo{booktitle}{\emph{International Conference on
  Conceptual Modeling}}. Springer, \bibinfo{pages}{525--538}.
\newblock


\bibitem[\protect\citeauthoryear{Wang, Zhang, Guo, Yin, Li, and Chen}{Wang
  et~al\mbox{.}}{2021b}]%
        {wang2021decoupling}
\bibfield{author}{\bibinfo{person}{Yanling Wang}, \bibinfo{person}{Jing Zhang},
  \bibinfo{person}{Shasha Guo}, \bibinfo{person}{Hongzhi Yin},
  \bibinfo{person}{Cuiping Li}, {and} \bibinfo{person}{Hong Chen}.}
  \bibinfo{year}{2021}\natexlab{b}.
\newblock \showarticletitle{Decoupling representation learning and
  classification for gnn-based anomaly detection}. In
  \bibinfo{booktitle}{\emph{Proceedings of the 44th International ACM SIGIR
  Conference on Research and Development in Information Retrieval}}.
  \bibinfo{pages}{1239--1248}.
\newblock


\bibitem[\protect\citeauthoryear{Wang, Wang, Pei, and Ye}{Wang
  et~al\mbox{.}}{2017}]%
        {wang2017multiple}
\bibfield{author}{\bibinfo{person}{Zheng Wang}, \bibinfo{person}{Chaokun Wang},
  \bibinfo{person}{Jisheng Pei}, {and} \bibinfo{person}{Xiaojun Ye}.}
  \bibinfo{year}{2017}\natexlab{}.
\newblock \showarticletitle{Multiple source detection without knowing the
  underlying propagation model}. In \bibinfo{booktitle}{\emph{Proceedings of
  the Thirty-First AAAI Conference on Artificial Intelligence}}.
  \bibinfo{pages}{217--223}.
\newblock


\bibitem[\protect\citeauthoryear{Welling and Kipf}{Welling and Kipf}{2017}]%
        {welling2017semi}
\bibfield{author}{\bibinfo{person}{Max Welling} {and} \bibinfo{person}{Thomas~N
  Kipf}.} \bibinfo{year}{2017}\natexlab{}.
\newblock \showarticletitle{Semi-supervised classification with graph
  convolutional networks}. In \bibinfo{booktitle}{\emph{J. International
  Conference on Learning Representations (ICLR 2017)}}.
\newblock


\bibitem[\protect\citeauthoryear{Wu, Lian, Xu, Wu, and Chen}{Wu
  et~al\mbox{.}}{2020a}]%
        {wu2020graph}
\bibfield{author}{\bibinfo{person}{Yongji Wu}, \bibinfo{person}{Defu Lian},
  \bibinfo{person}{Yiheng Xu}, \bibinfo{person}{Le Wu}, {and}
  \bibinfo{person}{Enhong Chen}.} \bibinfo{year}{2020}\natexlab{a}.
\newblock \showarticletitle{Graph convolutional networks with markov random
  field reasoning for social spammer detection}. In
  \bibinfo{booktitle}{\emph{Proceedings of the AAAI Conference on Artificial
  Intelligence}}, Vol.~\bibinfo{volume}{34}. \bibinfo{pages}{1054--1061}.
\newblock


\bibitem[\protect\citeauthoryear{Wu, Pan, Chen, Long, Zhang, and Philip}{Wu
  et~al\mbox{.}}{2020b}]%
        {wu2020comprehensive}
\bibfield{author}{\bibinfo{person}{Zonghan Wu}, \bibinfo{person}{Shirui Pan},
  \bibinfo{person}{Fengwen Chen}, \bibinfo{person}{Guodong Long},
  \bibinfo{person}{Chengqi Zhang}, {and} \bibinfo{person}{S~Yu Philip}.}
  \bibinfo{year}{2020}\natexlab{b}.
\newblock \showarticletitle{A comprehensive survey on graph neural networks}.
\newblock \bibinfo{journal}{\emph{IEEE transactions on neural networks and
  learning systems}} \bibinfo{volume}{32}, \bibinfo{number}{1}
  (\bibinfo{year}{2020}), \bibinfo{pages}{4--24}.
\newblock


\bibitem[\protect\citeauthoryear{Xia, Jedari, Yang, Ma, and Huang}{Xia
  et~al\mbox{.}}{2016}]%
        {xia2016signaling}
\bibfield{author}{\bibinfo{person}{Feng Xia}, \bibinfo{person}{Behrouz Jedari},
  \bibinfo{person}{Laurence~Tianruo Yang}, \bibinfo{person}{Jianhua Ma}, {and}
  \bibinfo{person}{Runhe Huang}.} \bibinfo{year}{2016}\natexlab{}.
\newblock \showarticletitle{A signaling game for uncertain data delivery in
  selfish mobile social networks}.
\newblock \bibinfo{journal}{\emph{IEEE Transactions on Computational Social
  Systems}} \bibinfo{volume}{3}, \bibinfo{number}{2} (\bibinfo{year}{2016}),
  \bibinfo{pages}{100--112}.
\newblock


\bibitem[\protect\citeauthoryear{Xia, Liu, Nie, Fu, Wan, and Kong}{Xia
  et~al\mbox{.}}{2019}]%
        {xia2019random}
\bibfield{author}{\bibinfo{person}{Feng Xia}, \bibinfo{person}{Jiaying Liu},
  \bibinfo{person}{Hansong Nie}, \bibinfo{person}{Yonghao Fu},
  \bibinfo{person}{Liangtian Wan}, {and} \bibinfo{person}{Xiangjie Kong}.}
  \bibinfo{year}{2019}\natexlab{}.
\newblock \showarticletitle{Random walks: A review of algorithms and
  applications}.
\newblock \bibinfo{journal}{\emph{IEEE Transactions on Emerging Topics in
  Computational Intelligence}} \bibinfo{volume}{4}, \bibinfo{number}{2}
  (\bibinfo{year}{2019}), \bibinfo{pages}{95--107}.
\newblock


\bibitem[\protect\citeauthoryear{Xia, Sun, Yu, Aziz, Wan, Pan, and Liu}{Xia
  et~al\mbox{.}}{2021b}]%
        {xia2021graph}
\bibfield{author}{\bibinfo{person}{Feng Xia}, \bibinfo{person}{Ke Sun},
  \bibinfo{person}{Shuo Yu}, \bibinfo{person}{Abdul Aziz},
  \bibinfo{person}{Liangtian Wan}, \bibinfo{person}{Shirui Pan}, {and}
  \bibinfo{person}{Huan Liu}.} \bibinfo{year}{2021}\natexlab{b}.
\newblock \showarticletitle{Graph Learning: A Survey}.
\newblock \bibinfo{journal}{\emph{IEEE Transactions on Artificial
  Intelligence}} \bibinfo{volume}{2}, \bibinfo{number}{2}
  (\bibinfo{year}{2021}), \bibinfo{pages}{109--127}.
\newblock


\bibitem[\protect\citeauthoryear{Xia, Wang, Bekele, and Liu}{Xia
  et~al\mbox{.}}{2017}]%
        {xia2017big}
\bibfield{author}{\bibinfo{person}{Feng Xia}, \bibinfo{person}{Wei Wang},
  \bibinfo{person}{Teshome~Megersa Bekele}, {and} \bibinfo{person}{Huan Liu}.}
  \bibinfo{year}{2017}\natexlab{}.
\newblock \showarticletitle{Big scholarly data: A survey}.
\newblock \bibinfo{journal}{\emph{IEEE Transactions on Big Data}}
  \bibinfo{volume}{3}, \bibinfo{number}{1} (\bibinfo{year}{2017}),
  \bibinfo{pages}{18--35}.
\newblock


\bibitem[\protect\citeauthoryear{Xia, Huang, Xu, Dai, Bo, Zhang, and Chen}{Xia
  et~al\mbox{.}}{2021a}]%
        {xia2021spatial}
\bibfield{author}{\bibinfo{person}{Lianghao Xia}, \bibinfo{person}{Chao Huang},
  \bibinfo{person}{Yong Xu}, \bibinfo{person}{Peng Dai},
  \bibinfo{person}{Liefeng Bo}, \bibinfo{person}{Xiyue Zhang}, {and}
  \bibinfo{person}{Tianyi Chen}.} \bibinfo{year}{2021}\natexlab{a}.
\newblock \showarticletitle{Spatial-Temporal Sequential Hypergraph Network for
  Crime Prediction with Dynamic Multiplex Relation Learning}. In
  \bibinfo{booktitle}{\emph{IJCAI}}. \bibinfo{pages}{1631--1637}.
\newblock


\bibitem[\protect\citeauthoryear{Xie, Guo, Chen, Xiao, Wang, and Zhao}{Xie
  et~al\mbox{.}}{2020}]%
        {xie2020deep}
\bibfield{author}{\bibinfo{person}{Qinge Xie}, \bibinfo{person}{Tiancheng Guo},
  \bibinfo{person}{Yang Chen}, \bibinfo{person}{Yu Xiao}, \bibinfo{person}{Xin
  Wang}, {and} \bibinfo{person}{Ben~Y Zhao}.} \bibinfo{year}{2020}\natexlab{}.
\newblock \showarticletitle{Deep graph convolutional networks for
  incident-driven traffic speed prediction}. In
  \bibinfo{booktitle}{\emph{Proceedings of the 29th ACM International
  Conference on Information \& Knowledge Management}}.
  \bibinfo{pages}{1665--1674}.
\newblock


\bibitem[\protect\citeauthoryear{Yang, Harkreader, Zhang, Shin, and Gu}{Yang
  et~al\mbox{.}}{2012}]%
        {yang2012analyzing}
\bibfield{author}{\bibinfo{person}{Chao Yang}, \bibinfo{person}{Robert
  Harkreader}, \bibinfo{person}{Jialong Zhang}, \bibinfo{person}{Seungwon
  Shin}, {and} \bibinfo{person}{Guofei Gu}.} \bibinfo{year}{2012}\natexlab{}.
\newblock \showarticletitle{Analyzing spammers' social networks for fun and
  profit: a case study of cybercriminal ecosystem on {Twitter}}. In
  \bibinfo{booktitle}{\emph{Proceedings of the 21st international conference on
  World Wide Web}}. \bibinfo{pages}{71--80}.
\newblock


\bibitem[\protect\citeauthoryear{Ying, Bourgeois, You, Zitnik, and
  Leskovec}{Ying et~al\mbox{.}}{2019}]%
        {ying2019gnnexplainer}
\bibfield{author}{\bibinfo{person}{Rex Ying}, \bibinfo{person}{Dylan
  Bourgeois}, \bibinfo{person}{Jiaxuan You}, \bibinfo{person}{Marinka Zitnik},
  {and} \bibinfo{person}{Jure Leskovec}.} \bibinfo{year}{2019}\natexlab{}.
\newblock \showarticletitle{{GNNExplainer}: Generating explanations for graph
  neural networks}.
\newblock \bibinfo{journal}{\emph{Advances in neural information processing
  systems}}  \bibinfo{volume}{32} (\bibinfo{year}{2019}),
  \bibinfo{pages}{9240}.
\newblock


\bibitem[\protect\citeauthoryear{Yu, Liu, Wei, Xia, and Tong}{Yu
  et~al\mbox{.}}{2020a}]%
        {yu2020optimize}
\bibfield{author}{\bibinfo{person}{Shuo Yu}, \bibinfo{person}{Jiaying Liu},
  \bibinfo{person}{Haoran Wei}, \bibinfo{person}{Feng Xia}, {and}
  \bibinfo{person}{Hanghang Tong}.} \bibinfo{year}{2020}\natexlab{a}.
\newblock \showarticletitle{How to optimize an academic team when the outlier
  member is leaving?}
\newblock \bibinfo{journal}{\emph{IEEE Intelligent Systems}}
  \bibinfo{volume}{36}, \bibinfo{number}{3} (\bibinfo{year}{2020}),
  \bibinfo{pages}{23--30}.
\newblock


\bibitem[\protect\citeauthoryear{Yu, Ren, Li, Naseriparsa, and Xia}{Yu
  et~al\mbox{.}}{2022}]%
        {yu2022graph}
\bibfield{author}{\bibinfo{person}{Shuo Yu}, \bibinfo{person}{Jing Ren},
  \bibinfo{person}{Shihao Li}, \bibinfo{person}{Mehdi Naseriparsa}, {and}
  \bibinfo{person}{Feng Xia}.} \bibinfo{year}{2022}\natexlab{}.
\newblock \showarticletitle{Graph Learning for Fake Review Detection}.
\newblock \bibinfo{journal}{\emph{Frontiers in Artificial Intelligence}}
  \bibinfo{volume}{5} (\bibinfo{year}{2022}), \bibinfo{pages}{229--270}.
\newblock


\bibitem[\protect\citeauthoryear{Yu, Xia, Sun, Tang, Yan, and Lee}{Yu
  et~al\mbox{.}}{2020b}]%
        {yu2020detecting}
\bibfield{author}{\bibinfo{person}{Shuo Yu}, \bibinfo{person}{Feng Xia},
  \bibinfo{person}{Yuchen Sun}, \bibinfo{person}{Tao Tang},
  \bibinfo{person}{Xiaoran Yan}, {and} \bibinfo{person}{Ivan Lee}.}
  \bibinfo{year}{2020}\natexlab{b}.
\newblock \showarticletitle{Detecting Outlier Patterns With Query-Based
  Artificially Generated Searching Conditions}.
\newblock \bibinfo{journal}{\emph{IEEE Transactions on Computational Social
  Systems}} \bibinfo{volume}{8}, \bibinfo{number}{1} (\bibinfo{year}{2020}),
  \bibinfo{pages}{134--147}.
\newblock


\bibitem[\protect\citeauthoryear{Zafarani, Zhou, Shu, and Liu}{Zafarani
  et~al\mbox{.}}{2019}]%
        {zafarani2019fake}
\bibfield{author}{\bibinfo{person}{Reza Zafarani}, \bibinfo{person}{Xinyi
  Zhou}, \bibinfo{person}{Kai Shu}, {and} \bibinfo{person}{Huan Liu}.}
  \bibinfo{year}{2019}\natexlab{}.
\newblock \showarticletitle{Fake news research: Theories, detection strategies,
  and open problems}. In \bibinfo{booktitle}{\emph{Proceedings of the 25th ACM
  SIGKDD International Conference on Knowledge Discovery \& Data Mining}}.
  \bibinfo{pages}{3207--3208}.
\newblock


\bibitem[\protect\citeauthoryear{Zamini and Hasheminejad}{Zamini and
  Hasheminejad}{2019}]%
        {zamini2019comprehensive}
\bibfield{author}{\bibinfo{person}{Mohamad Zamini} {and} \bibinfo{person}{Seyed
  Mohammad~Hossein Hasheminejad}.} \bibinfo{year}{2019}\natexlab{}.
\newblock \showarticletitle{A comprehensive survey of anomaly detection in
  banking, wireless sensor networks, social networks, and healthcare}.
\newblock \bibinfo{journal}{\emph{Intelligent Decision Technologies}}
  \bibinfo{volume}{13}, \bibinfo{number}{2} (\bibinfo{year}{2019}),
  \bibinfo{pages}{229--270}.
\newblock


\bibitem[\protect\citeauthoryear{Zeng, Shen, Zhou, Wu, Fan, Wang, and
  Stanley}{Zeng et~al\mbox{.}}{2017}]%
        {zeng2017science}
\bibfield{author}{\bibinfo{person}{An Zeng}, \bibinfo{person}{Zhesi Shen},
  \bibinfo{person}{Jianlin Zhou}, \bibinfo{person}{Jinshan Wu},
  \bibinfo{person}{Ying Fan}, \bibinfo{person}{Yougui Wang}, {and}
  \bibinfo{person}{H~Eugene Stanley}.} \bibinfo{year}{2017}\natexlab{}.
\newblock \showarticletitle{The science of science: From the perspective of
  complex systems}.
\newblock \bibinfo{journal}{\emph{Physics Reports}}  \bibinfo{volume}{714}
  (\bibinfo{year}{2017}), \bibinfo{pages}{1--73}.
\newblock


\bibitem[\protect\citeauthoryear{Zhang, Song, Huang, Swami, and Chawla}{Zhang
  et~al\mbox{.}}{2019}]%
        {zhang2019heterogeneous}
\bibfield{author}{\bibinfo{person}{Chuxu Zhang}, \bibinfo{person}{Dongjin
  Song}, \bibinfo{person}{Chao Huang}, \bibinfo{person}{Ananthram Swami}, {and}
  \bibinfo{person}{Nitesh~V Chawla}.} \bibinfo{year}{2019}\natexlab{}.
\newblock \showarticletitle{Heterogeneous graph neural network}. In
  \bibinfo{booktitle}{\emph{Proceedings of the 25th ACM SIGKDD international
  conference on knowledge discovery \& data mining}}.
  \bibinfo{pages}{793--803}.
\newblock


\bibitem[\protect\citeauthoryear{Zhang and Davidson}{Zhang and
  Davidson}{2021}]%
        {zhang2021towards}
\bibfield{author}{\bibinfo{person}{Hongjing Zhang} {and} \bibinfo{person}{Ian
  Davidson}.} \bibinfo{year}{2021}\natexlab{}.
\newblock \showarticletitle{Towards fair deep anomaly detection}. In
  \bibinfo{booktitle}{\emph{Proceedings of the 2021 ACM Conference on Fairness,
  Accountability, and Transparency}}. \bibinfo{pages}{138--148}.
\newblock


\bibitem[\protect\citeauthoryear{Zhang, Li, Yu, Li, Hui, and Zheng}{Zhang
  et~al\mbox{.}}{2020b}]%
        {zhang2020urban}
\bibfield{author}{\bibinfo{person}{Mingyang Zhang}, \bibinfo{person}{Tong Li},
  \bibinfo{person}{Yue Yu}, \bibinfo{person}{Yong Li}, \bibinfo{person}{Pan
  Hui}, {and} \bibinfo{person}{Yu Zheng}.} \bibinfo{year}{2020}\natexlab{b}.
\newblock \showarticletitle{Urban Anomaly Analytics: Description, Detection,
  and Prediction}.
\newblock \bibinfo{journal}{\emph{IEEE Transactions on Big Data}}
  \bibinfo{volume}{8}, \bibinfo{number}{3} (\bibinfo{year}{2020}),
  \bibinfo{pages}{809--826}.
\newblock


\bibitem[\protect\citeauthoryear{Zhang, Yin, Chen, Hung, Huang, and Cui}{Zhang
  et~al\mbox{.}}{2020c}]%
        {zhang2020gcn}
\bibfield{author}{\bibinfo{person}{Shijie Zhang}, \bibinfo{person}{Hongzhi
  Yin}, \bibinfo{person}{Tong Chen}, \bibinfo{person}{Quoc Viet~Nguyen Hung},
  \bibinfo{person}{Zi Huang}, {and} \bibinfo{person}{Lizhen Cui}.}
  \bibinfo{year}{2020}\natexlab{c}.
\newblock \showarticletitle{{GCN-Based} User Representation Learning for
  Unifying Robust Recommendation and Fraudster Detection}. In
  \bibinfo{booktitle}{\emph{Proceedings of the 43rd International ACM SIGIR
  Conference on Research and Development in Information Retrieval}}.
  \bibinfo{pages}{689--698}.
\newblock


\bibitem[\protect\citeauthoryear{Zhang, Cui, and Zhu}{Zhang
  et~al\mbox{.}}{2020a}]%
        {zhang2020deep}
\bibfield{author}{\bibinfo{person}{Ziwei Zhang}, \bibinfo{person}{Peng Cui},
  {and} \bibinfo{person}{Wenwu Zhu}.} \bibinfo{year}{2020}\natexlab{a}.
\newblock \showarticletitle{Deep learning on graphs: A survey}.
\newblock \bibinfo{journal}{\emph{IEEE Transactions on Knowledge and Data
  Engineering}} \bibinfo{volume}{34}, \bibinfo{number}{1}
  (\bibinfo{year}{2020}), \bibinfo{pages}{249--270}.
\newblock


\bibitem[\protect\citeauthoryear{Zhao, Wang, Shi, Hu, Song, and Ye}{Zhao
  et~al\mbox{.}}{2021}]%
        {zhao2021heterogeneous}
\bibfield{author}{\bibinfo{person}{Jianan Zhao}, \bibinfo{person}{Xiao Wang},
  \bibinfo{person}{Chuan Shi}, \bibinfo{person}{Binbin Hu},
  \bibinfo{person}{Guojie Song}, {and} \bibinfo{person}{Yanfang Ye}.}
  \bibinfo{year}{2021}\natexlab{}.
\newblock \showarticletitle{Heterogeneous graph structure learning for graph
  neural networks}. In \bibinfo{booktitle}{\emph{Proceedings of the AAAI
  Conference on Artificial Intelligence}}, Vol.~\bibinfo{volume}{35}.
  \bibinfo{pages}{4697--4705}.
\newblock


\bibitem[\protect\citeauthoryear{Zhao, Deng, Yu, Jiang, Wang, and Jiang}{Zhao
  et~al\mbox{.}}{2020}]%
        {zhao2020error}
\bibfield{author}{\bibinfo{person}{Tong Zhao}, \bibinfo{person}{Chuchen Deng},
  \bibinfo{person}{Kaifeng Yu}, \bibinfo{person}{Tianwen Jiang},
  \bibinfo{person}{Daheng Wang}, {and} \bibinfo{person}{Meng Jiang}.}
  \bibinfo{year}{2020}\natexlab{}.
\newblock \showarticletitle{{Error-Bounded} Graph Anomaly Loss for GNNs}. In
  \bibinfo{booktitle}{\emph{Proceedings of the 29th ACM International
  Conference on Information \& Knowledge Management}}.
  \bibinfo{pages}{1873--1882}.
\newblock


\bibitem[\protect\citeauthoryear{Zheng, Li, Li, Li, and Gao}{Zheng
  et~al\mbox{.}}{2019}]%
        {zheng2019addgraph}
\bibfield{author}{\bibinfo{person}{Li Zheng}, \bibinfo{person}{Zhenpeng Li},
  \bibinfo{person}{Jian Li}, \bibinfo{person}{Zhao Li}, {and}
  \bibinfo{person}{Jun Gao}.} \bibinfo{year}{2019}\natexlab{}.
\newblock \showarticletitle{{AddGraph}: Anomaly Detection in Dynamic Graph
  Using Attention-based Temporal {GCN}.}. In \bibinfo{booktitle}{\emph{IJCAI}}.
  \bibinfo{pages}{4419--4425}.
\newblock


\bibitem[\protect\citeauthoryear{Zheng, Liu, and Hsieh}{Zheng
  et~al\mbox{.}}{2013}]%
        {zheng2013u}
\bibfield{author}{\bibinfo{person}{Yu Zheng}, \bibinfo{person}{Furui Liu},
  {and} \bibinfo{person}{Hsun-Ping Hsieh}.} \bibinfo{year}{2013}\natexlab{}.
\newblock \showarticletitle{U-air: When urban air quality inference meets big
  data}. In \bibinfo{booktitle}{\emph{Proceedings of the 19th ACM SIGKDD
  international conference on Knowledge discovery and data mining}}.
  \bibinfo{pages}{1436--1444}.
\newblock


\bibitem[\protect\citeauthoryear{Zhong, Li, Kong, Liu, Li, and Li}{Zhong
  et~al\mbox{.}}{2019}]%
        {zhong2019graph}
\bibfield{author}{\bibinfo{person}{Jia-Xing Zhong}, \bibinfo{person}{Nannan
  Li}, \bibinfo{person}{Weijie Kong}, \bibinfo{person}{Shan Liu},
  \bibinfo{person}{Thomas~H Li}, {and} \bibinfo{person}{Ge Li}.}
  \bibinfo{year}{2019}\natexlab{}.
\newblock \showarticletitle{Graph convolutional label noise cleaner: Train a
  plug-and-play action classifier for anomaly detection}. In
  \bibinfo{booktitle}{\emph{Proceedings of the IEEE/CVF Conference on Computer
  Vision and Pattern Recognition}}. \bibinfo{pages}{1237--1246}.
\newblock


\bibitem[\protect\citeauthoryear{Zhong, Cao, Sheng, Guo, and Wang}{Zhong
  et~al\mbox{.}}{2020a}]%
        {zhong2020integrating}
\bibfield{author}{\bibinfo{person}{Lei Zhong}, \bibinfo{person}{Juan Cao},
  \bibinfo{person}{Qiang Sheng}, \bibinfo{person}{Junbo Guo}, {and}
  \bibinfo{person}{Ziang Wang}.} \bibinfo{year}{2020}\natexlab{a}.
\newblock \showarticletitle{Integrating Semantic and Structural Information
  with Graph Convolutional Network for Controversy Detection}. In
  \bibinfo{booktitle}{\emph{Proceedings of the 58th Annual Meeting of the
  Association for Computational Linguistics}}. \bibinfo{pages}{515--526}.
\newblock


\bibitem[\protect\citeauthoryear{Zhong, Liu, Ao, Hu, Feng, Tang, and He}{Zhong
  et~al\mbox{.}}{2020b}]%
        {zhong2020financial}
\bibfield{author}{\bibinfo{person}{Qiwei Zhong}, \bibinfo{person}{Yang Liu},
  \bibinfo{person}{Xiang Ao}, \bibinfo{person}{Binbin Hu},
  \bibinfo{person}{Jinghua Feng}, \bibinfo{person}{Jiayu Tang}, {and}
  \bibinfo{person}{Qing He}.} \bibinfo{year}{2020}\natexlab{b}.
\newblock \showarticletitle{Financial defaulter detection on online credit
  payment via multi-view attributed heterogeneous information network}. In
  \bibinfo{booktitle}{\emph{Proceedings of The Web Conference 2020}}.
  \bibinfo{pages}{785--795}.
\newblock


\bibitem[\protect\citeauthoryear{Zhou, Cao, Zhang, Trajcevski, Zhong, and
  Geng}{Zhou et~al\mbox{.}}{2019a}]%
        {zhou2019meta}
\bibfield{author}{\bibinfo{person}{Fan Zhou}, \bibinfo{person}{Chengtai Cao},
  \bibinfo{person}{Kunpeng Zhang}, \bibinfo{person}{Goce Trajcevski},
  \bibinfo{person}{Ting Zhong}, {and} \bibinfo{person}{Ji Geng}.}
  \bibinfo{year}{2019}\natexlab{a}.
\newblock \showarticletitle{Meta-gnn: On few-shot node classification in graph
  meta-learning}. In \bibinfo{booktitle}{\emph{Proceedings of the 28th ACM
  International Conference on Information and Knowledge Management}}.
  \bibinfo{pages}{2357--2360}.
\newblock


\bibitem[\protect\citeauthoryear{Zhou, Cui, Hu, Zhang, Yang, Liu, Wang, Li, and
  Sun}{Zhou et~al\mbox{.}}{2020}]%
        {zhou2020graph}
\bibfield{author}{\bibinfo{person}{Jie Zhou}, \bibinfo{person}{Ganqu Cui},
  \bibinfo{person}{Shengding Hu}, \bibinfo{person}{Zhengyan Zhang},
  \bibinfo{person}{Cheng Yang}, \bibinfo{person}{Zhiyuan Liu},
  \bibinfo{person}{Lifeng Wang}, \bibinfo{person}{Changcheng Li}, {and}
  \bibinfo{person}{Maosong Sun}.} \bibinfo{year}{2020}\natexlab{}.
\newblock \showarticletitle{Graph neural networks: A review of methods and
  applications}.
\newblock \bibinfo{journal}{\emph{AI Open}}  \bibinfo{volume}{1}
  (\bibinfo{year}{2020}), \bibinfo{pages}{57--81}.
\newblock


\bibitem[\protect\citeauthoryear{Zhou and Zafarani}{Zhou and Zafarani}{2020}]%
        {zhou2020survey}
\bibfield{author}{\bibinfo{person}{Xinyi Zhou} {and} \bibinfo{person}{Reza
  Zafarani}.} \bibinfo{year}{2020}\natexlab{}.
\newblock \showarticletitle{A survey of fake news: Fundamental theories,
  detection methods, and opportunities}.
\newblock \bibinfo{journal}{\emph{ACM Computing Surveys (CSUR)}}
  \bibinfo{volume}{53}, \bibinfo{number}{5} (\bibinfo{year}{2020}),
  \bibinfo{pages}{1--40}.
\newblock


\bibitem[\protect\citeauthoryear{Zhou, Zafarani, Shu, and Liu}{Zhou
  et~al\mbox{.}}{2019b}]%
        {zhou2019fake}
\bibfield{author}{\bibinfo{person}{Xinyi Zhou}, \bibinfo{person}{Reza
  Zafarani}, \bibinfo{person}{Kai Shu}, {and} \bibinfo{person}{Huan Liu}.}
  \bibinfo{year}{2019}\natexlab{b}.
\newblock \showarticletitle{Fake news: Fundamental theories, detection
  strategies and challenges}. In \bibinfo{booktitle}{\emph{Proceedings of the
  twelfth ACM international conference on web search and data mining}}.
  \bibinfo{pages}{836--837}.
\newblock


\bibitem[\protect\citeauthoryear{Zhu, Ma, and Liu}{Zhu et~al\mbox{.}}{2020b}]%
        {zhu2020deepad}
\bibfield{author}{\bibinfo{person}{Dali Zhu}, \bibinfo{person}{Yuchen Ma},
  {and} \bibinfo{person}{Yinlong Liu}.} \bibinfo{year}{2020}\natexlab{b}.
\newblock \showarticletitle{{DeepAD}: A Joint Embedding Approach for Anomaly
  Detection on Attributed Networks}. In \bibinfo{booktitle}{\emph{International
  Conference on Computational Science}}. Springer, \bibinfo{pages}{294--307}.
\newblock


\bibitem[\protect\citeauthoryear{Zhu, Zhang, Cui, and Zhu}{Zhu
  et~al\mbox{.}}{2019}]%
        {zhu2019robust}
\bibfield{author}{\bibinfo{person}{Dingyuan Zhu}, \bibinfo{person}{Ziwei
  Zhang}, \bibinfo{person}{Peng Cui}, {and} \bibinfo{person}{Wenwu Zhu}.}
  \bibinfo{year}{2019}\natexlab{}.
\newblock \showarticletitle{Robust graph convolutional networks against
  adversarial attacks}. In \bibinfo{booktitle}{\emph{Proceedings of the 25th
  ACM SIGKDD International Conference on Knowledge Discovery \& Data Mining}}.
  \bibinfo{pages}{1399--1407}.
\newblock


\bibitem[\protect\citeauthoryear{Zhu, Luo, Li, Bu, Zhou, Zhang, and Lu}{Zhu
  et~al\mbox{.}}{2020a}]%
        {zhu2020heterogeneous}
\bibfield{author}{\bibinfo{person}{Yong-Nan Zhu}, \bibinfo{person}{Xiaotian
  Luo}, \bibinfo{person}{Yu-Feng Li}, \bibinfo{person}{Bin Bu},
  \bibinfo{person}{Kaibo Zhou}, \bibinfo{person}{Wenbin Zhang}, {and}
  \bibinfo{person}{Mingfan Lu}.} \bibinfo{year}{2020}\natexlab{a}.
\newblock \showarticletitle{Heterogeneous Mini-Graph Neural Network and Its
  Application to Fraud Invitation Detection}. In \bibinfo{booktitle}{\emph{2020
  IEEE International Conference on Data Mining (ICDM)}}. IEEE,
  \bibinfo{pages}{891--899}.
\newblock


\end{thebibliography}
	
	
\end{document}